\documentclass{article}
\usepackage[main, final]{neurips_2026}
\makeatletter
\renewcommand{\@noticestring}{Accepted at the \@neuripsordinal\ Conference on Neural Information Processing Systems (NeurIPS \@neuripsyear).}
\makeatother

\usepackage[utf8]{inputenc}
\usepackage[T1]{fontenc}
\usepackage{hyperref}
\hypersetup{hidelinks}
\usepackage{url}
\usepackage{booktabs}
\usepackage{amsfonts}
\usepackage{amsmath}
\usepackage{amssymb}
\usepackage{amsthm}
\usepackage{nicefrac}
\usepackage{xcolor}
\usepackage{algorithm}
\usepackage{algorithmic}
\usepackage{graphicx}
\usepackage{subcaption}
\usepackage{enumitem}

\newtheorem{theorem}{Theorem}
\newtheorem{lemma}[theorem]{Lemma}
\newtheorem{proposition}[theorem]{Proposition}
\newtheorem{corollary}[theorem]{Corollary}
\newtheorem{definition}[theorem]{Definition}
\newtheorem{remark}[theorem]{Remark}

\newcommand{\Exploit}{\mathrm{Exploit}}
\newcommand{\Regret}{\mathrm{Regret}}

\newcommand{\bbR}{\mathbb{R}}

\newcommand{\pibar}{\bar{\pi}}
\newcommand{\Tbar}{\bar{T}}
\newcommand{\dbar}{\bar{d}}
\newcommand{\disc}{\beta}

\title{Virtual Double Oracle: Faster Convergence
by Leveraging Strategies That Were Watching All
Along}

\author{
  Ruilan Wang\\
  Department of Mathematics\\
  London School of Economics\\
  London, UK\\
  \texttt{R.Wang27@lse.ac.uk}
  \And
  Francisco Aristi Reina\\
  Department of Mathematics\\
  London School of Economics\\
  London, UK\\
  \texttt{f.aristi-reina@lse.ac.uk}
  \And
  Katerina Papadaki\\
  Department of Mathematics\\
  London School of Economics\\
  London, UK\\
  \texttt{k.p.papadaki@lse.ac.uk}
}

\begin{document}
\maketitle

\begin{abstract} Double Oracle (DO) and PSRO approximate Nash equilibria in two-player zero-sum games by iteratively expanding a strategy population with best responses supplied by an oracle. Online Double Oracle (ODO) interleaves multiplicative weights updates (MWU) with population discovery and obtains an $O(\sqrt{k\ln k/T})$ rate, where $T$ is the number of learning rounds and $k$ is the larger of the two players' final population sizes. We
identify the \emph{reinitialisation} step as an
underexplored design choice: at each expansion,
MWU-based methods reset newly discovered
strategies to a uniform prior. We introduce
Virtual Double Oracle (VDO), which exploits a
bilinear identity---a strategy's full hindsight
loss history collapses to its utility against the
time-averaged opponent---to initialise each new
strategy with the MWU weight it would have
had from the start. We prove that VDO achieves $O(\sqrt{k/T})$ exploitability over the discovered population as it grows, provided the cumulative cost of adding strategies is at most linear in $k$, a condition supported by our empirical measurements.
Discounted VDO (dVDO) moderates historical
preferences to reduce over-concentration at
reinitialisation. Experiments across matrix games,
poker, and Goofspiel show that, with a suitable
discount, dVDO can substantially reduce
finite-horizon exploitability relative to both VDO
and uniform-reset ODO, and can improve the
discovered strategy populations.
\end{abstract}

\section{Introduction}
\label{sec:intro}

Computing Nash equilibria (NE) in large two-player
zero-sum games is a fundamental challenge in
multi-agent systems, with applications spanning
security games~\citep{tambe2011security},
competitive game
playing~\citep{vinyals2019grandmaster}, and
multi-agent reinforcement
learning~\citep{lanctot2017unified}. While the
minimax theorem guarantees existence, computing NE
in games with large strategy spaces remains
demanding. Empirical evidence suggests that many
benchmark games admit compact effective strategy
sets~\citep{dinh2022online}, while real-world
games exhibit structured non-transitive geometry
that motivates population-based
methods~\citep{czarnecki2020real}. These methods, such as Double Oracle (DO)~\citep{mcmahan2003planning} and PSRO~\citep{lanctot2017unified}, maintain a restricted game over a subset of each player's strategy set (\emph{population}). They solve the
restricted game for an NE and then expand each
population with a best response computed by an
oracle over the full set of pure strategies,
repeating until the populations contain the
support of an NE. Although a linear
program solves an explicit two-player zero-sum
matrix game in polynomial time, the full matrix is
too large to enumerate in the intended
applications.

\textbf{PSRO in practice mimics online
learning.}\quad In practice, the restricted
game is rarely solved to exact equilibrium. To
scale to large games, or to games in which neither
the exact solution nor the game model is well
understood---as in StarCraft~II, where
AlphaStar~\citep{vinyals2019grandmaster} trains a
league of agents with RL-based best
responses---PSRO methods approximate both the
restricted-game solution and the best-response
oracle. For example, the original PSRO approximates the restricted-game NE with projected replicator dynamics~\citep{lanctot2017unified}, and regularised replicator dynamics~\citep{wang2023regularization} deliberately truncates the solver at a regret threshold. In practice, these iterative solvers run within a finite computational budget. We argue that approximate
population-based game solving is therefore best
viewed as a continuous online learning process
over an expanding strategy set, not a sequence of
subgame solves. In online learning, play
proceeds in rounds of the same game; in each
round, every player chooses a strategy and
receives a loss, treating the other player's
choices as part of the environment (nature). This online perspective is not only more faithful to
practice but also yields theoretical traction:
Online Double Oracle (ODO)~\citep{dinh2022online}
formalises this view and obtains an $O(\sqrt{k\ln k/T})$ rate when MWU steps and best-response checks alternate at every step.
Here $k$ is the larger of the two players' final population sizes and $T$ is the number of learning rounds.

\textbf{Reinitialisation as a hidden design
axis.}\quad When a new strategy is
discovered, an online meta-solver must decide how
it joins the weights already accumulated by the
incumbent strategies. Classical DO sidesteps this
by re-solving restricted games from scratch;
ODO runs MWU on each restricted population
but restarts with uniform weights at expansion,
treating the newcomer as if it had no history,
although it was selected precisely because it
responds best to the opponent's play. The
uniform reset is natural only if each restricted
game is viewed as an independent problem, solved
by a fresh MWU instance that assigns equal initial
weight to every strategy in the population. We
argue that this choice is load-bearing: the
$\sqrt{k}$ factor in ODO's windowed regret
analysis is the cost of treating successive
populations as separate learners. The principled
choice follows from a simple identity: in bilinear
games, any strategy's hindsight loss history
compresses into its utility against the
time-average opponent, so we can recover exactly
the MWU weight it would have had from the start.
New strategies were watching all along.

\textbf{VDO and discounted VDO.}\quad We use
the bilinear identity not as a warm start, but to
preserve a single expanding-support MWU trajectory
and to account explicitly for the cost of
inserting new strategies, instead of restarting
the regret analysis at every expansion. We call
the resulting algorithm Virtual Double Oracle (VDO).
Discounted VDO (dVDO) adjusts the influence
of the reconstructed history using a single
discount $\disc > 0$, controlling the strength of
the initialisation over finite horizons.
Both are drop-in replacements for uniform
reinitialisation that require only one running
average per player.

\textbf{Contributions.}\quad
(1)~We identify reinitialisation as a
    distinct design axis in expanding-population
    online learning and derive VDO from a
    bilinear identity (\S\ref{sec:vdo}).
    (2)~We show that VDO plays a virtual
    MWU distribution conditioned on the available
    strategies, and quantify the regret caused by
    revealing new strategies
    (Theorem~\ref{thm:vdo}); a linear
    insertion-cost budget yields $O(\sqrt{k/T})$
    discovered-population exploitability
    (Corollary~\ref{cor:budget}), and we extend
    the same insertion accounting to dVDO
    (Theorem~\ref{thm:unified}).
    (3)~We evaluate historical
    initialisation and discounting across matrix
    games and tabular imperfect-information games
    (Kuhn, Leduc, Goofspiel-4, etc.), using
    online exploitability and restricted-NE exploitability (RNE)
    to distinguish effects on the played mixture
    and the discovered population
    (\S\ref{sec:experiments}).
    (4)~We evaluate the robustness of VDO through PSRO experiments with approximate best responses.

\section{Background}
\label{sec:background}

\textbf{Two-player zero-sum games.}\quad A
two-player zero-sum normal-form game is a triple
$(S_1, S_2, A)$ where $S_1, S_2$ are finite pure
strategy sets with $|S_1| = n$, $|S_2| = m$, and
$A \in [-1, 1]^{n \times m}$ is player~1's payoff
matrix; player~2's payoff is $-A$. For our
extensive-form experiments (Kuhn, Leduc,
Goofspiel-4), $S_i$ denotes the finite set of pure
normal-form strategies induced by the game. A mixed strategy is a probability distribution over the finite set of pure strategies, $\pi_i\in\Delta(S_i)$, and we call the pair $(\pi_1,\pi_2)$ a strategy profile. For player $i$, $-i$ denotes the other player.  The utility
functions map strategy profiles to real numbers:
$u_1(\pi_1, \pi_2) = \pi_1^\top A \pi_2$,
$u_2(\pi_2, \pi_1) = -\pi_1^\top A \pi_2$. A population is a subset of a player's pure strategies, $\Pi_1\subseteq S_1$, $\Pi_2\subseteq S_2$, and strategies restricted to a population are distributions in $\Delta(\Pi_i)$. In other words, $k=\max\{|\Pi_1|,|\Pi_2|\}$ for the final populations.
We define the restricted exploitability of
a strategy profile, $\Exploit_{\Pi_1,
\Pi_2}(\pi_1, \pi_2) := \max_{s \in \Pi_1} u_1(s,
\pi_2) + \max_{s \in \Pi_2} u_2(s, \pi_1) \geq 0$,
with full-game $\Exploit_S := \Exploit_{S_1,
S_2}$. The main objective of this work is to
minimise the full-game exploitability
$\Exploit_S$. In a two-player zero-sum
game, $\Exploit_S = 0$ is a necessary and
sufficient condition for the strategy profile
$(\pi_1, \pi_2)$ to be a Nash equilibrium.

\quad

\textbf{Loss convention and no regret.} Let $\pi_i^t$ be the mixed strategy played by player $i$ in round $t$, for $t=1,\ldots,T$. The loss of pure strategy $s$ in that round is its negative utility against the opponent's played strategy, $\ell_i^t(s)=-u_i(s,\pi_{-i}^t)$. A player's regret is the difference between the cumulative loss of the strategies it played and that of the best fixed strategy in hindsight, $\Regret_i(T)=\sum_{t=1}^T\langle\pi_i^t,\ell_i^t\rangle-\min_{s\in S_i}\sum_{t=1}^T\ell_i^t(s)$. Write $s^*$ for any strategy attaining this minimum. Regret measures the
player's learning progress; a player is
\emph{no-regret} if its average regret
$\Regret_i(T)/T$ converges to zero as $T \to
\infty$. Writing $\pibar_i^T=T^{-1}\sum_{t=1}^T\pi_i^t$ for player $i$'s average strategy, when both players run no-regret algorithms, the regret-exploitability
bridge~\citep{freund1999adaptive} gives
$\Exploit_S(\pibar^T_1, \pibar^T_2) \leq
(\Regret_1(T) + \Regret_2(T)) / T$.

\textbf{Online learning and MWU.}\quad
For a single MWU learner, let $\Pi=\Pi_i$ be its fixed population, $k_i=|\Pi|$ its size, and $\eta>0$ its learning rate. Multiplicative Weights Update (MWU)~\citep{arora2012multiplicative,freund1999adaptive} maintains weights $w^t\in\bbR^{|\Pi|}_{>0}$, plays $\pi^t=w^t/\|w^t\|_1$, and updates $w^{t+1}(s)=w^t(s)\exp(-\eta\ell^t(s))$. With uniform
initialisation and $\eta = \sqrt{2 \ln |\Pi| /
T}$, MWU achieves regret at most $\sqrt{2T \ln
|\Pi|}$~\citep{arora2012multiplicative}. The
non-uniform-initialisation
extension~\citep[Lemma~5.1]{cesabianchi2006prediction}
is central to our analysis: for any initial
probability distribution $\pi^1 \in \Delta(\Pi)$
with $\pi^1(s^*) > 0$,
\begin{equation}
    \sum_{t=1}^T \langle \pi^t, \ell^t \rangle -
    \sum_{t=1}^T \ell^t(s^*) \leq \frac{1}{\eta}
    \ln \frac{1}{\pi^1(s^*)} + \frac{\eta T}{2}.
    \label{eq:nonuniform_mwu}
\end{equation}
The initialisation affects this regret bound through the probability assigned to the best fixed strategy in hindsight.
Restarting MWU uniformly discards the preferences learned in previous rounds.

\textbf{The DO/PSRO family.}\quad Double Oracle (DO)~\citep{mcmahan2003planning} solves a game by expanding each player's restricted strategy population. At each
iteration, DO computes a Nash equilibrium of the
restricted game induced by the current
populations; a best-response oracle then finds,
for each player, the pure strategy in the
\emph{full} game that best responds to the
opponent's restricted equilibrium strategy, and
adds it to that player's population. DO terminates
when no best response improves on the restricted
equilibrium, i.e., no new strategy is added.
PSRO~\citep{lanctot2017unified} extends DO to large games by approximating the best-response oracle with reinforcement learning and by allowing general meta-strategy solvers for the restricted game. ODO~\citep{dinh2022online}
replaces exact restricted-game solving with MWU
over the restricted population, expanding
it with oracle best responses to the opponent's current average strategy. As in standard online
learning, each ODO player observes only the losses
of the current iteration and treats the other
player as nature.

We assume a richer information
structure: a centralised DO/PSRO
setting in which the meta-solver, which computes
each player's mixture over its current
population, has access to
both players' current mixed strategies and
maintains their historical averages. The
best-response oracle is a separate
strategy-generation subroutine and provides no
additional information to the learner. A payoff
evaluator suffices; the full normal-form payoff
matrix need not be explicitly available, and our theoretical
guarantees assume exact payoff evaluations.

ODO partitions iterations into
\emph{windows}: window $T_j$ is an interval
between successive population-expansion resets of the player under analysis, with the final interval truncated at the evaluation horizon. The players' reset times may differ; we suppress the player index on window quantities. Write $J$ for the number of
windows, $|T_j|$ for their lengths,
$\Tbar_j=\sum_{r<j}|T_r|$, and
$T=\sum_{j=1}^J|T_j|$ for the number of learning
rounds. Let $\Pi_i^{\mathrm{fin}}$ include all
strategies inserted through the end of round $T$,
$k_i=|\Pi_i^{\mathrm{fin}}|$, and
$k=\max\{k_1,k_2\}$.

\begin{theorem}[ODO convergence, {\citealt[Theorem
5]{dinh2022online}}]
\label{thm:odo} Under its window-local response
and learning-rate protocol, ODO has exploitability
$O(\sqrt{k\ln k/T})$.
\end{theorem}

Its analysis sums the window-wise MWU bounds. If a
player has $K_j$ strategies in window $j$,
Cauchy--Schwarz gives $\sum_{j=1}^J\sqrt{|T_j|\ln K_j}
\leq\sqrt{T\sum_{j=1}^J\ln K_j}
\leq\sqrt{TJ\ln k}$. When $J$ scales with $k$, this yields the
additional $\sqrt{k}$ factor. Each window restarts from a uniform prior over its whole population and pays a prior cost $\ln K_j$ again.
For window sizes $K_j=j$, $j=1,\ldots,k$, these costs sum to $\sum_j\ln K_j=\ln k!$, of order $k\ln k$. VDO retains incumbents'
historical weights and charges only the weight
added at expansion (Theorem~\ref{thm:vdo}).

\section{Virtual Double Oracle}
\label{sec:vdo}

\subsection{Reconstructing Historical Weights}

In ODO, strategies are
\emph{discovered} incrementally, and whenever the
oracle adds a new strategy, MWU is reinitialised
over the enlarged population, discarding all loss
information accumulated in previous windows.
To recover the weights implied by this discarded history, imagine that strategies discovered later
had been observing the opponent's actual play from
the beginning. Although unavailable for play, they
have a well-defined \emph{virtual history} of
losses, from which VDO reconstructs the MWU
weights they would have accumulated. They enter
with these weights rather than starting from
scratch.

This relies on the richer information
structure of Section~\ref{sec:background}, in which
each player records the opponent's past play (in
bilinear games, its running average suffices), so
that the virtual losses of a strategy can be
computed at the moment the oracle reveals it,
unlike the standard online-learning setting,
where the opponent is treated as
nature.

\begin{definition}[Virtual cumulative loss]
For player $i$, strategy $s\in S_i$ and rounds $1\le a\le b\le T$, define $L_i^{[a:b]}(s)=\sum_{t=a}^{b}\ell_i^t(s)=-\sum_{t=a}^{b}u_i(s,\pi_{-i}^t)$. Write $L_i^\tau(s)=L_i^{[1:\tau]}(s)$ for the cumulative loss through round $\tau\ge1$, and set $L_i^0(s)=0$.
\end{definition}

\begin{lemma}[Cumulative loss against the average opponent]
\label{lem:bilinear}
For any $s\in S_i$, any opponent sequence $\pi^1_{-i},\ldots,\pi^T_{-i}$, and $1\le\tau\le T$, $L_i^\tau(s)=-\tau\,u_i(s,\pibar_{-i}^{[1:\tau]})$, where $\pibar_{-i}^{[1:\tau]}=(1/\tau)\sum_{t=1}^{\tau}\pi_{-i}^t$ is the opponent's average strategy over rounds $1$ through $\tau$.
\end{lemma}
\begin{proof}
For player 1: $L^\tau_1(s) = -\sum_{t=1}^\tau
e_s^\top A \pi^t_2 = -e_s^\top A \sum_{t=1}^\tau
\pi^t_2 = -\tau \cdot u_1(s,
\pibar^{[1:\tau]}_2)$. Player~2 follows by
symmetry.
\end{proof}

Lemma~\ref{lem:bilinear} makes this
reconstruction practical: one payoff evaluation
against the historical opponent average, scaled by
the number of observed rounds, gives the
cumulative loss of a strategy, including one
discovered only later. The identity uses the opponent strategies actually played. Thus newcomers and
incumbents can be compared using the same history,
without replaying every past round.

\subsection{VDO Reinitialisation}

Suppose strategy $s$ joins the population after
$\tau := \Tbar_{j+1}$ rounds have been observed.
Had MWU run over all $k_i$ strategies from $t=1$
with uniform initialisation $w^1_V = 1/k_i$, the
\emph{virtual weight} entering round $\tau+1$
would be
\begin{equation}
    w^{\tau+1}_V(s) = \frac{1}{k_i} \exp\!\left(\eta
    \tau \cdot u_i\!\left(s,
    \pibar^{[1:\tau]}_{-i}\right)\right), \qquad
    \tau = \Tbar_{j+1},
    \label{eq:virtual_weight}
\end{equation}
by Lemma~\ref{lem:bilinear}. Normalising over the enlarged population $\Pi^{(j+1)}$ (the $1/k_i$ factors cancel), and writing $\pibar^{\mathrm{gl}}_{-i}=\pibar^{[1:\tau]}_{-i}$ and $\lambda=\eta\tau$ gives the virtual
reinitialisation, a softmax over historical utilities:
\begin{equation}
    \pi_{\text{VDO}}(s) =
    \frac{\exp\!\left(\lambda \cdot u_i(s,
    \pibar^{\mathrm{gl}}_{-i})\right)}{\sum_{s'
    \in \Pi^{(j+1)}} \exp\!\left(\lambda \cdot
    u_i(s', \pibar^{\mathrm{gl}}_{-i})\right)},
    \qquad \lambda = \eta \Tbar_{j+1}.
    \label{eq:vdo_reinit}
\end{equation}

This reinitialisation has a
\emph{virtual-play} interpretation. Imagine an MWU
learner tracking all strategies in the final
discovered population from round~1, updating
against the opponents \emph{actually played}. Let
$\widehat\pi_i^t$ be its distribution before
round~$t$ and $\Pi_i^t$ the strategies available
to VDO. Exact VDO plays the virtual distribution
conditioned on those available strategies:
\begin{equation}
 \pi_i^t(s)=\frac{\widehat\pi_i^t(s)}
 {\sum_{a\in\Pi_i^t}\widehat\pi_i^t(a)},
 \qquad s\in\Pi_i^t.
 \label{eq:virtual_conditioning}
\end{equation}

With $\disc=1$, incumbent strategies
accumulate weights proportional to $\exp[-\eta
L_i^{t-1}(s)]$ through ordinary MWU updates.
Equation~\eqref{eq:virtual_weight} recovers these
same historical weights for newcomers.
Reinitialising incumbents therefore preserves
their relative weights, and normalising across
available strategies gives exactly the
conditioning in~\eqref{eq:virtual_conditioning}.

The VDO/dVDO algorithm (Algorithm~\ref{alg:vdo})
uses virtual reinitialisation
(Line~\ref{line:reinit}) and a configurable
per-player response-checking schedule $m(k)$. Each player is reinitialised only when its own population grows. Its responses target the global opponent average used to reconstruct historical weights. The only
additional quantity maintained is the global
opponent average $\pibar^{\mathrm{gl}}_{-i}$.
Stored sparsely, the average uses memory
linear in the number of distinct opponent
strategies played and needs one incremental
update per step. At each expansion, computing
$u_i(s,\pibar^{\mathrm{gl}}_{-i})$ for all
$s\in\Pi_i$ requires a single matrix-vector
product in our tabular normal-form implementation;
in extensive-form implementations it would require
an evaluator for the value of each strategy
against the sparse opponent average.

In centralised PSRO, the meta-solver
already has the opponent historical policy networks that were played, and therefore a historical average distribution across networks.
Once a new policy's payoff row against the stored
opponent policy networks is evaluated, its historical
utility is a weighted sum of that
row; no replay of the historical sequence or
averaging of neural-network parameters is needed.
With simulation-based evaluation, the same
estimates can be reused. Our theorems assume exact
payoffs; finite-rollout estimation and learned
responses require empirical evaluation of the
resulting trajectories.

\begin{algorithm}[t]
\caption{Virtual Double Oracle (VDO and dVDO)}
\label{alg:vdo}
\begin{algorithmic}[1]
\REQUIRE Full strategy sets $S_1,S_2$; horizon $T$ learning rounds; learning rate $\eta$; per-player checking schedule $m(k)$; discount $\disc\in(0,\infty]$.
\STATE For each player $i$, initialise $\Pi_i=\{s_i^0\}$, $w_i^1(s_i^0)=1$, countdown $c_i=0$, and global average $\pibar_i^{\mathrm{gl}}=0$.
\FOR{$t=1,\ldots,T$}
\STATE Both players simultaneously play $\pi_i^t=w_i^t/\|w_i^t\|_1$.
\STATE Update $\pibar_i^{\mathrm{gl}}\leftarrow ((t-1)\pibar_i^{\mathrm{gl}}+\pi_i^t)/t$ for each $i$; set $\Tbar=t$.
\STATE For each $i$ and $s\in\Pi_i$, update $w_i^{t+1}(s)\leftarrow w_i^t(s)\exp(\eta u_i(s,\pi_{-i}^t))$.
\STATE Decrease each countdown: $c_i\leftarrow c_i-1$; let $D_t=\{i:c_i\leq0\}$.
\STATE For each $i\in D_t$, compute $b_i^*\in\arg\max_{s\in S_i}u_i(s,\pibar_{-i}^{\mathrm{gl}})$ using the current global averages.
\FOR{each $i\in D_t$}
\IF{$b_i^*\notin\Pi_i$}
\STATE Add $b_i^*$ to $\Pi_i$ and assign it zero mass in the accumulated play average.
\STATE Reinitialise player $i$ only: for all $s\in\Pi_i$, set $w_i^{t+1}(s)\leftarrow\exp\!\left(\frac{\eta\Tbar}{\disc}u_i(s,\pibar_{-i}^{\mathrm{gl}})\right)$. \label{line:reinit}
\ENDIF
\STATE Set $c_i\leftarrow m(|\Pi_i|)$, whether or not a strategy was added.
\ENDFOR
\ENDFOR
\STATE \textbf{return} $\pibar_i^T=T^{-1}\sum_{t=1}^T\pi_i^t$ (averages of played distributions).
\end{algorithmic}
\end{algorithm}

\subsection{From Virtual Learning to Actual
Play}
\label{sec:convergence}

The virtual-play
identity~\eqref{eq:virtual_conditioning} shows
that VDO's distribution coincides with the
virtual learner's once every
strategy in the final population has been
revealed. It does not change choices made before
discovery: the virtual learner might have assigned
probability to strategies that VDO could not yet
play. Since MWU's regret guarantee concerns
cumulative play, we must account for this
difference over the complete run.

An \emph{insertion} adds one or more
previously unavailable strategies to the
population. VDO keeps the incumbents' historical
weights and reconstructs those of the newcomers.
After $\tau$ rounds, let $W_{i,e}^-$ and
$W_{i,e}^+$ be the total weights immediately
before and after insertion~$e$, evaluated on the
same history, and let
$q_{i,e}=1-W_{i,e}^-/W_{i,e}^+$ be the share of
play assigned to the newcomers on arrival.
The insertion cost is
\begin{equation}
 \mathcal I_{i,T}=\sum_e\ln\frac{W_{i,e}^+}{W_{i,e}^-}
 \;=\;\sum_e\ln\frac{1}{1-q_{i,e}}.
 \label{eq:insertion_charge}
\end{equation}
For example, if incumbents have weights $2$
and $1$ and a newcomer receives weight $3$, VDO
plays $(1/3,1/6,1/2)$, preserving the incumbents'
$2{:}1$ ratio; the newcomer takes half the play
and the insertion costs $\ln 2$. A uniform reset
instead plays $(1/3,1/3,1/3)$. If each newcomer's
share stays bounded away from one, each insertion
has bounded cost, giving a linear cumulative
budget (Appendix~\ref{app:insertion}).

\begin{theorem}[VDO: regret across population
expansions]
\label{thm:vdo} Consider Algorithm~\ref{alg:vdo}
with $\disc=1$, exact payoff feedback, losses in
$[-1,1]$, populations that only grow, and a fixed
$\eta>0$. Let $\Pi_i^t$ be player~$i$'s population
before round~$t$, $\Pi_i^{\mathrm{fin}}$ its final
population, and $k_{i,0}=|\Pi_i^1|$. For each
player, regret against the final discovered
population satisfies
\begin{equation}
 R_{i,T}:=\sum_{t=1}^T\langle\pi_i^t,\ell_i^t\rangle
 -\min_{s\in\Pi_i^{\mathrm{fin}}}L_i^T(s)
 \leq\frac{\ln k_{i,0}+\mathcal I_{i,T}}{\eta}+\frac{\eta T}{2}.
 \label{eq:insertion_regret}
\end{equation}
Consequently, writing $C_{i,T}=\ln
k_{i,0}+\mathcal I_{i,T}$,
\begin{equation}
 \Exploit_{\Pi^{\mathrm{fin}}}(\pibar_1^T,\pibar_2^T)
 \leq\frac{C_{1,T}+C_{2,T}}{\eta T}+\eta .
 \label{eq:insertion_gap}
\end{equation}
These inequalities hold for every realised
expansion sequence.
\end{theorem}

\begin{proof}[Proof sketch; details in
Appendix~\ref{app:proof_vdo}] Between insertions,
the weights follow ordinary MWU. When new
strategies join, their reconstructed weights add
exactly the log-potential change
in~\eqref{eq:insertion_charge}; the incumbents'
historical weights are unchanged. These ordinary
updates and expansion jumps telescope over the
same run. Comparing the final sum of weights with
any final strategy's full-history weight
proves~\eqref{eq:insertion_regret}; two-player
regret gives~\eqref{eq:insertion_gap}.
\end{proof}

\begin{corollary}[A linear insertion-budget rate]
\label{cor:budget} Fix a constant $c>0$. Suppose the run uses a fixed learning rate $\eta=\sqrt{2ck/T}$ and $C_{i,T}\leq ck$ for both players at this rate. Then
$\Exploit_{\Pi^{\mathrm{fin}}}(\pibar_1^T,\pibar_2^T)
\leq 2\sqrt{2ck/T}=O(\sqrt{k/T})$.
\end{corollary} Appendix~\ref{app:insertion} further discusses the linear insertion-cost condition and provides empirical evidence supporting its plausibility.

\begin{remark}[Full-game quality]
\label{rem:scope} For the reported average,
$\Exploit_S=\Exploit_{\Pi^{\mathrm{fin}}}
+\varepsilon_{\mathrm{disc}}$, where
$\varepsilon_{\mathrm{disc}}\geq0$ is the sum of
the two players' gains from allowing deviations
outside the final populations. An exact
best-response evaluation of this average computes
the remaining gap. This separates learning over
the discovered strategies from their coverage of
profitable full-game deviations.
If a Nash equilibrium of the final
restricted game admits no profitable deviation
under an exact full-game best-response audit, it
is also a Nash equilibrium of the full game. This
terminal certificate does not establish a
population-independent finite-time rate for VDO.
\end{remark}

\section{Discounted VDO}
\label{sec:dvdo}

\subsection{The Over-Concentration Problem}
\label{sec:overconc}

VDO's precision $\lambda = \eta \Tbar$ grows
linearly with the total number of steps. As
$\Tbar$ accumulates, the reinitialisation can
concentrate almost all mass on whichever strategy
performed best against the historical average.
Play can then remain concentrated between
best-response checks, and a newly revealed
response can take a substantial share on arrival.
Discounted VDO controls the strength of
this historical preference: it reconstructs past
weights at rate $\eta/\disc$, then learns from new
observations within each window at rate $\eta$.
The theorem below accounts for both; our
experiments examine how $\disc$ affects actual
play and subsequent discovery.

\subsection{Constant Discount}
\label{sec:constant_discount}

We introduce a constant discount $\disc > 0$ and
reinitialise with $q_{j+1}(s) \propto
\exp((\eta\Tbar_{j+1}/\disc) \cdot u_i(s,
\pibar^{[1:\Tbar_{j+1}]}_{-i}))$.
The effective precision is
$\eta\Tbar/\disc$, so smaller $\disc$ exploits
history more aggressively and larger $\disc$ is more
conservative. Here discounting rescales the
entire history uniformly; it changes the concentration of the historical weights rather than assigning greater weight to recent rounds.
The limiting cases are a hard historical response
($\disc\to0$), the exact virtual weights
($\disc=1$), and the uniform reset
($\disc\to\infty$). We use the convention
$1/\infty=0$, so $\disc=\infty$ yields
$q_{j+1}(s)=1/|\Pi^{(j+1)}|$, recovering ODO's
uniform reinitialisation rule under the selected
response protocol.

\subsection{Discounted Convergence Guarantees}
\label{sec:dvdo_bound}

Let $\mathcal I^{\gamma}_{i,T}$ be the
insertion cost~\eqref{eq:insertion_charge}
computed with historical weights $\exp(-\gamma
L_i^{\tau}(s))$, $\gamma=\eta/\disc$; both sums
use the same rate and history.

\begin{theorem}[dVDO: regret across population
expansions]
\label{thm:unified} Assume exact payoff feedback,
losses in $[-1,1]$, populations that only grow,
the historical reconstructions above, a fixed
learning rate $\eta>0$ and a fixed finite discount
$\disc\geq1$, and let $\gamma=\eta/\disc$. For
each player, dVDO's regret against its final
population obeys
\begin{equation}
 R_{i,T}\leq\frac{\disc\bigl(\ln k_{i,0}+\mathcal I^{\gamma}_{i,T}\bigr)}{\eta}
 +\frac{\eta T}{2}.
 \label{eq:dvdo_insertion}
\end{equation}
At $\disc=1$ this is Theorem~\ref{thm:vdo}.
Summing over the two players bounds
$\Exploit_{\Pi^{\mathrm{fin}}}(\pibar_1^T,\pibar_2^T)$;
full-game exploitability additionally includes
$\varepsilon_{\mathrm{disc}}$.
\end{theorem}

Appendix~\ref{app:proof_dvdo} proves Theorem~\ref{thm:unified} and gives two further regret estimates: one describes how reconstruction-window lengths affect regret, and the other describes convergence after the final population expansion.

\textbf{Historical weighting and the choice of
$\disc$.}\quad For a fixed linear budget $\ln k_{i,0}+\mathcal
I^{\gamma}_{i,T}\leq ck$ for both players at the
chosen rate $\gamma=\eta/\disc$, taking
$\eta=\sqrt{2\disc ck/T}$ gives
$\Exploit_{\Pi^{\mathrm{fin}}}\leq2\sqrt{2\disc
ck/T}$, hence $O(\sqrt{k/T})$ for fixed $\disc$.
For a common budget the displayed bound grows with
$\disc$, but $\disc$ also changes insertion costs
and subsequent play, so this does not rank
discount choices. In practice, fix $\disc$
in advance or select it on separate validation
games; the dVDO-best curves are hindsight
benchmarks.

\subsection{A Three-Axis View of Population
Expansion}
\label{sec:three_axes}

Once reinitialisation is made explicit, population
expansion decomposes naturally into three
independent design axes: (1)~\textbf{when to reveal} (BR frequency
$m(j)$): every step, periodic, adaptive, or only after subgame
convergence; (2)~\textbf{what to reveal} (BR target): exact best
response, approximate BR, regularised BR, or diverse/exploratory
policies; and (3)~\textbf{how to integrate} (reinitialisation): uniform
reset, warm start, VDO, or discounted VDO.
Existing work has primarily studied the first two
axes: PDO~\citep{tang2023regret} analyses
BR-frequency schedules with CFR inner solvers in
extensive-form games; the diversity-focused PSRO
literature~\citep{balduzzi2019open,
perez2021modelling, yao2023policy} studies what to
reveal. Our contribution targets the third axis.
Table~\ref{tab:unification}
(Appendix~\ref{app:additional}) situates DO, ODO,
VDO, and dVDO along the reinitialisation axis.
Variable-discount schedules ($\disc_j$ depending
on the window) lie outside the constant-$\disc$
bound; Appendix~\ref{app:variable_disc} gives a bound obtained by summing regret over reconstruction windows.

\section{Experiments}
\label{sec:experiments}

\subsection{Setup}

We evaluate \textbf{Gaussian matrix games} at sizes $25$, $50$ and $100$, and the OpenSpiel implementations of \textbf{Kuhn poker}, \textbf{Leduc poker} and \textbf{Goofspiel-4}~\citep{lanctot2019openspiel}. For each matrix size, ten seeds vary the initial strategies on one normalised Gaussian matrix. Additional games and full specifications appear in Appendices~\ref{app:more_experiments} and~\ref{app:results}.

\textbf{Baselines and protocol.}\quad
ODO~\citep{dinh2022online} ($\disc=\infty$), VDO
($\disc=1$), dVDO-best (a hindsight choice by final mean online exploitability; grids in Appendix~\ref{app:results}),
DO~\citep{mcmahan2003planning}, and an FP-style
hard-BR baseline (the $\disc\to0$ limit, maximising payoff against the global-average opponent at every round). All MWU-based variants use the
same BR target (the global time-averaged opponent)
and the same BR oracle, so differences between
them reflect only the reinitialisation rule;
our ODO baseline therefore uses this target rather
than the window-local average of published ODO,
and DO responds to the restricted-game
equilibrium. Each MWU player $i$ checks for a best response on its own countdown
$m(k_i)=\lfloor50\ln k_i\rfloor$, where $k_i$ is its population size,
restarting it after every check; only a player whose response adds a new
policy is reinitialised. The horizontal axes count response events, which
are distinct from the learning rounds $T$ of the theory; at each index,
the plotted metric averages the two players' corresponding evaluations.
We use a constant learning rate $\eta=1/11$
(Appendix~\ref{app:eta_robust} reports
sensitivity over $\eta\in[0.01,0.5]$), 10 seeds
(5 for Goofspiel-4 due to higher per-seed cost)
and 95\% confidence intervals.

\textbf{Approximate best-response oracles.}
In PSRO with reinforcement learning, the trained response may be suboptimal, and its training opponents may be drawn from a finite sample. We isolate each source with a
controlled, tabular approximation of the exact best response
$b_i = \mathrm{BR}(\sigma_{-i})$ to the opponent's BR target $\sigma_{-i}$.
(i)~\emph{$\varepsilon$-perturbed} (correct
  target, imperfect execution): the oracle
  computes $b_i$ and then, independently at each
  of player $i$'s infostates, replaces its action
  by a uniformly random legal action with
  probability $\varepsilon = 0.1$, so the result is still a pure strategy.
(ii)~\emph{Sampled target} (exact execution,
  wrong target): the oracle draws $N = 10$ i.i.d.\
  pure strategies from $\sigma_{-i}$ and returns the exact best response to their empirical mixture, mimicking a BR
  trained against sampled opponents. Both approximations recover the
exact oracle in the limit ($\varepsilon=0$,
$N\to\infty$) and draw from their own random
stream, seeded per run. We run both oracles on
Kuhn, Leduc and Goofspiel-4, and on the three
further games of Appendix~\ref{app:results}.

\textbf{Metrics.} We report \emph{online exploitability} of the time-averaged profile and \emph{restricted-NE exploitability (RNE)}: the full-game exploitability of a restricted-game Nash equilibrium computed by linear programming~\citep[Section~4.2]{wang2022evaluating}. Since the MWU-based variants share the BR target and oracle, RNE comparisons measure how reinitialisation affects subsequent discovery. Both metrics are reported as half NashConv on
the normalised game: native-payoff NashConv divided by $2M$, where $M$ is
the maximum absolute native game payoff. The PSRO results
(Figures~\ref{fig:main_results}--\ref{fig:sampled},
Figure~\ref{fig:traj_imperfect_info} and Appendix~\ref{app:results}) were
trained with native payoffs; Figures~\ref{fig:eta_robust}
and~\ref{fig:traj_matrix}--\ref{fig:long_horizon} use payoffs normalised
before learning but the same response schedules and resets, with the
protocol in Appendix~\ref{app:more_experiments}.

\textbf{Compute and reproducibility.} Experiments use a single CPU core, no GPU, and under 100 CPU-hours in total. Extensive-form games and exact BR oracles use OpenSpiel (Apache~2.0). Code, numerical results and reproduction instructions are available at \url{https://github.com/ruilan-wang/vdo-reproducibility}.

\subsection{Main Results}
\label{sec:main_results}

\begin{figure}[t]
\centering
\begin{subfigure}[b]{0.32\textwidth}
\includegraphics[width=\textwidth]{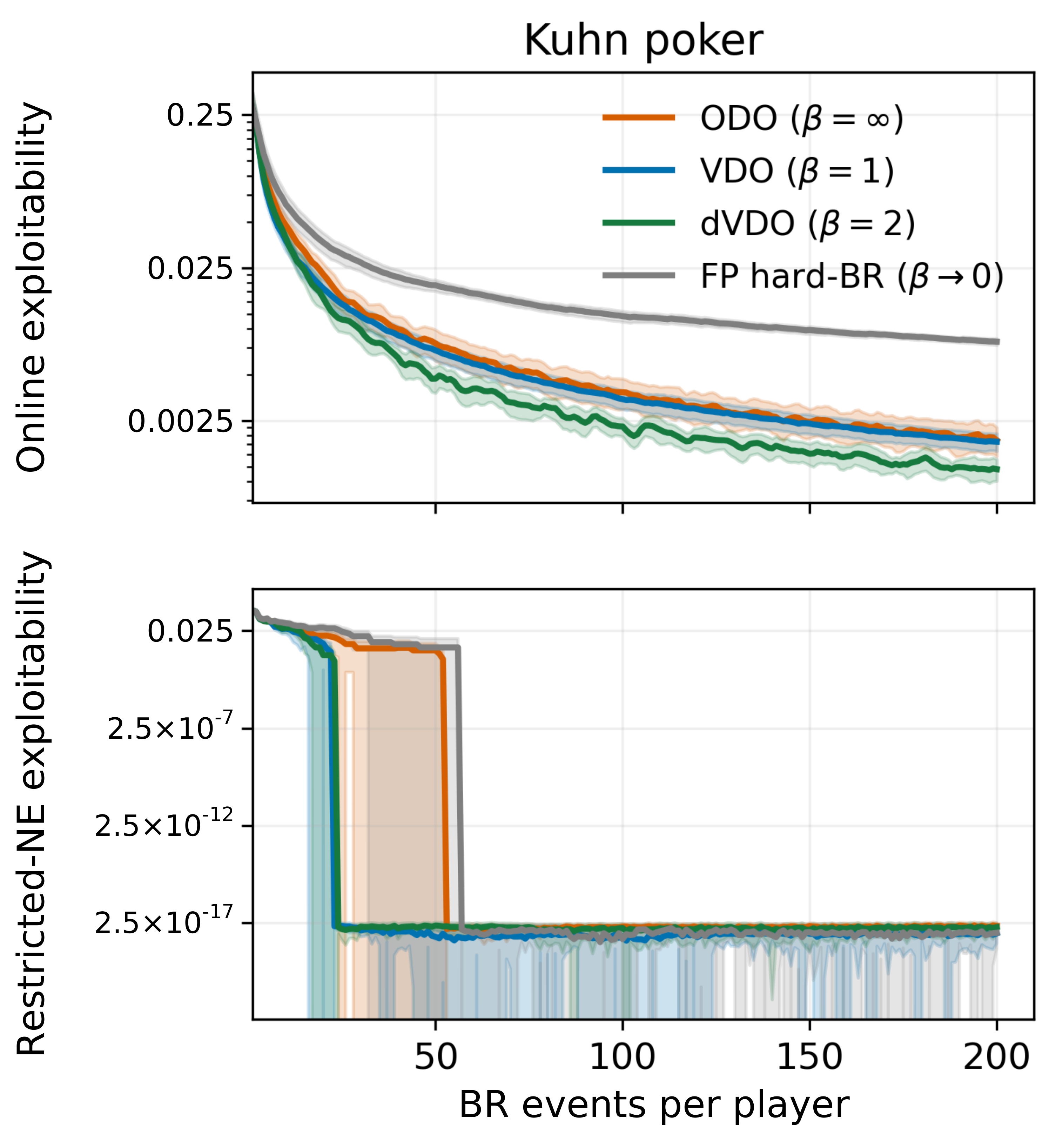}
\caption{Kuhn poker.}
\end{subfigure}\hfill
\begin{subfigure}[b]{0.32\textwidth}
\includegraphics[width=\textwidth]{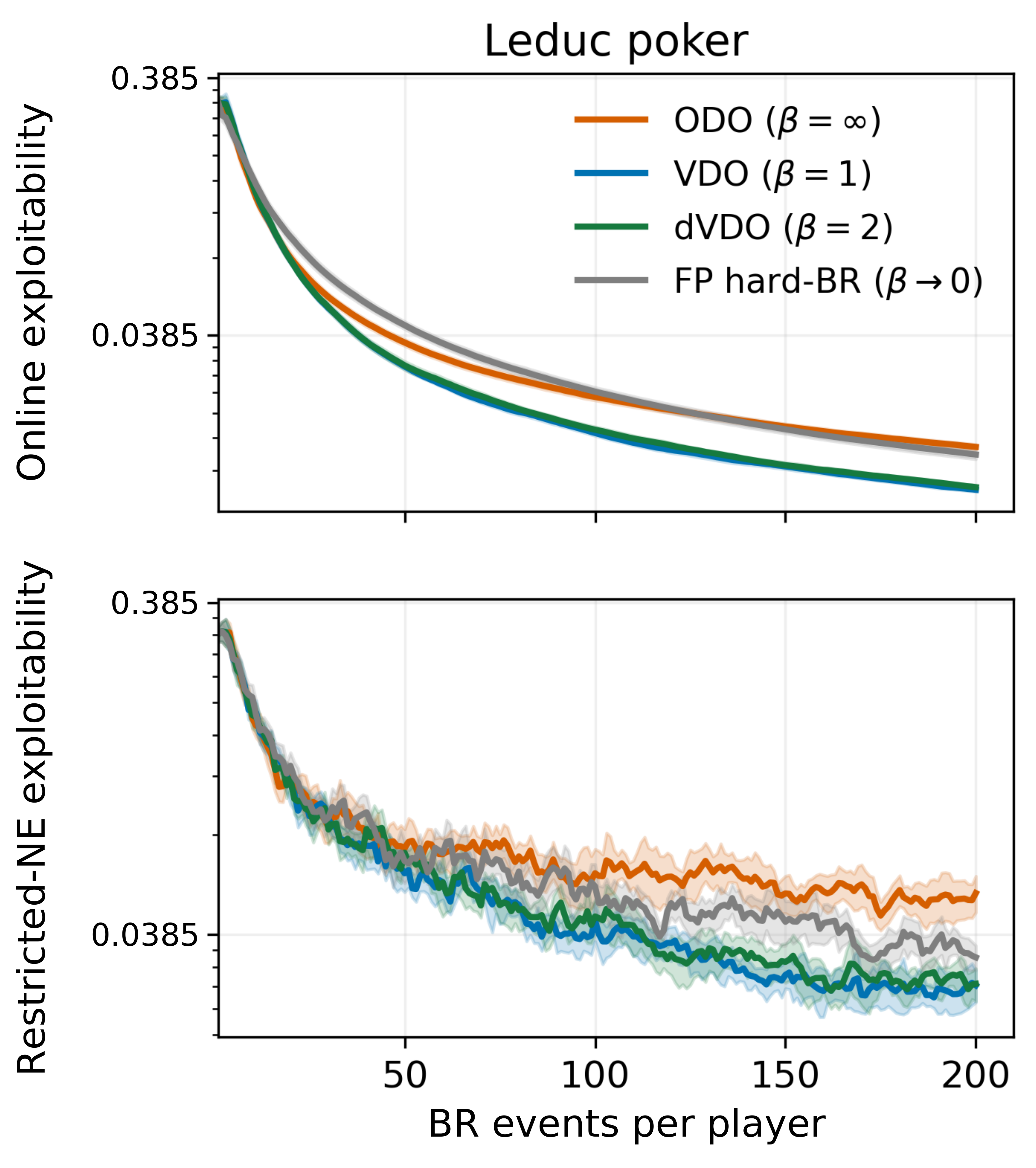}
\caption{Leduc poker.}
\end{subfigure}\hfill
\begin{subfigure}[b]{0.32\textwidth}
\includegraphics[width=\textwidth]{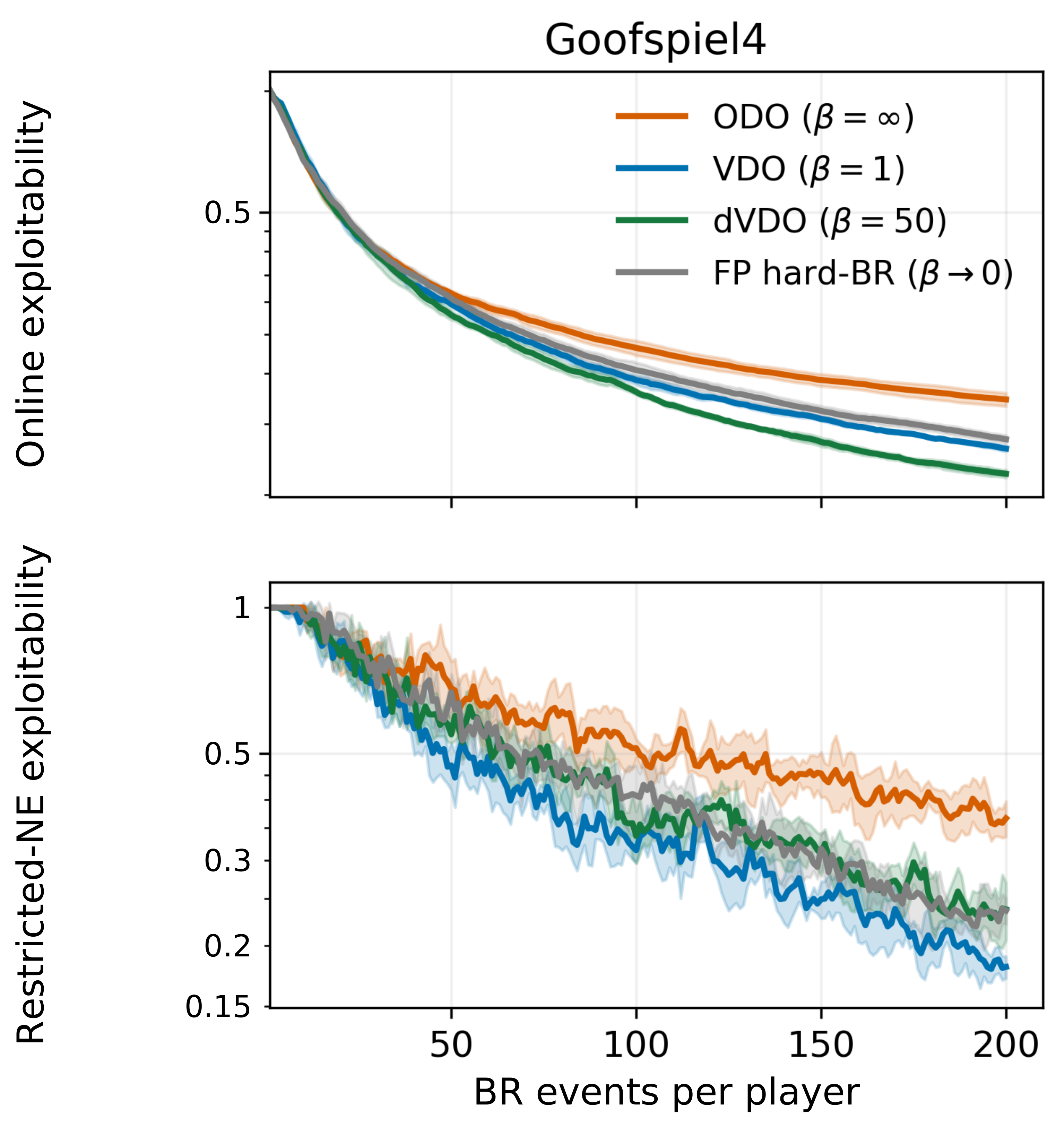}
\caption{Goofspiel-4.}
\end{subfigure}
\caption{Online (top) and
restricted-NE (bottom, RNE) exploitability
versus BR queries on three imperfect-information
games. ODO is the uniform-reset control with the
historical BR target; VDO uses $\disc=1$;
dVDO-best is the per-game hindsight choice. Full
$\disc$ sweeps are in
Appendix~\ref{app:full_traj}; matrix-game
baselines and the long-horizon run are in
Appendices~\ref{app:matrix_baselines}
and~\ref{app:long_horizon}.}
\label{fig:main_results}
\end{figure}

Figure~\ref{fig:main_results} compares
online and restricted-NE (RNE)
exploitability. On \textbf{Kuhn} (small $k$),
dVDO has the lowest mean online exploitability: $\disc=2$ (shown) and $\disc=5$ both give $0.00120$ to three significant figures. VDO ($\disc=1$) and ODO are close behind, and FP-hardBR is the worst. VDO and dVDO reach an RNE of $\approx 0$ after about 20 BR queries, against about 55 for ODO and FP-hardBR.
On \textbf{Leduc} ($k\approx 130$), VDO and dVDO ($\disc=2$) are close on both metrics in Figure~\ref{fig:main_results}(b). At the endpoint, VDO has the lowest mean online exploitability and the lowest mean RNE among the MWU-based methods. On \textbf{Goofspiel-4}, dVDO ($\disc=50$) has the lowest mean online exploitability in Figure~\ref{fig:main_results}(c), whereas VDO ($\disc=1$) has the lowest mean RNE.
The preferred discount differs across games
and between the two metrics
(Appendix~\ref{app:full_ushape}); at response event 200 per player, $\disc=5$ gives the lowest mean online exploitability on each tested matrix instance.

\begin{figure}[t]
\centering
\begin{subfigure}[b]{0.32\textwidth}
\includegraphics[width=\textwidth]{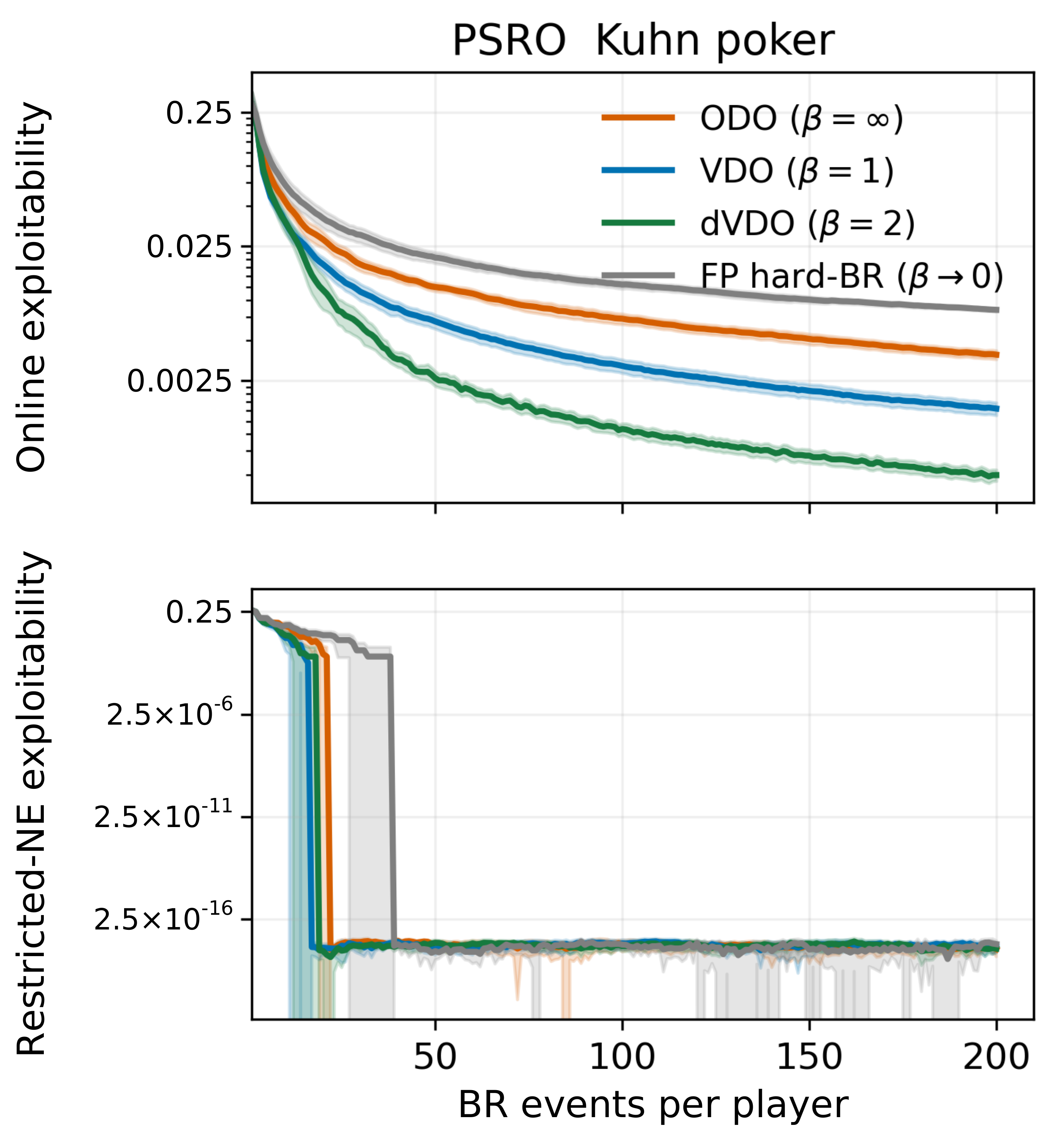}
\caption{Kuhn poker.}
\end{subfigure}\hfill
\begin{subfigure}[b]{0.32\textwidth}
\includegraphics[width=\textwidth]{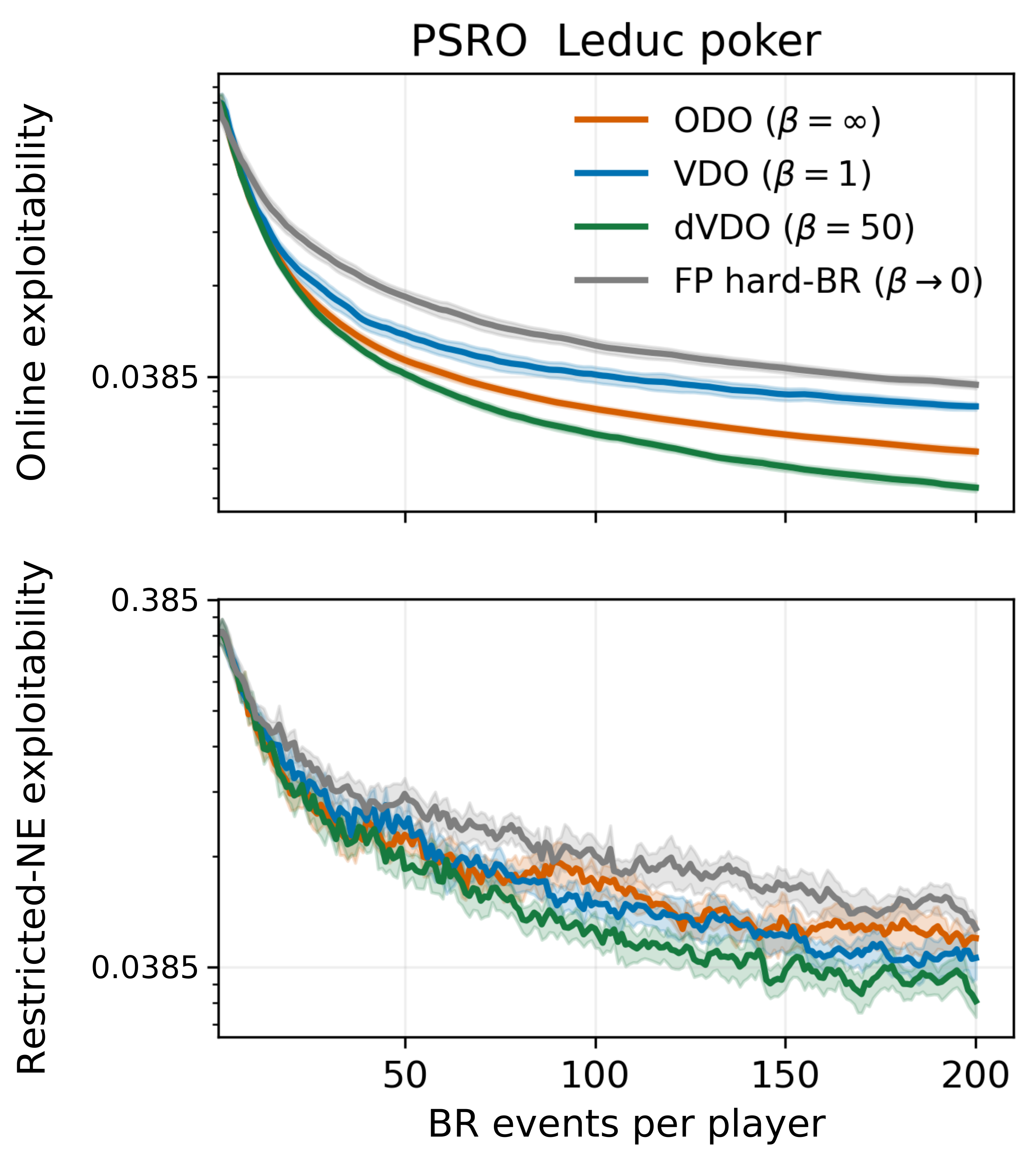}
\caption{Leduc poker.}
\end{subfigure}\hfill
\begin{subfigure}[b]{0.32\textwidth}
\includegraphics[width=\textwidth]{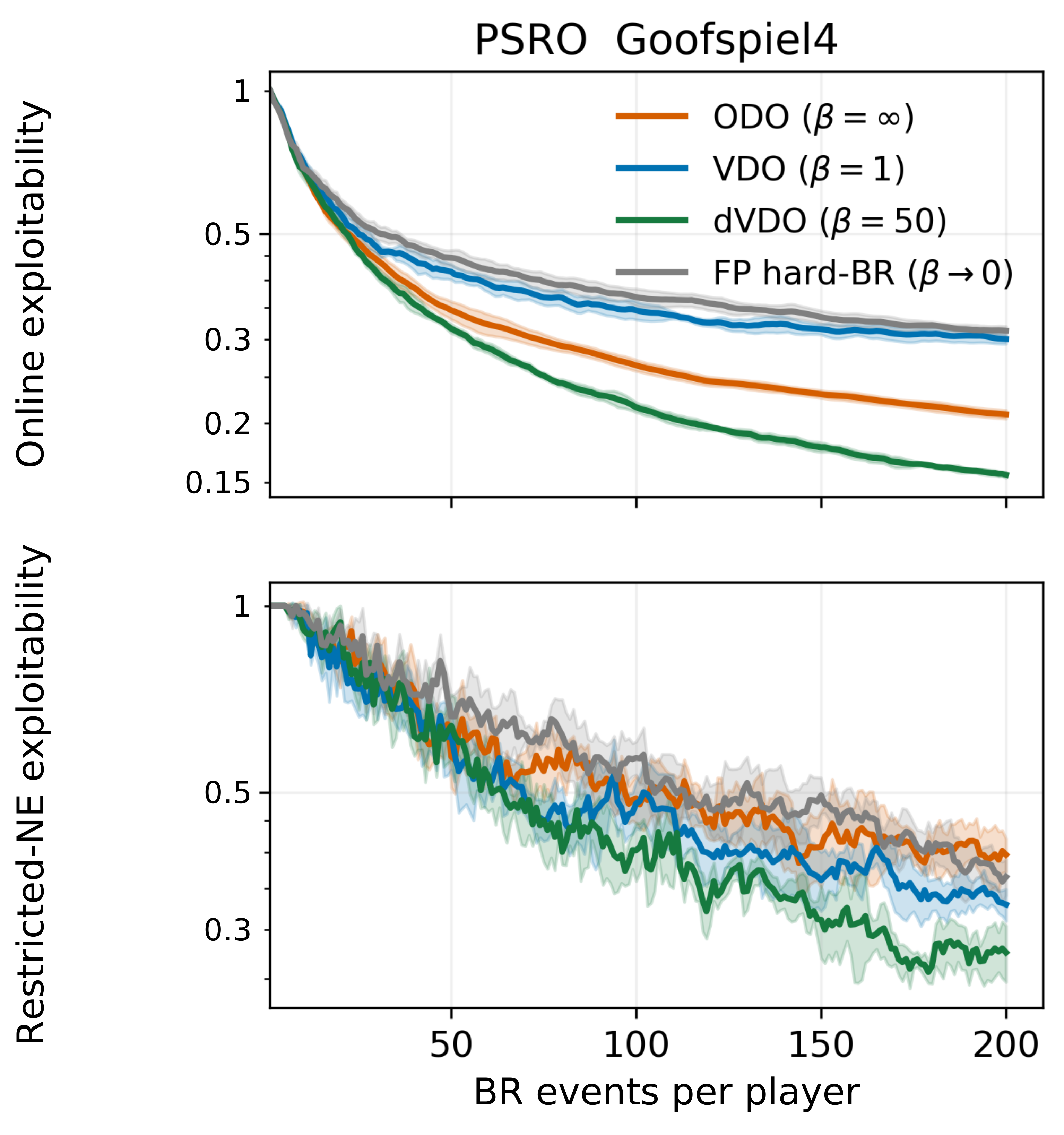}
\caption{Goofspiel-4.}
\end{subfigure}
\caption{$\varepsilon$-perturbed best
responses ($\varepsilon=0.1$). Online (top) and
restricted-NE (bottom, RNE) exploitability versus
BR queries, with the same baselines as
Figure~\ref{fig:main_results}. Of the tested discounts, $\disc=50$ gives the lowest mean online exploitability on Leduc and Goofspiel-4. VDO has higher online exploitability than ODO on both games.
dVDO-best has the lowest online exploitability and
the lowest or tied-lowest RNE on every game.}
\label{fig:eps}
\end{figure}

\begin{figure}[t]
\centering
\begin{subfigure}[b]{0.32\textwidth}
\includegraphics[width=\textwidth]{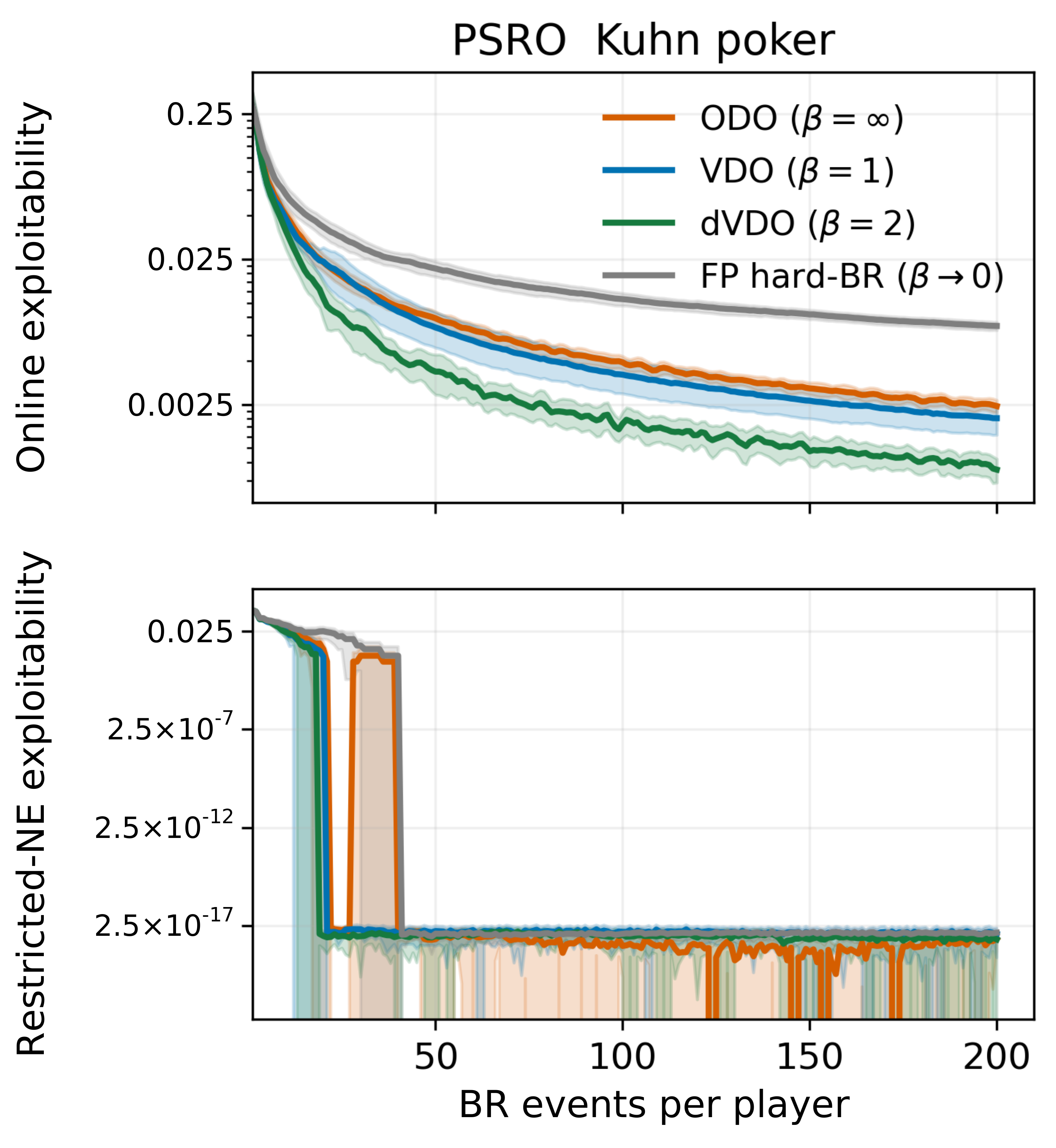}
\caption{Kuhn poker.}
\end{subfigure}\hfill
\begin{subfigure}[b]{0.32\textwidth}
\includegraphics[width=\textwidth]{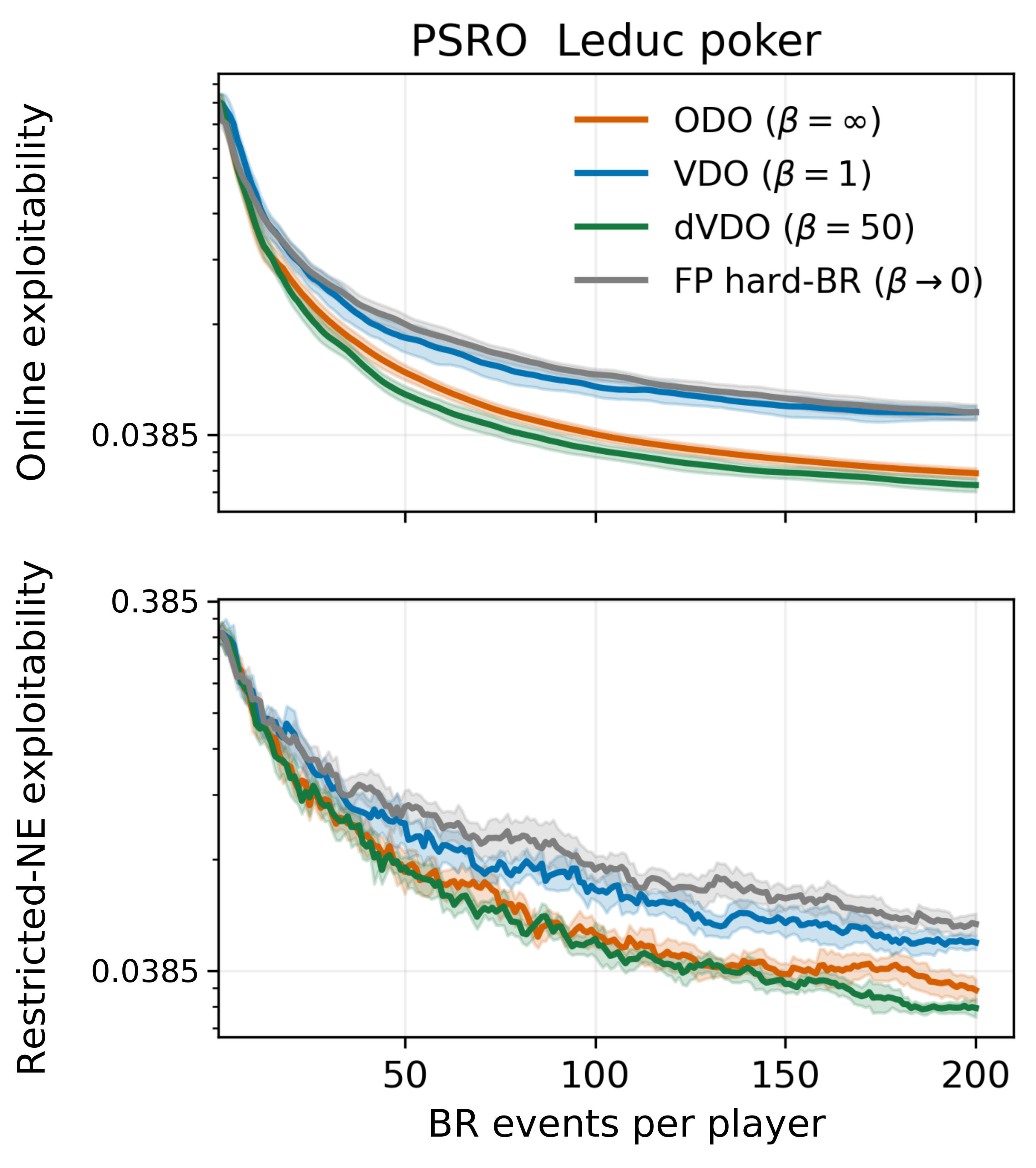}
\caption{Leduc poker.}
\end{subfigure}\hfill
\begin{subfigure}[b]{0.32\textwidth}
\includegraphics[width=\textwidth]{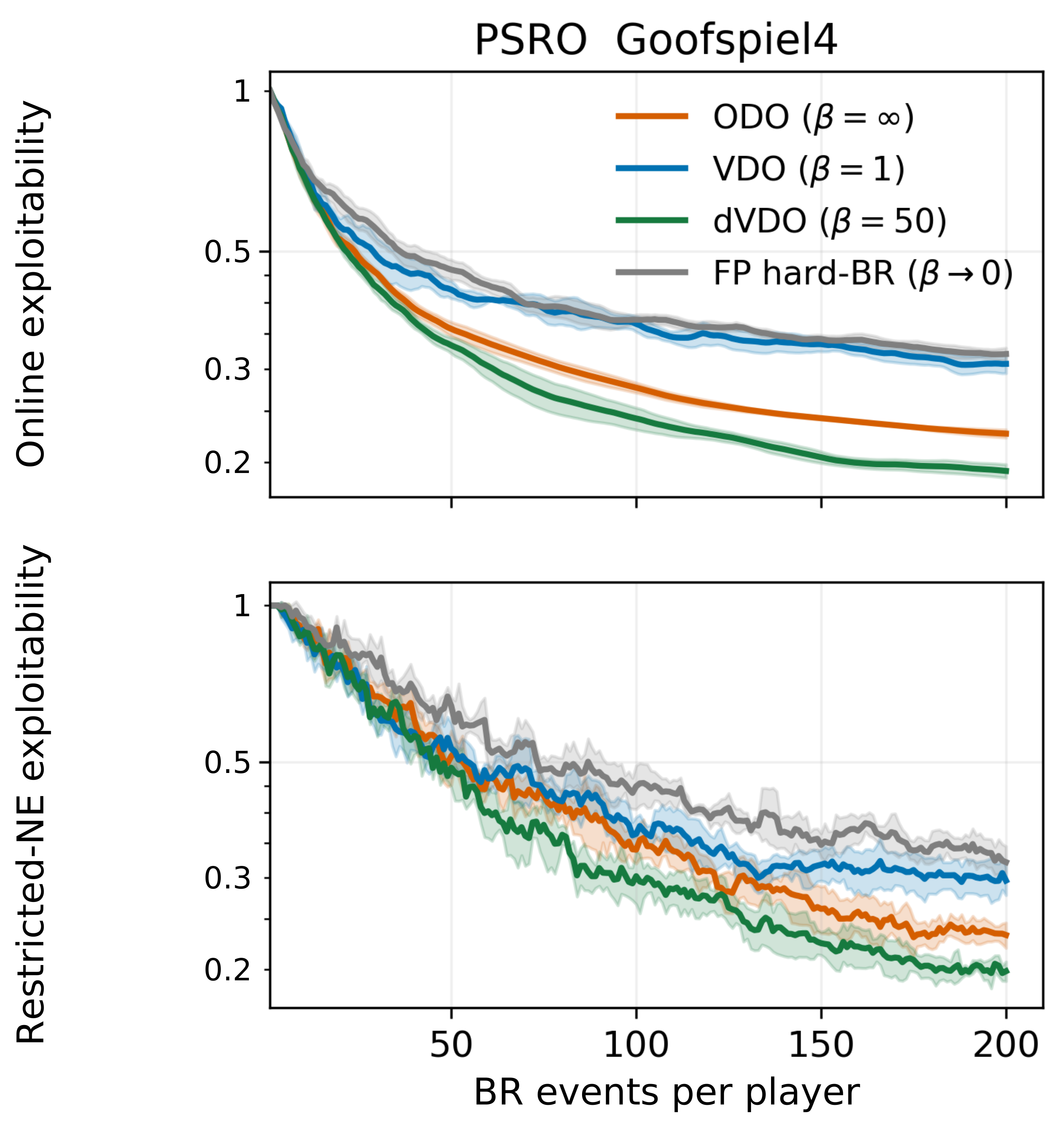}
\caption{Goofspiel-4.}
\end{subfigure}
\caption{Sampled-target best responses
($N=10$ opponent samples). Online (top) and
restricted-NE (bottom, RNE) exploitability versus
BR queries, with the same baselines as
Figure~\ref{fig:main_results}. dVDO-best is lowest
online and lowest or tied-lowest on RNE on all
three games; on Leduc and Goofspiel-4, $\disc=50$ is the best of the tested discounts online, and VDO has higher online exploitability than ODO.}
\label{fig:sampled}
\end{figure}

Under both approximate-response oracles (Figures~\ref{fig:eps} and~\ref{fig:sampled}), the lowest mean online exploitability on Leduc is attained at $\disc=50$ among the two discounts tested, $2$ and $50$. Goofspiel-4 also favours $\disc=50$, while Kuhn favours $\disc=2$. On Leduc and Goofspiel-4, VDO has higher online exploitability than ODO under both oracle models. Among the methods plotted, dVDO-best has the lowest mean online exploitability on all three games and the lowest or tied-lowest mean RNE at the reported precision. These results are consistent with the over-concentration argument of Section~\ref{sec:overconc}: an undiscounted all-history prior over-weights early, noisy responses, so stronger discounting is advisable with approximate best responses.

\subsection{Long-horizon behaviour and discount sensitivity}
\label{sec:long_horizon}

Figure~\ref{fig:long_horizon} (Appendix~\ref{app:long_horizon}) follows one normalised $50\times50$ matrix for 3,000 response events per player, using ten seeds. Among the methods displayed, dVDO with $\disc=10$ has the lowest mean online exploitability at event 3,000. The fixed-precision control reconstructs at $\lambda=10$ whenever that player adds a strategy; the other methods use their specified historical precision.

\section{Related Work}
\label{sec:related}

\textbf{Double Oracle family.}\quad
DO~\citep{mcmahan2003planning} and
ADO~\citep{mcaleer2022anytime} solve restricted
games exactly but lack rate bounds.
ODO~\citep{dinh2022online} introduced MWU-based
solving with $O(\sqrt{k \ln k / T})$.
RMDO/PDO~\citep{tang2023regret} extend regret
minimisation to extensive-form games with an
explicit per-window reset step, defaulting to
uniform reinitialisation;
XDO~\citep{mcaleer2021xdo} extends DO to
extensive-form via action-level expansion.

\textbf{PSRO, regularisation, and discounted
regret.}\quad PSRO~\citep{lanctot2017unified}
generalises DO with RL best responses; APSRO uses
an exploitability-minimising restricted
distribution via no-regret
updates~\citep{mcaleer2022anytime}; SP-PSRO adds
approximately optimal stochastic
policies~\citep{mcaleer2024toward}. Prior
warm-start work in PSRO targets BR
training~\citep[e.g.,][]{mcaleer2020pipeline}, not
meta-strategy reinitialisation. PSRO
overfitting~\citep{bighashdel2024psro} has been
addressed via meta-solver
regularisation~\citep{lanctot2017unified,
wang2023regularization} and history-aware oracle
training~\citep{wright2019history}; dVDO addresses
related symptoms at the reinitialisation
layer. Unlike discounted
regret~\citep[Ch.~2]{cesabianchi2006prediction}
and discounted CFR~\citep{brown2019solving}, which
down-weight older rounds, it rescales the whole
history uniformly.

\section{Discussion and Limitations}
\label{sec:discussion}

\paragraph{What VDO does.}
The bilinear identity (Lemma~\ref{lem:bilinear}) gives a closed-form
reinitialisation: VDO plays the distribution of a virtual MWU learner
conditioned on the strategies currently available, and
Theorem~\ref{thm:vdo} charges each expansion by its historical weight,
yielding $O(\sqrt{k/T})$ discovered-population exploitability under a
linear insertion-cost budget. Discounted VDO tempers that history; its
benefit at moderate $\disc$ is an empirical, finite-horizon effect. The
guarantee concerns integration over the discovered population;
improvements in later discovery are supported by RNE comparisons but not
guaranteed by the regret bound.

\paragraph{Scope and limitations.}
Our guarantees assume two-player zero-sum payoffs, exact payoff
feedback, retained strategies and a fixed learning rate; the rates also
need their insertion budgets, and full-game quality depends on the
discovery gap (Remark~\ref{rem:scope}). The rule applies to any finite
policy population, including learned best responses, but noisy payoff
estimates are not covered. The rule does not provide enough information for researchers to set discounted hyperparameters before running experiments; some intuition based on noise can be used to define the discount rate, but empirical evaluations are needed. Our experiments are tabular, dVDO-best is a
hindsight benchmark, and the Gaussian results use one matrix per size.

\begin{ack}
The authors received no third-party funding or support for this work.
The authors declare no competing interests.
\end{ack}

\bibliographystyle{plainnat}

\clearpage
\appendix

\section{Proofs for VDO}
\label{app:proof_vdo}

\subsection{Insertion-sensitive regret:
Theorem~\ref{thm:vdo}}
\label{app:insertion}

Fix one player and suppress its index. Let $A_t$
be its active population before round~$t$, let
$A^{\mathrm{fin}}$ be the population after all
round-$T$ insertions, and set $K_0=|A_1|$ and $k_i=|A^{\mathrm{fin}}|$ for the player under analysis. Define
\[
 L_t(s)=\sum_{r=1}^t\ell^r(s),\quad L_0(s)=0,\quad
 w_t(s)=e^{-\eta L_{t-1}(s)}\ (s\in A_t),\quad
 W_t=\sum_{s\in A_t}w_t(s).
\]
Lemma~\ref{lem:bilinear} reconstructs this same weight on arrival. For $\disc=1$, reinitialising an
incumbent reproduces its existing unnormalised
weight, so $p_t(s)=w_t(s)/W_t$ throughout the run.
This is the weight consistency behind the
conditioning
identity~\eqref{eq:virtual_conditioning}. All
losses refer to the actual opponent sequence,
rather than a counterfactual self-play trajectory.

\paragraph{Ordinary learning steps.} Let
$\widetilde W_{t+1}=\sum_{s\in A_t}e^{-\eta
L_t(s)}$ denote the potential after the round-$t$
loss update and before any insertion. Then
\[
 \frac{\widetilde W_{t+1}}{W_t}=\sum_{s\in A_t}p_t(s)e^{-\eta\ell^t(s)}.
\]
Hoeffding's lemma, with loss range two, gives
\begin{equation}
 \ln\widetilde W_{t+1}-\ln W_t\leq-\eta\langle p_t,\ell^t\rangle
 +\frac{\eta^2}{2}.
 \label{eq:A_step}
\end{equation}

\paragraph{Population expansions.} An expansion
after the loss update adds the reconstructed
weights of its new strategies. If $U_e$ is their
total weight, then $W_e^+=W_e^-+U_e$. Write
\[
 \Delta_e=\ln\frac{W_e^+}{W_e^-},\qquad \mathcal I_T=\sum_e\Delta_e .
\]
Repeated queries that add no strategy have zero
insertion charge. Any strategies inserted after
the last played round are included both in
$A^{\mathrm{fin}}$ and in $\mathcal I_T$.

\paragraph{Telescoping and comparison.}
Adding~\eqref{eq:A_step} and all expansion jumps
gives
\[
 \ln W_{T+1}\leq\ln K_0-\eta\sum_{t=1}^T\langle p_t,\ell^t\rangle
 +\frac{\eta^2T}{2}+\mathcal I_T .
\]
For every $s\in A^{\mathrm{fin}}$, the final
potential contains its full historical weight:
$W_{T+1}\geq e^{-\eta L_T(s)}$. Therefore
\[
 \sum_{t=1}^T\langle p_t,\ell^t\rangle-L_T(s)\leq
 \frac{\ln K_0+\mathcal I_T}{\eta}+\frac{\eta T}{2}.
\]
Taking the best comparator in $A^{\mathrm{fin}}$
proves the per-player bound. In a zero-sum game
the realised payoff terms cancel when the two
regrets are added, giving
\[
 \Exploit_{\Pi^{\mathrm{fin}}}(\pibar_1^T,\pibar_2^T)
 =\frac{R_{1,T}+R_{2,T}}{T}.
\]
This proves Theorem~\ref{thm:vdo}. For
Corollary~\ref{cor:budget}, substitute
$C_{i,T}\leq ck$ and $\eta=\sqrt{2ck/T}$ into
\eqref{eq:insertion_gap}. The resulting
$O(\sqrt{k/T})$ rate holds when the budget is
satisfied at that learning rate.

\paragraph{Interpreting insertion budgets.} For
single-strategy insertions, suppose the newcomers'
weight share at each insertion satisfies $q_e\leq
1-\delta$ for a constant $\delta>0$. Then
$\Delta_e=-\ln(1-q_e)\leq\ln(1/\delta)$. With at
most $k-K_0$ insertions, $\mathcal
I_T\leq(k-K_0)\ln(1/\delta)$ and $\ln K_0+\mathcal
I_T\leq ck$ for a constant $c$ depending only on
$\delta$. This condition concerns the realised
historical weights.

Finite-horizon empirical measurements support the plausibility of the linear insertion-budget condition (Appendix~\ref{app:insertion_experiments}).

\subsection{The virtual learner and post-discovery
convergence}
\label{app:postdiscovery}

\begin{corollary}[Post-discovery convergence]
\label{cor:postdiscovery} If the final expansion
occurs after $\tau$ observed rounds, exact VDO
satisfies
\[
 \Exploit_{\Pi^{\mathrm{fin}}}(\pibar_1^T,\pibar_2^T)\leq
 \frac{\ln k_1+\ln k_2}{\eta T}+\eta+\frac{4\tau}{T}.
\]
With the fixed-rate calibration $\eta=\sqrt{2k/T}$, this gives the bound $2\sqrt{2k/T}+4\tau/T$.
\end{corollary}

\begin{proof} Define a virtual MWU learner on
$A^{\mathrm{fin}}$ from the beginning, processing
the same realised losses. Its distribution before
round~$t$ is
\[
 v_t(s)=\frac{e^{-\eta L_{t-1}(s)}}{\sum_{a\in A^{\mathrm{fin}}}
 e^{-\eta L_{t-1}(a)}},\qquad Z_t=v_t(A_t).
\]
Equation~\eqref{eq:virtual_conditioning} gives
$p_t=v_t(\cdot\mid A_t)$, where the virtual
learner processes the same actual opponent
sequence. Standard MWU on the final population of
size $k_i$ implies
\[
 \sum_{t=1}^T\langle v_t,\ell^t\rangle-L_T(s)\leq\frac{\ln k_i}{\eta}
 +\frac{\eta T}{2}\qquad(s\in A^{\mathrm{fin}}).
\]
For $Z_t<1$, write $v_t=Z_tp_t+(1-Z_t)h_t$, where
$h_t$ is a distribution on the unrevealed
strategies. Since losses lie in $[-1,1]$,
\[
 \langle p_t-v_t,\ell^t\rangle=(1-Z_t)\langle p_t-h_t,\ell^t\rangle
 \leq2(1-Z_t).
\]
The same bound is zero when $Z_t=1$. Thus
\begin{equation}
 R_T(A^{\mathrm{fin}})\leq\frac{\ln k_i}{\eta}+\frac{\eta T}{2}
 +2\sum_{t=1}^T(1-Z_t).
 \label{eq:A_hidden}
\end{equation}
If the last expansion is after $\tau$ observed
rounds, all final strategies are available for
rounds $t>\tau$, and $\sum_t(1-Z_t)\leq\tau$.
Summing the two per-player estimates proves the
stated bound. The fixed-rate
inequality~\eqref{eq:A_hidden} applies at every
horizon.
\end{proof}

For uniform-reset ODO, the prefix contributes at
most $2\tau$ per player and the final
uniform-initialised MWU phase contributes at most
$\ln k_i/\eta+\eta(T-\tau)/2$. Hence it also admits
the coarse post-discovery form.
Theorem~\ref{thm:vdo} retains the individual
expansion charges instead of replacing the whole
prefix by $2\tau$.

\subsection{From discovered-population to
full-game exploitability}
\label{app:discovery_gap}

For $i=1,2$, define
\[
 \delta_{i,T}=\max_{s\in S_i}u_i(s,\pibar^T_{-i})
 -\max_{s\in\Pi_i^{\mathrm{fin}}}u_i(s,\pibar^T_{-i})\geq0 .
\]
Then
$\varepsilon_{\mathrm{disc}}=\delta_{1,T}+\delta_{2,T}$
and
$\Exploit_S(\pibar^T)=\Exploit_{\Pi^{\mathrm{fin}}}(\pibar^T)
+\varepsilon_{\mathrm{disc}}$. An exact response
to each reported opponent average evaluates this
gap. If the two response values are certified to
within $\epsilon_1$ and $\epsilon_2$, their sum
estimates full-game exploitability with an
additive error at most $\epsilon_1+\epsilon_2$.
Inclusion of the support of some full-game
equilibrium does not by itself replace this
output-specific certificate.

\section{Proofs for discounted VDO}
\label{app:proof_dvdo}

\subsection{Proof of Theorem~\ref{thm:unified}}
Fix a player and suppress its index; write
$\gamma=\eta/\disc\leq\eta$. Partition the rounds
into this player's successive reset windows.
Window~$j$ starts after $\tau_j$ observed rounds
from $q_j(s)=e^{-\gamma
L_{\tau_j}(s)}/W_j^{\mathrm{start}}$ on its active
population $A_j$, where
$W_j^{\mathrm{start}}=\sum_{s\in A_j}e^{-\gamma
L_{\tau_j}(s)}$, and has length $n_j$ and block
losses $G_j$. Let $W_j^{\mathrm{end}}=\sum_{s\in
A_j}e^{-\gamma L_{\tau_{j+1}}(s)}$ with
$\tau_{J+1}=T$, and
$W_{J+1}^{\mathrm{start}}=\sum_{s\in
A^{\mathrm{fin}}}e^{-\gamma L_T(s)}$; then
$\mathcal
I^\gamma_T=\sum_j\ln(W_{j+1}^{\mathrm{start}}/W_j^{\mathrm{end}})$,
whose terms vanish at resets where this player's
population does not grow. The potential argument
of Appendix~\ref{app:insertion}, applied to MWU at
rate $\eta$ inside the window, gives $\sum_{t\in
T_j}\langle
p_t,\ell^t\rangle\leq-f_j(\eta)/\eta+\eta n_j/2$,
where $f_j(a)=\ln\sum_{s\in
A_j}q_j(s)e^{-aG_j(s)}$. As a
log-moment-generating function, $f_j$ is convex
with $f_j(0)=0$, so
$f_j(\gamma)\leq(\gamma/\eta)f_j(\eta)$, that is,
$-f_j(\eta)/\eta\leq-f_j(\gamma)/\gamma$. Since
$q_j$ is proportional to the rate-$\gamma$
historical weights, $f_j(\gamma)=\ln
W_j^{\mathrm{end}}-\ln W_j^{\mathrm{start}}$.
Summing over windows and using
$W_1^{\mathrm{start}}=K_0$,
\[
 \sum_{t=1}^T\langle p_t,\ell^t\rangle\leq
 \frac{\ln K_0+\mathcal I^\gamma_T-\ln W_{J+1}^{\mathrm{start}}}{\gamma}
 +\frac{\eta T}{2}.
\]
For every $s\in A^{\mathrm{fin}}$,
$W_{J+1}^{\mathrm{start}}\geq e^{-\gamma L_T(s)}$,
which gives~\eqref{eq:dvdo_insertion}.

\subsection{Regret in terms of window lengths} Use the
same player's reconstruction windows, suppressing
its index on losses and strategies. Let $k_i=|A^{\mathrm{fin}}|$,
$L_j(s)=\sum_{t\in T_j}\ell^t(s)$, and
$\dbar=T^{-1}\sum_jn_j^2$, the mean length of the window containing a uniformly selected learning round. Define a virtual distribution
\[
 b_j(s)\propto\exp\Bigl(-\frac{\eta}{\disc}\sum_{r<j}L_r(s)\Bigr)
 \qquad(s\in A^{\mathrm{fin}})
\]
using the actual historical losses on the full final population. The actual initial distribution
in window $j$ is $q_j=b_j(\cdot\mid A_j)$, with
$Z^B_j=b_j(A_j)$ and $H^B=\sum_jn_j(1-Z^B_j)$. Here $Z^B_j$ is the virtual probability mass of the strategies already available in window $j$, and $H^B$ sums the unavailable mass, weighted by window length.

\begin{proposition}[Regret in terms of window lengths]
\label{prop:block} Under the assumptions of
Theorem~\ref{thm:unified},
\[
 \frac{R_T}{T}\leq\frac{\disc\ln k_i}{\eta T}+\frac{\eta\dbar}{2\disc}
 +\frac{\eta}{2}+\frac{2H^B}{T}.
\]
\end{proposition}

\begin{proof} The sequence $b_j$ follows exponential weights with window losses $L_j$ and learning rate $\eta/\disc$. The range of $L_j/\disc$ is at most
$2n_j/\disc$. Applying the same
exponential-potential inequality as in
Appendix~\ref{app:insertion} with these block
ranges and multiplying by $\disc$ gives, for any
$s\in A^{\mathrm{fin}}$,
\[
 \sum_j\langle b_j,L_j\rangle-\sum_jL_j(s)\leq\frac{\disc\ln k_i}{\eta}
 +\frac{\eta}{2\disc}\sum_jn_j^2 .
\]
Conditioning $b_j$ onto $A_j$ changes expected
block loss by at most $2n_j(1-Z^B_j)$, because the
block-loss range is at most $2n_j$, so
$\sum_j\langle q_j-b_j,L_j\rangle\leq2H^B$. Within
window $j$, ordinary MWU starts at $q_j$ and uses
rate $\eta$; the non-uniform-prior regret
inequality with comparator distribution $q_j$
gives $\sum_j(\sum_{t\in T_j}\langle
p_t,\ell^t\rangle-\langle q_j,L_j\rangle) \leq\eta
T/2$. Adding the three comparisons and dividing by
$T$ proves the claim.
\end{proof}

\subsection{Regret after the final population expansion}

\begin{proposition}[Regret after the final population expansion]
\label{prop:dvdo_postdiscovery} Under the
assumptions of Theorem~\ref{thm:unified}, if this
player's final reset occurs after $\tau$ observed
rounds, then
\[
 \frac{R_T}{T}\leq\frac{\ln k_i}{\eta T}+\frac{\eta}{2}
 +\Bigl(2+\frac{2}{\disc}\Bigr)\frac{\tau}{T}.
\]
\end{proposition}

\begin{proof} The prior on the final population
for round $\tau+1$ is $q_\tau(s)\propto
e^{-(\eta/\disc)L_\tau(s)}$. For every $s\in
A^{\mathrm{fin}}$,
$L_\tau(s)-\min_aL_\tau(a)\leq2\tau$, so
$\ln(1/q_\tau(s))\leq\ln k_i+2\eta\tau/\disc$. The
first $\tau$ rounds contribute at most $2\tau$
regret. Applying~\eqref{eq:nonuniform_mwu} to the
remaining $T-\tau$ rounds gives
\[
 R_T(s)\leq2\tau+\frac{\ln(1/q_\tau(s))}{\eta}+\frac{\eta(T-\tau)}{2}
 \leq\frac{\ln k_i}{\eta}+\frac{\eta T}{2}+\Bigl(2+\frac2\disc\Bigr)\tau .
\]
A uniform final reset has $q_\tau(s)=1/k_i$, giving
the same expression with $1/\disc=0$.
\end{proof}

\section{Learning Rate Discussion}
\label{app:eta}

The bilinear identity concerns losses and holds
independently of the learning rate. The displayed
VDO and dVDO bounds analyse one fixed $\eta$ on
each full run. Their rate corollaries use the
stated horizon and insertion-budget calibration using the final population size $k$.
A learning rate calibrated to $k$ can be specified in advance when that size is known.
Since the insertion cost and discovery trajectory
can depend on $\eta$, a rate selected using one
realised trajectory cannot be substituted into
another trajectory without checking the
corresponding conditions.

The experiments use the fixed value $\eta=1/11$.
This is an empirical choice assessed by the
sensitivity study, rather than an implementation
of the horizon-calibrated rate corollaries. With
time-varying $\eta_t$, retrospective weights can
instead be reconstructed from the weighted
opponent average
$\sum_t\eta_t\pi^t_{-i}/\sum_t\eta_t$ and the
total weight $\sum_t\eta_t$. An anytime or
adaptive-rate guarantee requires analysis of that
specified schedule; it is separate from the
fixed-rate results proved here.

\section{BR Computation Frequency}
\label{app:br_freq}

Algorithm~\ref{alg:vdo} allows a configurable
per-player checking schedule $m(k)$. Each player has its own countdown and reconstruction windows. A check need not
discover a new strategy, so a reset window may
contain several checks. Thus the checking
interval, window length and final insertion time
are distinct. The insertion-sensitive bounds
account for population growth through the
insertion costs; window lengths enter only the
estimate in
Proposition~\ref{prop:block}.

BR frequency and historical initialisation are
separate design choices. A checking schedule alone
does not bound the time of the final discovery.
PDO~\citep{tang2023regret} studies response
frequency as a separate axis in extensive-form
regret minimisation.

\section{The FP-style Hard-BR Limit}
\label{app:fp_limit}

At a reinitialisation with a nonempty history,
letting $\disc\to0$ concentrates the softmax on
maximisers of the historical payoff within the
available population (uniformly over ties for the
equal-base weights used here). This explains the
hard-response endpoint of the initialisation rule.

The separately evaluated FP-style comparator uses one-round checks and, after every response event, puts equal weight on the historical-payoff maximisers in the available population, including when no new policy was found. Classical fictitious-play convergence~\citep{robinson1951iterative} concerns its specified empirical-response update. Algorithm~\ref{alg:vdo} instead reinitialises only on growth of the player's own population. The comparator is reported separately from the finite-$\disc\geq1$ guarantee of Theorem~\ref{thm:unified} and from neural fictitious self-play~\citep{heinrich2016deep}.

\section{Window-wise Accounting with Non-uniform
Priors}
\label{app:general_bound}

The standard non-uniform MWU inequality also
provides direct accounting when the discount
varies. It retains the actual prior probability
rather than optimising a separate rate
retrospectively.

\begin{proposition}[Window-wise regret with actual
initial probabilities]
\label{prop:windowwise} Suppose window $j$ starts
from a positive distribution $q_j$ on $\Pi^{(j)}$
and uses fixed rate $\eta>0$ for its $n_j$ rounds.
For a comparator $s$ first available in window
$j^*$,
\[
 \sum_{j=j^*}^J\sum_{t\in T_j}\bigl(\langle\pi^t,\ell^t\rangle-\ell^t(s)\bigr)
 \leq\frac1\eta\sum_{j=j^*}^J\ln\frac1{q_j(s)}
 +\frac\eta2\sum_{j=j^*}^Jn_j .
\]
If regret is measured from round one, add at most
$2\Tbar_{j^*}$ for the rounds before $s$ was
available.
\end{proposition}

\begin{proof} Apply~\eqref{eq:nonuniform_mwu} to
$s$ in every window in which it is available, and
sum. The earlier per-round loss difference is at
most two.
\end{proof}

For uniform initialisation $q_j(s)=1/|\Pi^{(j)}|$.
For historical initialisation the same expression
uses its actual $q_j(s)$; the eventual hindsight
comparator need not receive above-uniform
probability in every window. Thus the comparison
is through the realised prior costs, not a
universal ordering of priors.

\section{Additional Results from VDO Analysis}
\label{app:additional}

The uniform endpoint $\disc=\infty$ recovers the
uniform reset rule. Matching the full published
ODO algorithm additionally requires its response
target and checking protocol.
Proposition~\ref{prop:windowwise} accounts for the
initialisation cost under either chosen protocol.

\subsection{Three-axis design view: classical
algorithms and our contributions}
\label{app:three_axis_table}

Table~\ref{tab:unification} below summarises the
three-axis view of population expansion introduced
in \S\ref{sec:three_axes}.

\begin{table}[h]
\caption{Classical algorithms and our
contributions across the three axes of population
expansion. The $\disc$ column shows the discount
parameter (where applicable), $m(j)$ the BR check
frequency, and ``History view'' the time horizon
over which the meta-strategy solver integrates
information at each window start. The ODO row
reflects the theoretical analysis
of~\citet{dinh2022online} ($m(j)=1$, window-local
average); our experimental ODO baseline uses the
same shared protocol (BR target, oracle, window
schedule) as VDO/dVDO and differs only in the
reinitialisation rule. Convergence rates and bound
regimes are stated in Theorems~\ref{thm:vdo}
and~\ref{thm:unified}; the FP-style hard-BR limit
($\disc\to 0$) is included as a structural extreme
of the reinitialisation axis and lies outside our
bound.}
\label{tab:unification}
\centering
\small
\setlength{\tabcolsep}{4pt}
\begin{tabular}{lccl}
\toprule
Algorithm & $\disc$ & $m(j)$ & History view \\
\midrule
DO~\citep{mcmahan2003planning} & --- & $\infty$ &
restricted-game NE \\
ODO~\citep{dinh2022online} & $\infty$ & 1 &
window-local average \\
PDO~\citep{tang2023regret} & $\infty$ & schedule &
window-local (CFR, EFG) \\
FP-style hard-BR limit$^*$ & $\to 0$ & 1 & full
historical avg \\
\midrule
VDO (ours) & 1 & any & full historical avg \\
Discounted VDO (ours) & $\disc \geq 1$ & any &
discounted historical avg \\
\bottomrule
\end{tabular}
\\[2pt]
{\footnotesize $^*$ The FP-style row describes a structural
hard-response limit of the reinitialisation rule, not a guarantee for the
full trajectory of classical fictitious play; see
Appendix~\ref{app:fp_limit}. It lies outside Theorem~\ref{thm:unified}.}
\end{table}

\subsection{Variable-discount schedules}
\label{app:variable_disc}

For a positive window-dependent discount
$\disc_j$, set $q_j(s)\propto\exp[-\eta
L_{\Tbar_j}(s)/\disc_j]$ and apply
Proposition~\ref{prop:windowwise}. This gives a
bound in the actual prior costs for comparators
once they are available. A vanishing
average-regret rate additionally requires control
of those costs and of the pre-arrival
contribution. For constant $\disc$,
Theorem~\ref{thm:unified} gives the more specific
insertion-cost analysis.

\section{Connection to Quantal Response
Equilibria}
\label{app:qre}

The reinitialisation formula~\eqref{eq:vdo_reinit}
has the structure of a logit quantal
response~\citep{mckelvey1995quantal}: a softmax
over the time-average opponent with rationality
parameter $\eta\Tbar$. As the population grows and
$\Tbar$ accumulates, the rationality
sharpens---tracing a sequence of
restricted-population logit responses with growing
precision. This is suggestive of connections to
QRE-based meta-strategy solvers:
\citet{wang2023regularization} use QRE explicitly
as a regularised meta-solver in PSRO, and recent
work on generalised QRE establishes existence and
efficient learning in multi-player general-sum
games~\citep{shukla2025generalized}. Whether VDO's
trajectory of restricted-population responses
converges to a QRE-style equilibrium in
multi-player settings is an interesting open
question that we leave to future work.

\section{Additional Experimental Results}
\label{app:more_experiments}

\begin{figure}[H]
\centering
\begin{subfigure}[b]{0.32\textwidth}
\includegraphics[width=\textwidth]{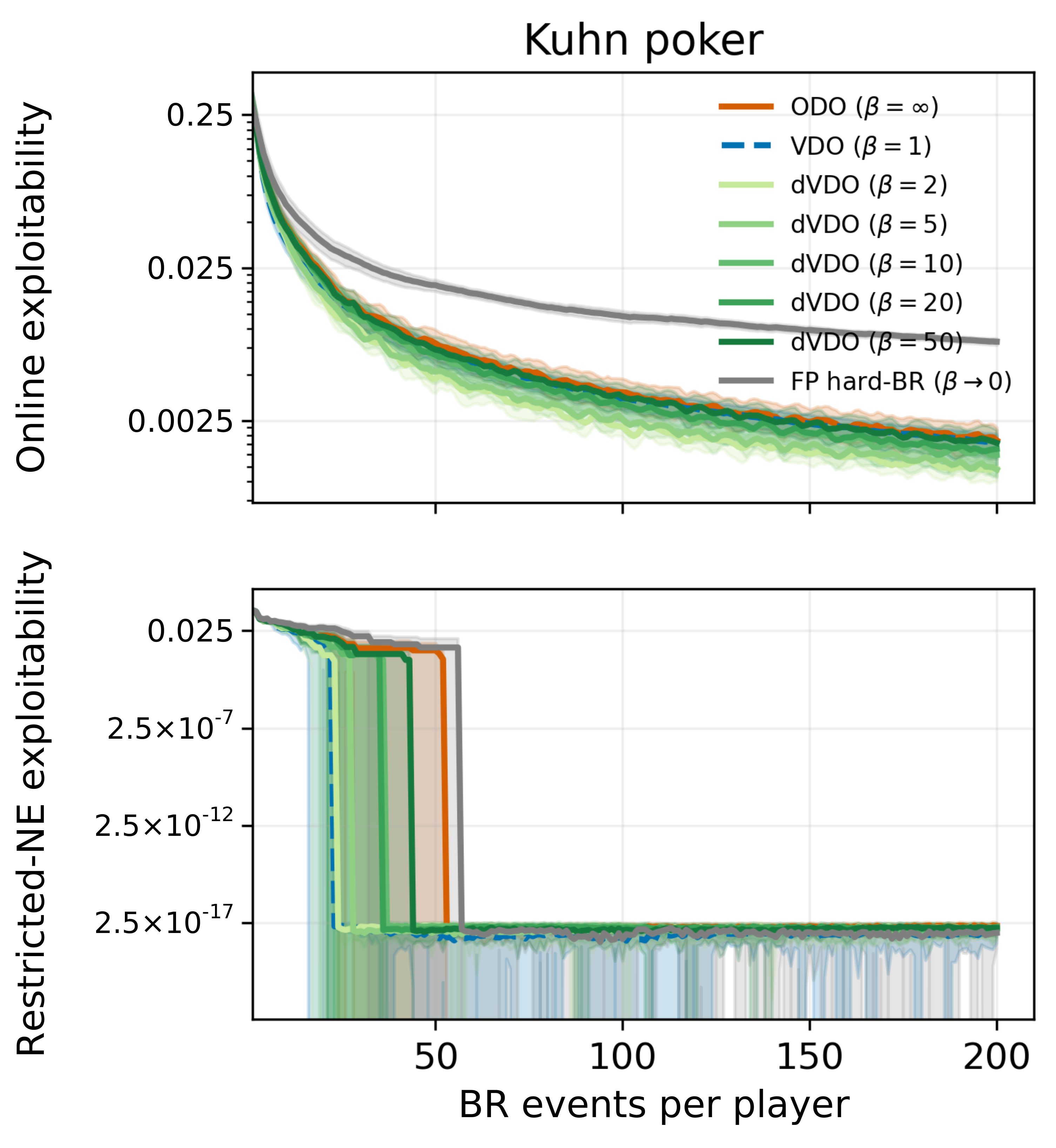}
\caption{Kuhn poker.}
\end{subfigure}\hfill
\begin{subfigure}[b]{0.32\textwidth}
\includegraphics[width=\textwidth]{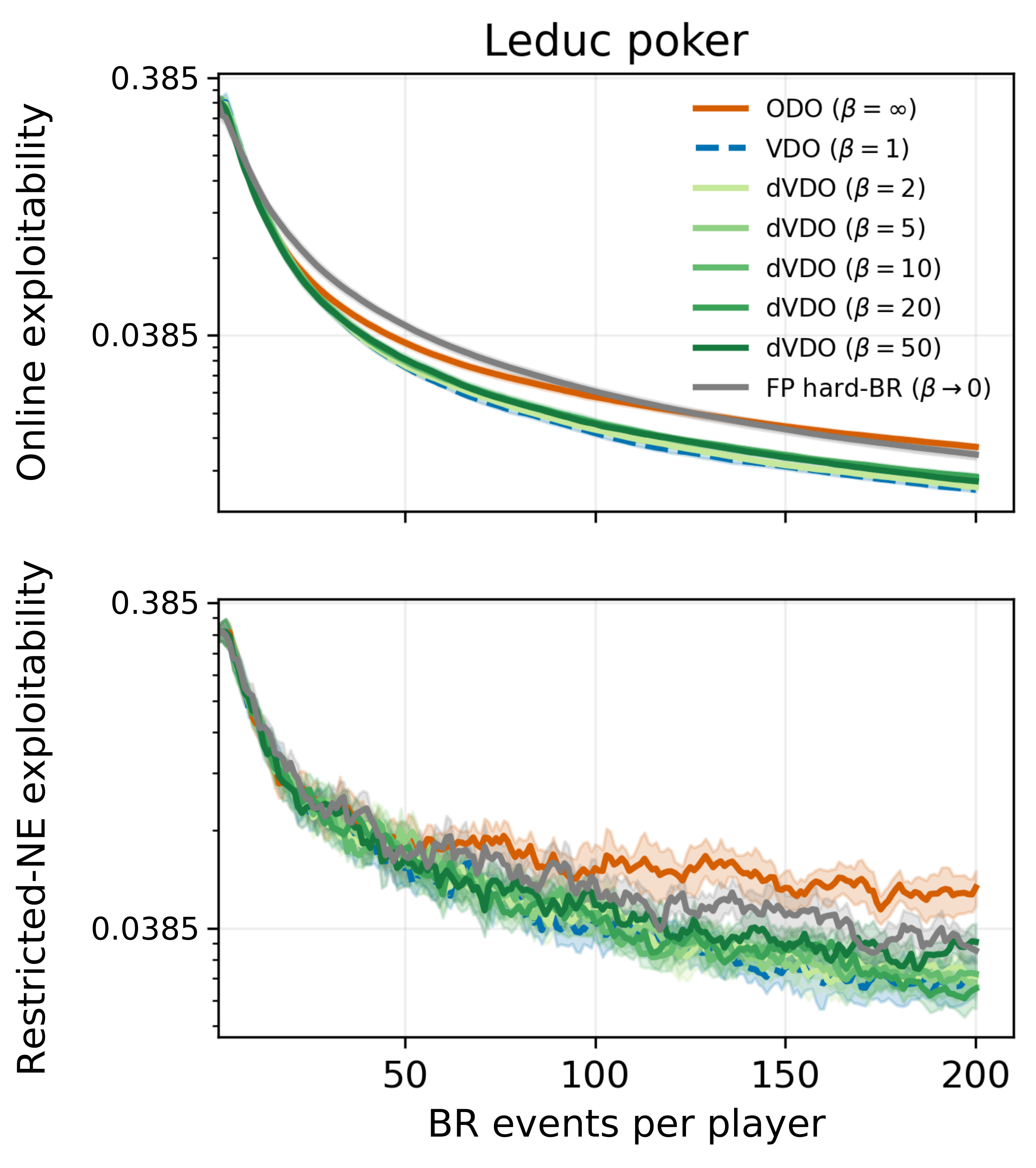}
\caption{Leduc poker.}
\end{subfigure}\hfill
\begin{subfigure}[b]{0.32\textwidth}
\includegraphics[width=\textwidth]{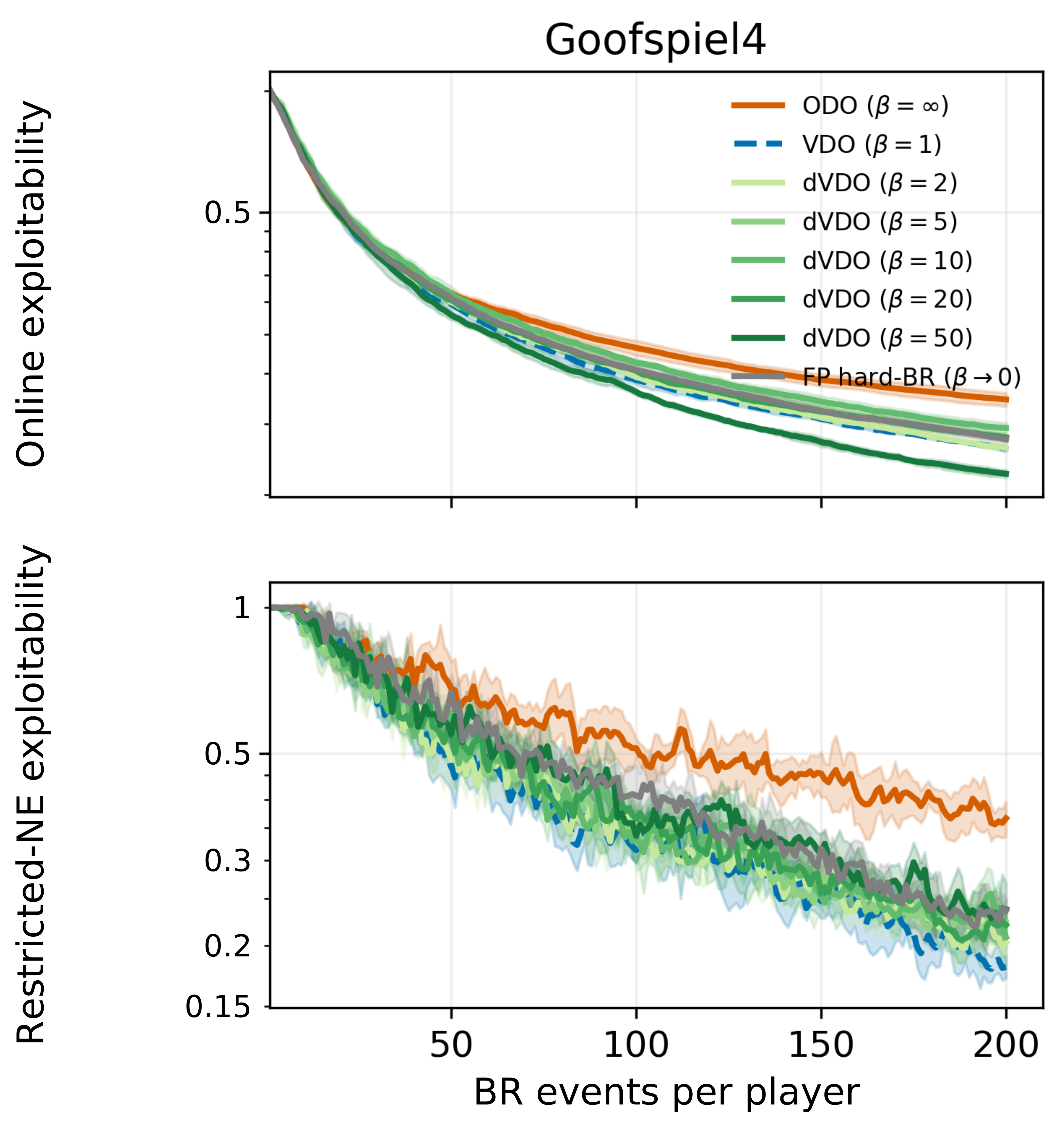}
\caption{Goofspiel-4.}
\end{subfigure}
\caption{Fixed-$\disc$ trajectories on
Kuhn, Leduc and Goofspiel-4; each curve is one
value of $\disc$. RNE panels use the
diagnostic defined in
Section~\ref{sec:experiments}. ODO
($\disc=\infty$) is solid orange, VDO ($\disc=1$)
dashed blue and FP-hardBR ($\disc\to 0$) grey;
dVDO runs from light to dark green as $\disc$
increases from 2 to 50.}
\label{fig:traj_imperfect_info}
\end{figure}

Figure~\ref{fig:traj_imperfect_info} gives the full discount sweeps for the PSRO experiments of Section~\ref{sec:experiments}. The remaining figures use the same per-player discovery and reset protocol with normalised training payoffs. For the MWU-based methods, both players update simultaneously, and a player queries a best response to the full-history opponent average when its own countdown expires. Only a response that expands that player's population triggers reinitialisation. The FP hard-BR comparator instead reselects payoff maximisers after every round.

Utilities are divided by a fixed game-wide scale: 2 for Kuhn, 13 for Leduc, 1 for Goofspiel-4, and $\max_{a,b}|A_{ab}|$ for each sampled Gaussian matrix. Extensive-form mixtures are realisation equivalent. Runs start from random pure policies with seeds $1,\ldots,10$ ($1,\ldots,5$ for Goofspiel-4). Methods share initial policies within each game and seed. Each run records 200 response events per player, except for the long-horizon study. At index $j$, the metric averages the evaluations immediately before the two players' $j$th responses, which can occur at different learning rounds.

\begin{figure}[H]
\centering
\begin{subfigure}[b]{0.32\textwidth}
\includegraphics[width=\textwidth]{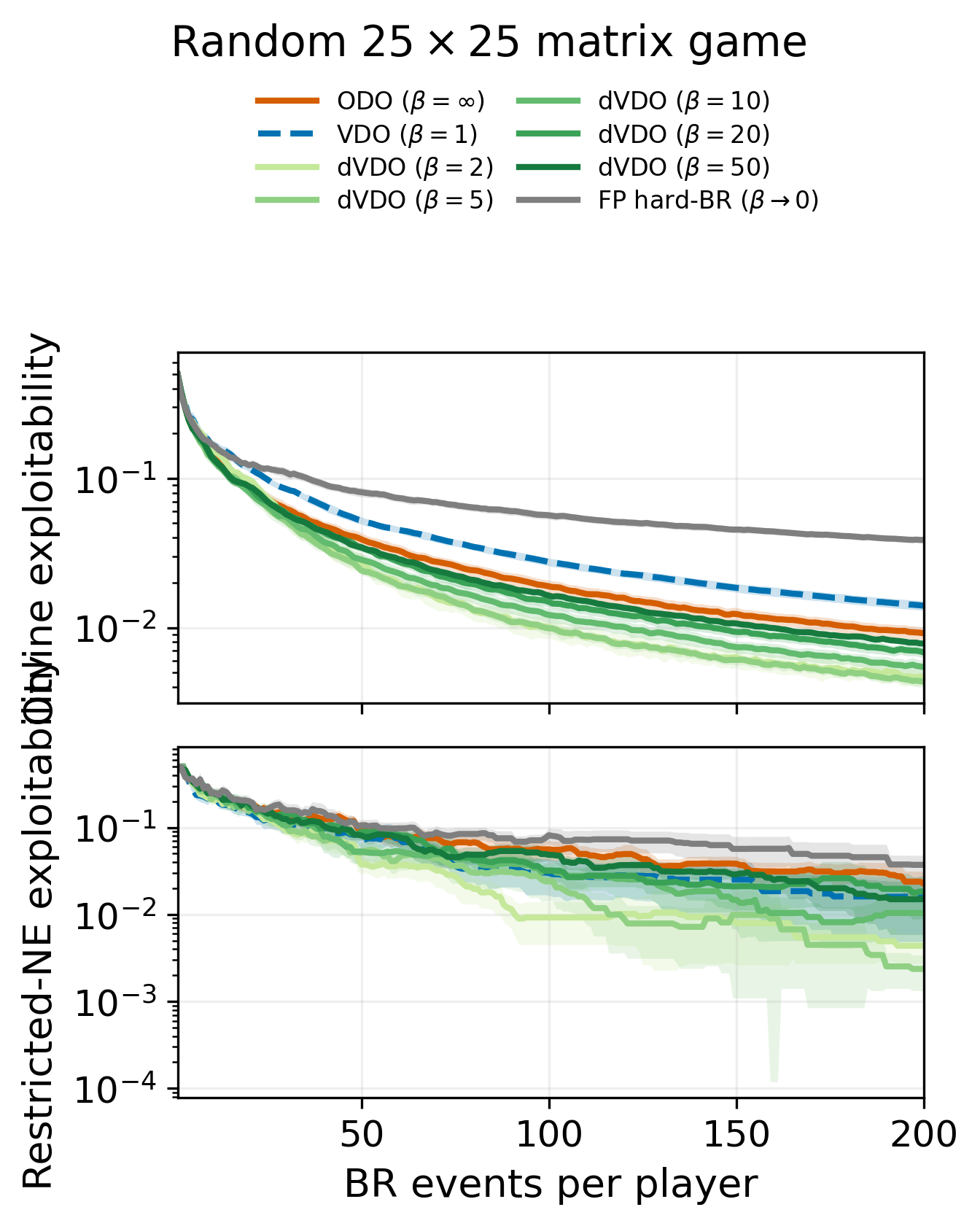}
\caption{$n=25$.}
\end{subfigure}\hfill
\begin{subfigure}[b]{0.32\textwidth}
\includegraphics[width=\textwidth]{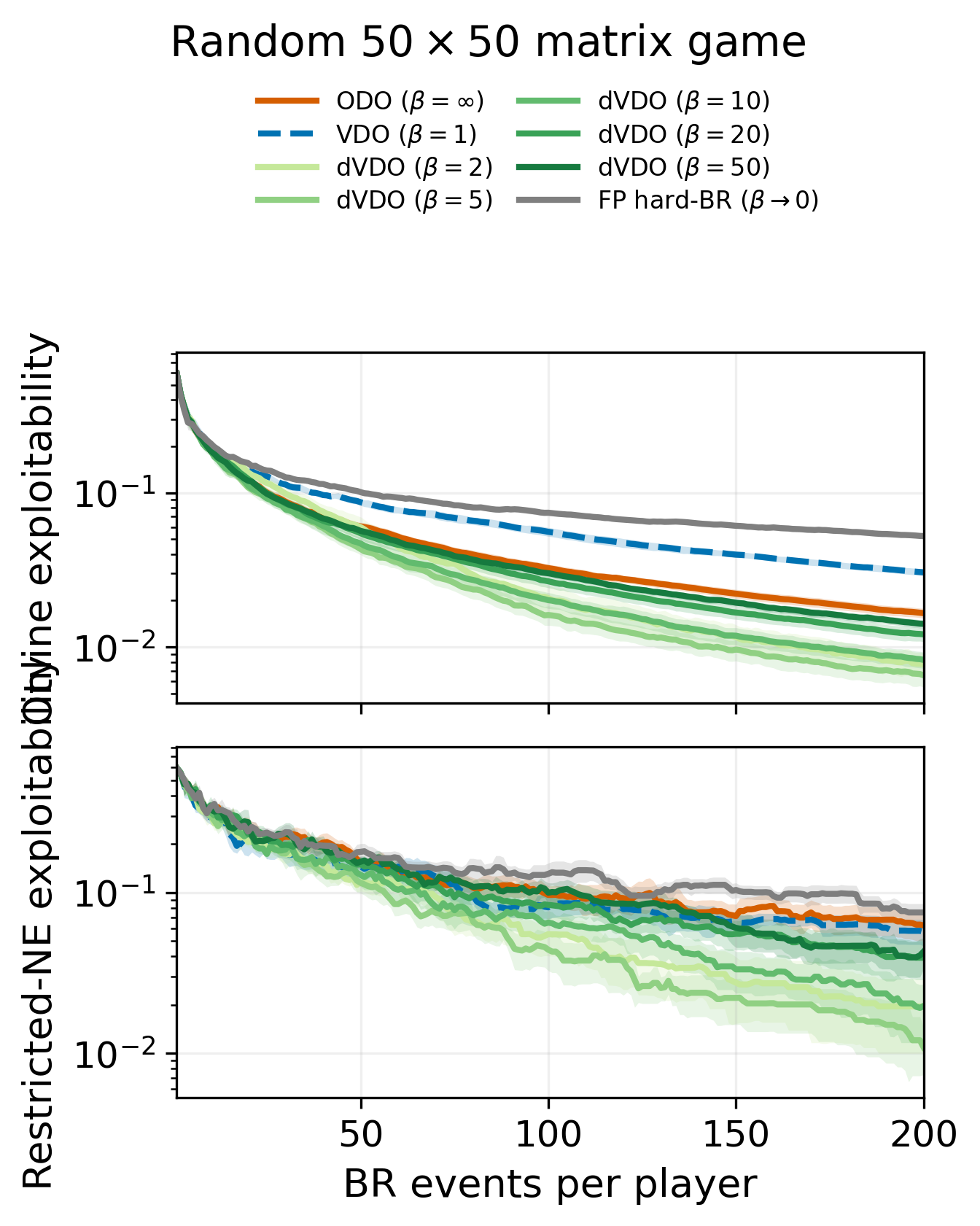}
\caption{$n=50$.}
\end{subfigure}\hfill
\begin{subfigure}[b]{0.32\textwidth}
\includegraphics[width=\textwidth]{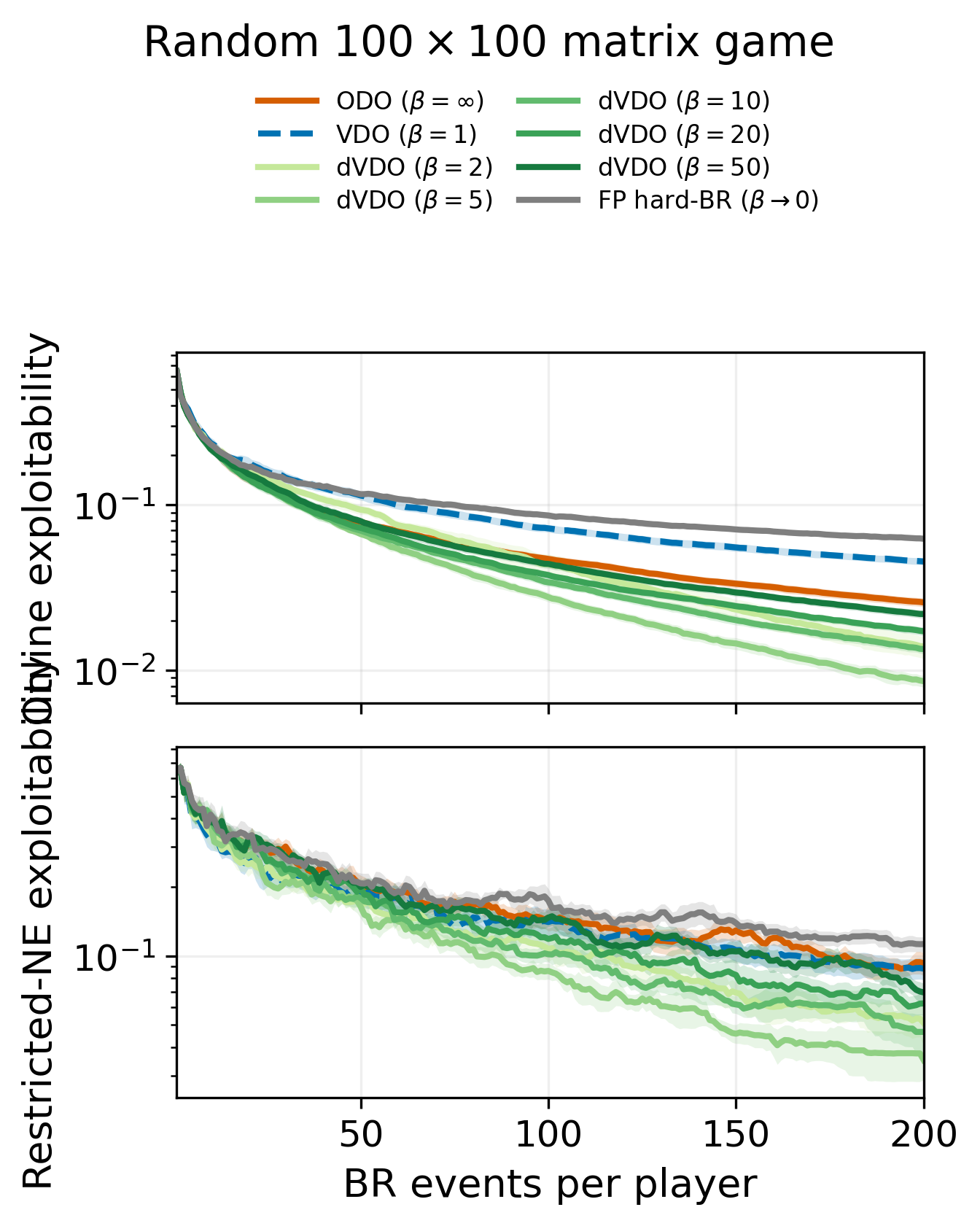}
\caption{$n=100$.}
\end{subfigure}
\caption{Fixed-$\disc$ trajectories and the FP hard-BR baseline on normalised Gaussian matrices of sizes 25, 50 and 100. Top: online half NashConv; bottom: restricted-NE half NashConv. Each panel uses one fixed matrix and ten trajectories. The horizontal axis indexes each player's response events; values average the evaluations at the two corresponding events.}
\label{fig:traj_matrix}
\end{figure}

Endpoint summaries use the final recorded event index, then average across seeds. Bands and error bars show means plus or minus 1.96 standard errors, with the population standard deviation convention used for the PSRO plots. Each matrix size uses one fixed normalised matrix, so its interval describes variation over initial strategies rather than over matrix instances. All metrics are normalised half NashConv. In the logarithmic panels of Figures~\ref{fig:traj_matrix}--\ref{fig:ushape_all}, means and interval limits below $10^{-4}$ are drawn at $10^{-4}$.

\begin{figure}[H]
\centering
\includegraphics[width=0.75\textwidth]{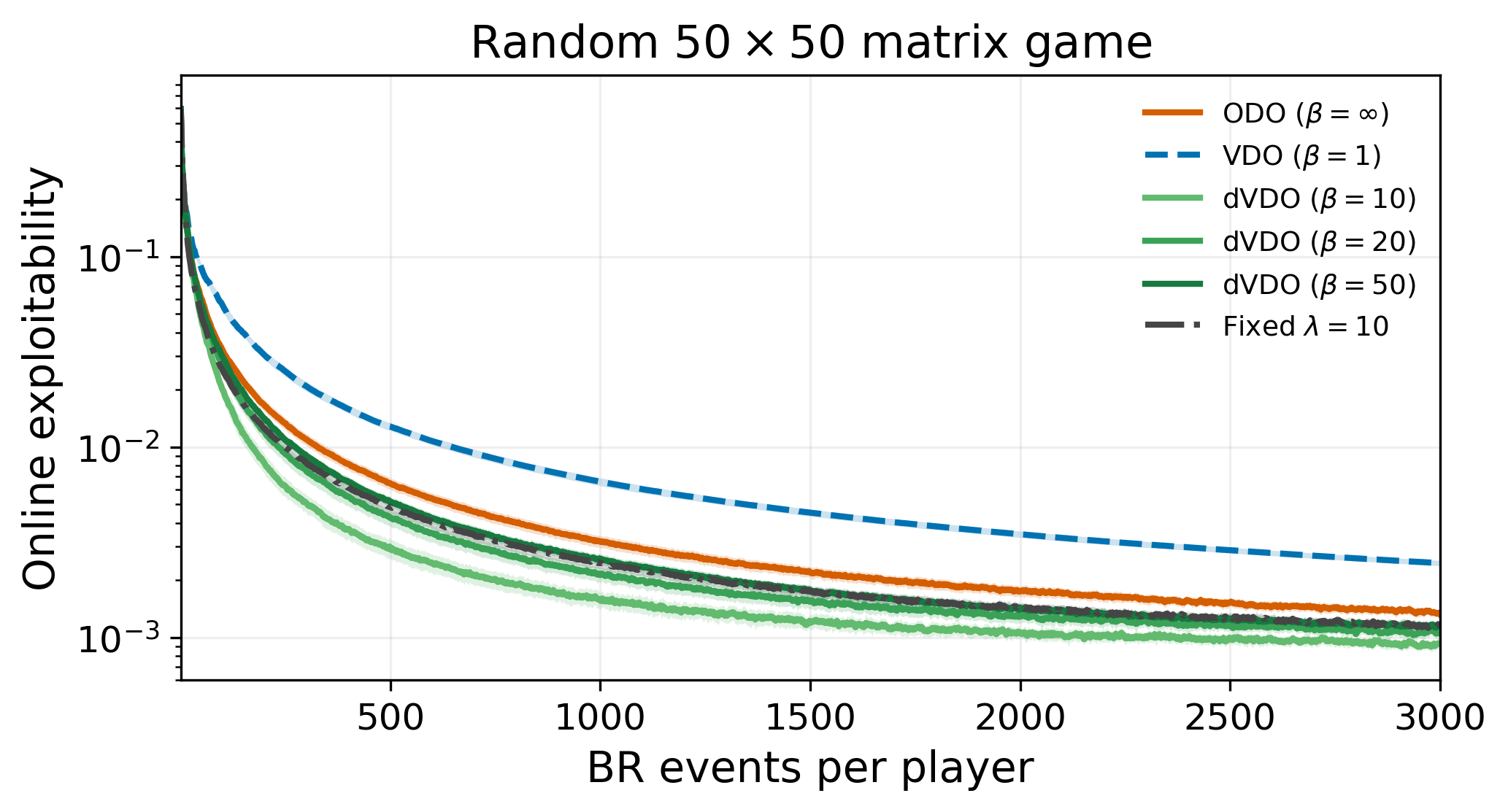}
\caption{Long-horizon comparison on one normalised Gaussian $50\times50$ matrix, with ten seeds and 3,000 response events per player. Online exploitability is half NashConv. VDO reconstructs with precision $\eta t$, dVDO with $\eta t/\disc$, and the fixed-precision control with $\lambda=10$, where $t$ counts learning rounds. Only the player adding a new strategy is reinitialised.}
\label{fig:long_horizon}
\end{figure}

\paragraph{Protocol details.} A singleton MWU population is checked after one learning round. The FP hard-BR comparator assigns equal weight to all restricted strategies maximising payoff against the global-average opponent, including after rediscovery. For Goofspiel-4, $M=1$, and Figure~\ref{fig:ushape_all}(b) uses the endpoints of the exact-response trajectories in Figure~\ref{fig:traj_imperfect_info}(c). The long-horizon grid in Figure~\ref{fig:long_horizon} contains $\disc\in\{1,10,20,50,\infty\}$ and the fixed-precision control; $\disc=5$ is not included.

\paragraph{Metric interpretation.} RNE evaluates the restricted-game equilibrium selected by the linear program; it does not minimise full-game exploitability over all profiles supported on that population, as MRCP does~\citep{jordan2010strategy}. Theorems~\ref{thm:vdo} and~\ref{thm:unified} bound online exploitability over the discovered population; full-game exploitability also includes the discovery gap in Remark~\ref{rem:scope}.

\subsection{Learning-rate sensitivity}
\label{app:eta_robust}

\begin{figure}[H]
\centering
\includegraphics[width=0.65\textwidth]{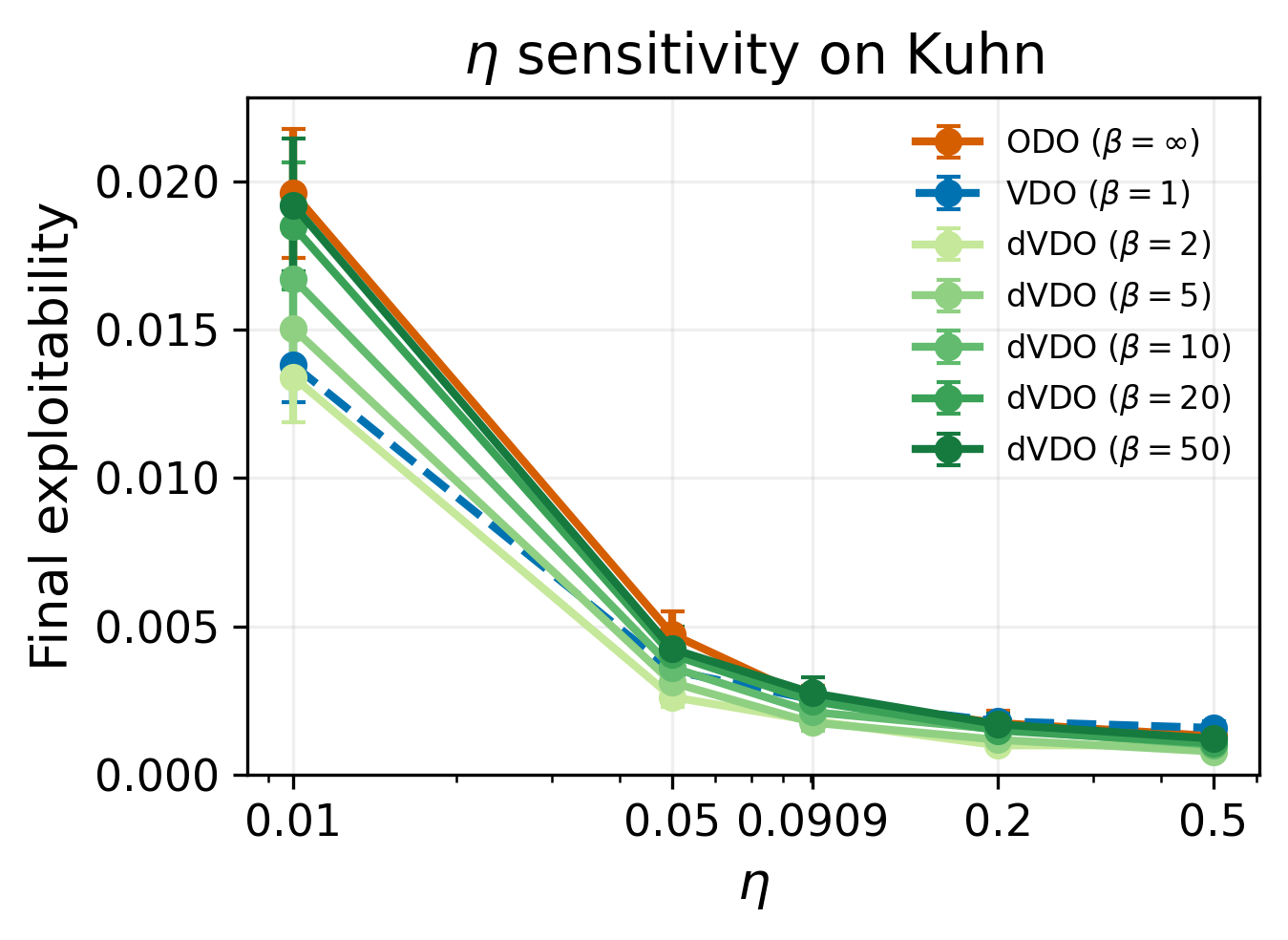}
\caption{Kuhn learning-rate sensitivity with normalised training payoffs and per-player discovery. Points show normalised online half NashConv at the final event index (200 events per player), averaged over ten seeds. The curves show ODO, VDO and dVDO at $\disc\in\{2,5,10,20,50\}$.}
\label{fig:eta_robust}
\end{figure}

\subsection{Full trajectory plots}
\label{app:full_traj}

\subsection{Matrix-game baselines comparison}
\label{app:matrix_baselines}

\begin{figure}[H]
\centering
\begin{subfigure}[b]{0.32\textwidth}
\includegraphics[width=\textwidth]{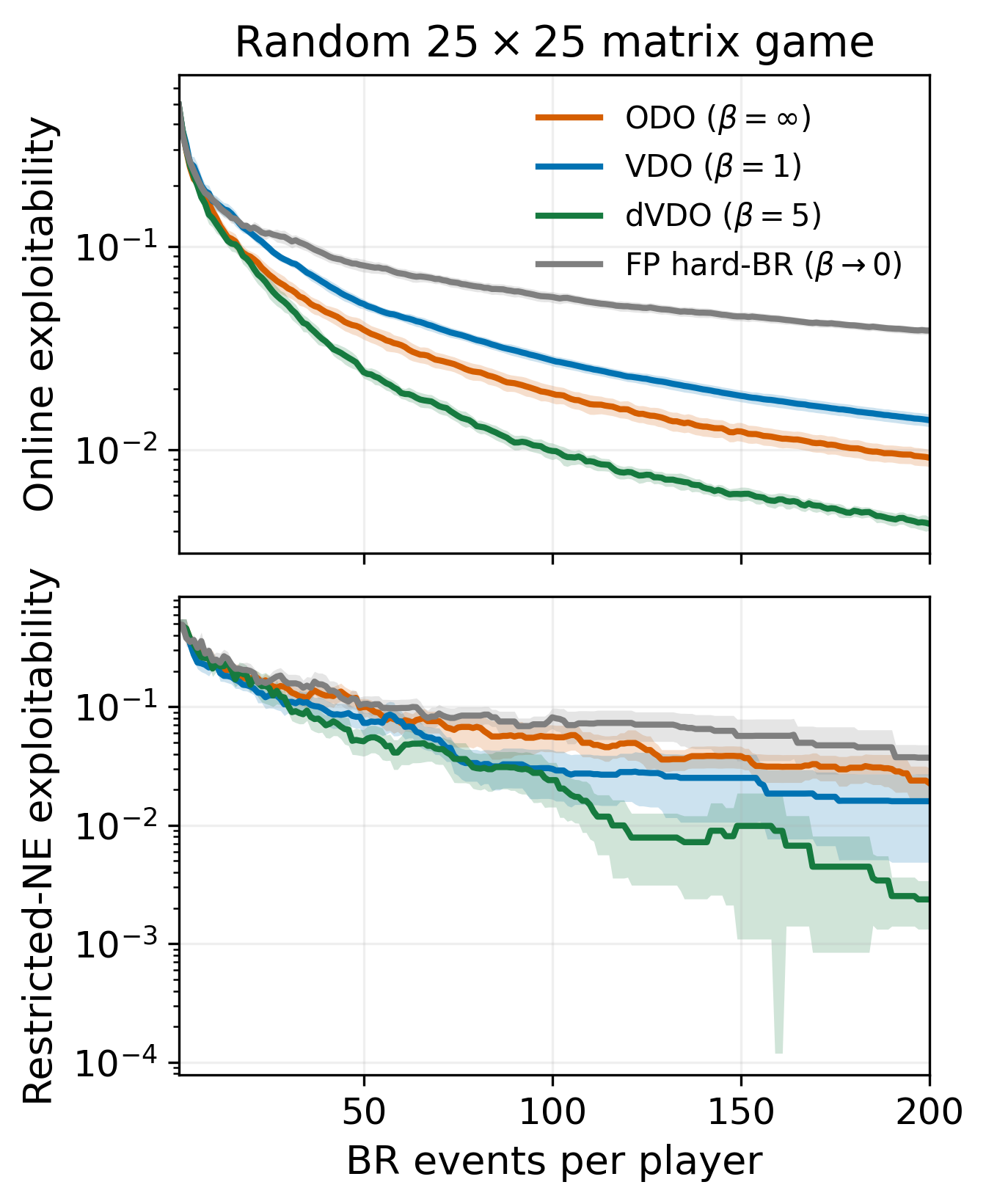}
\caption{$n=25$.}
\end{subfigure}\hfill
\begin{subfigure}[b]{0.32\textwidth}
\includegraphics[width=\textwidth]{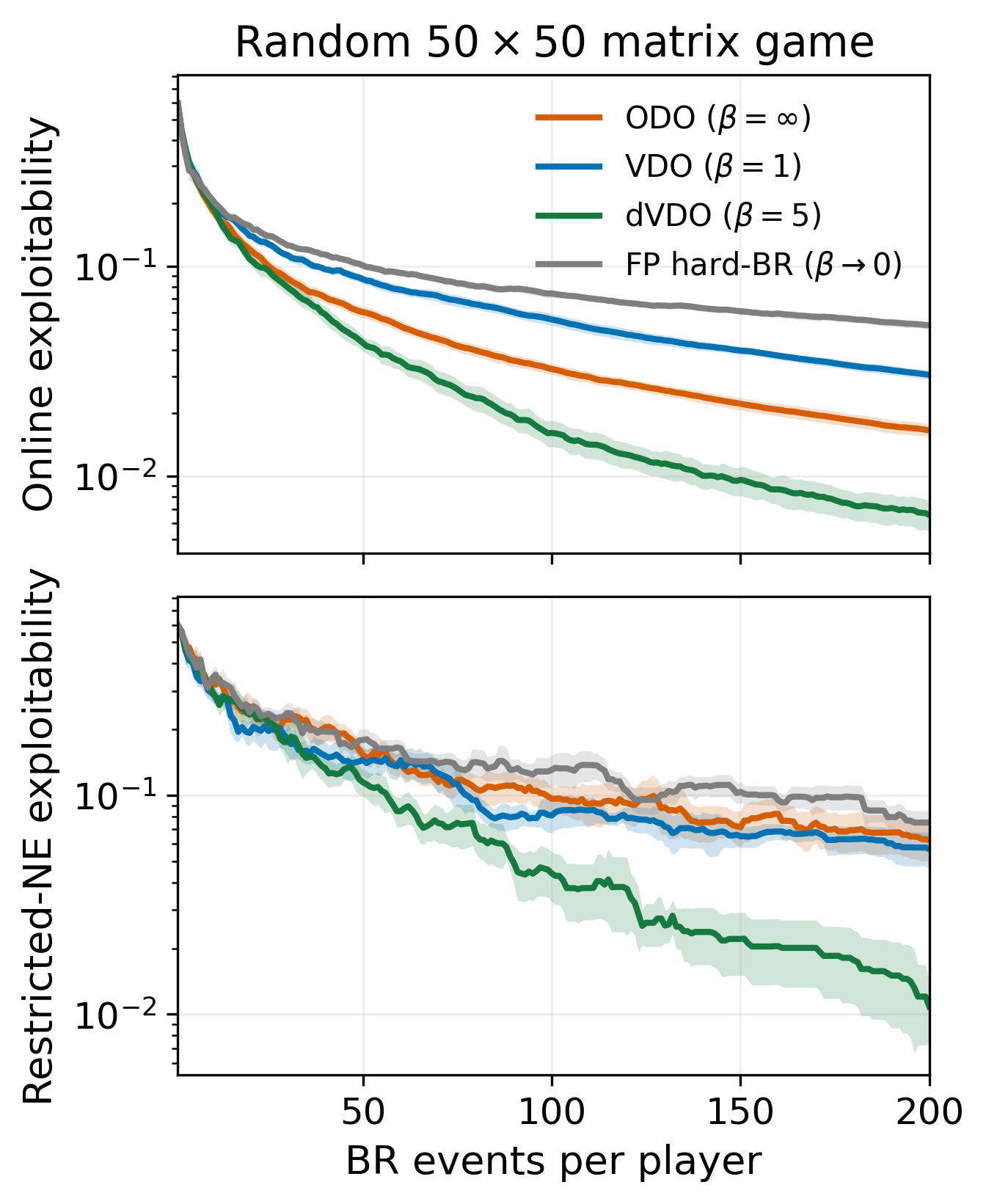}
\caption{$n=50$.}
\end{subfigure}\hfill
\begin{subfigure}[b]{0.32\textwidth}
\includegraphics[width=\textwidth]{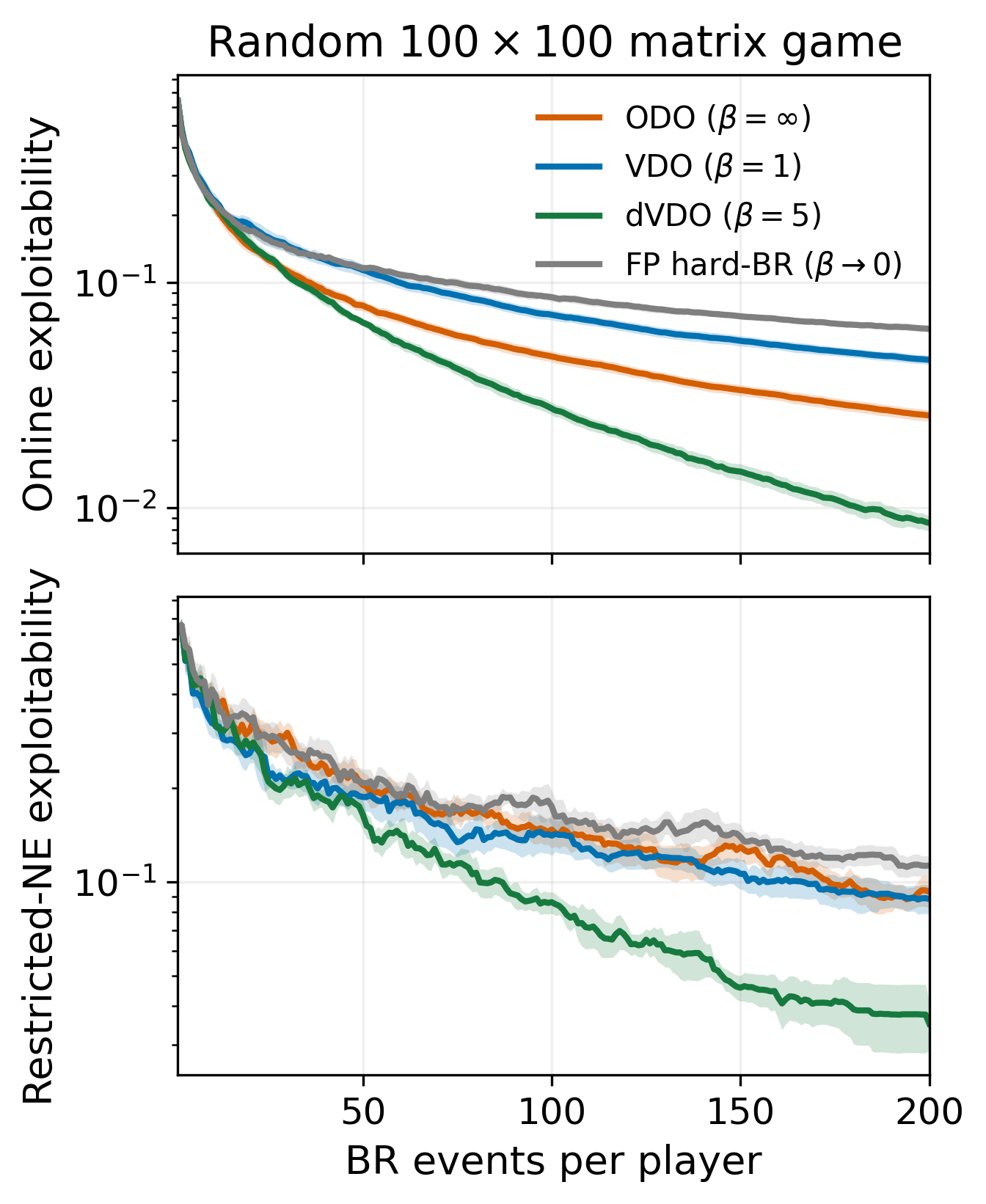}
\caption{$n=100$.}
\end{subfigure}
\caption{Online (top of each panel) and restricted-NE (RNE, bottom of each panel) half NashConv on normalised matrix games with per-player discovery and resets. VDO uses $\disc=1$; dVDO-best uses $\disc=5$, selected in hindsight from $\{2,5,10,20,50\}$ by mean online exploitability at the final event index. Each size uses one fixed matrix, ten seeds and 200 events per player.}
\label{fig:matrix_baselines}
\end{figure}

Figure~\ref{fig:matrix_baselines} compares ODO, VDO, dVDO and the FP-style hard-BR comparator on the same normalised matrices. The hard-BR comparator uses $\disc=0$ and one-round checks, retargeting the historical payoff maximisers at every response event.

\subsection{Discount sensitivity}
\label{app:full_ushape}

\begin{figure}[H]
\centering
\begin{subfigure}[b]{0.32\textwidth}
\includegraphics[width=\textwidth]{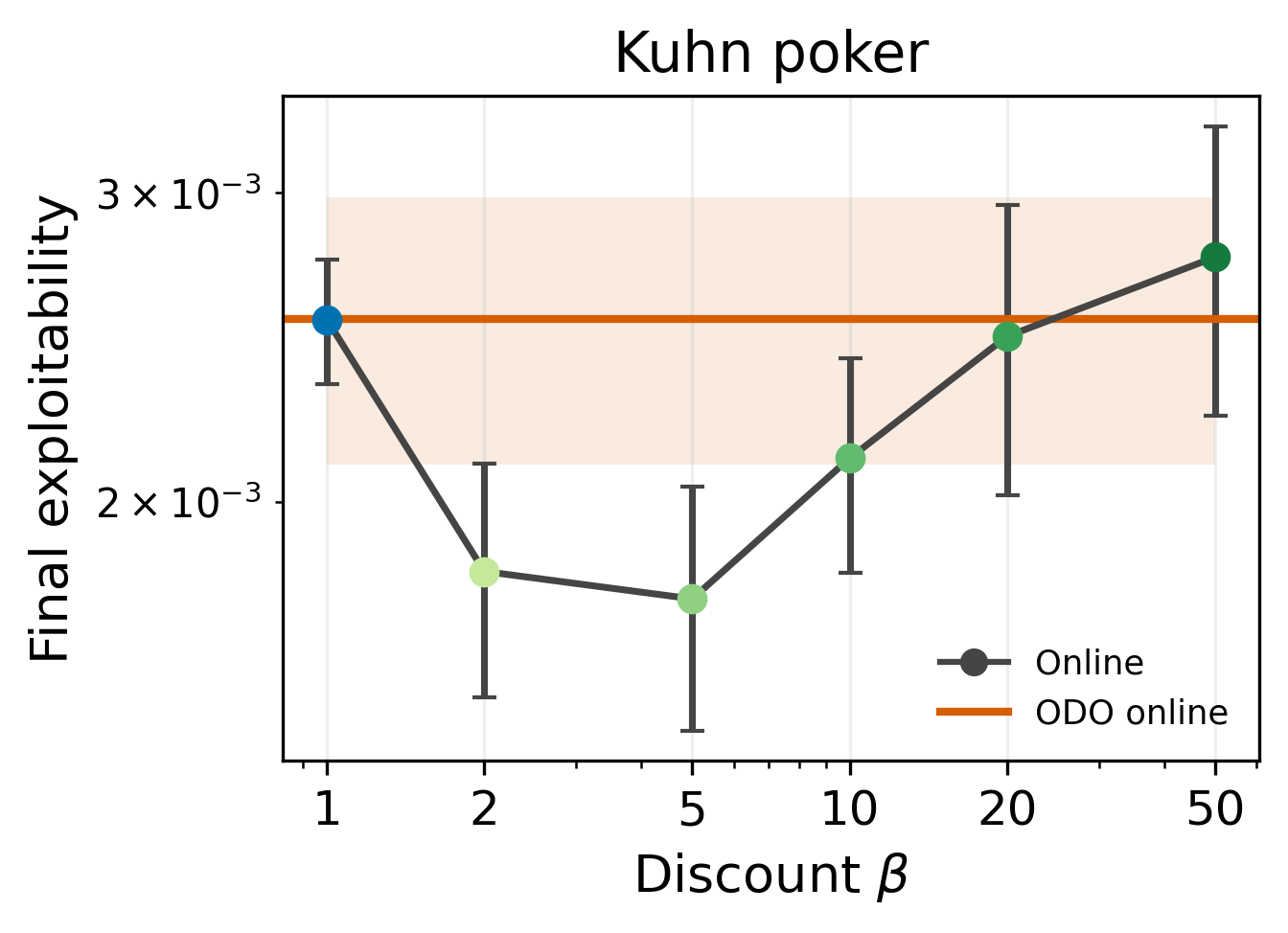}
\caption{Kuhn poker.}
\end{subfigure}\hfill
\begin{subfigure}[b]{0.32\textwidth}
\includegraphics[width=\textwidth]{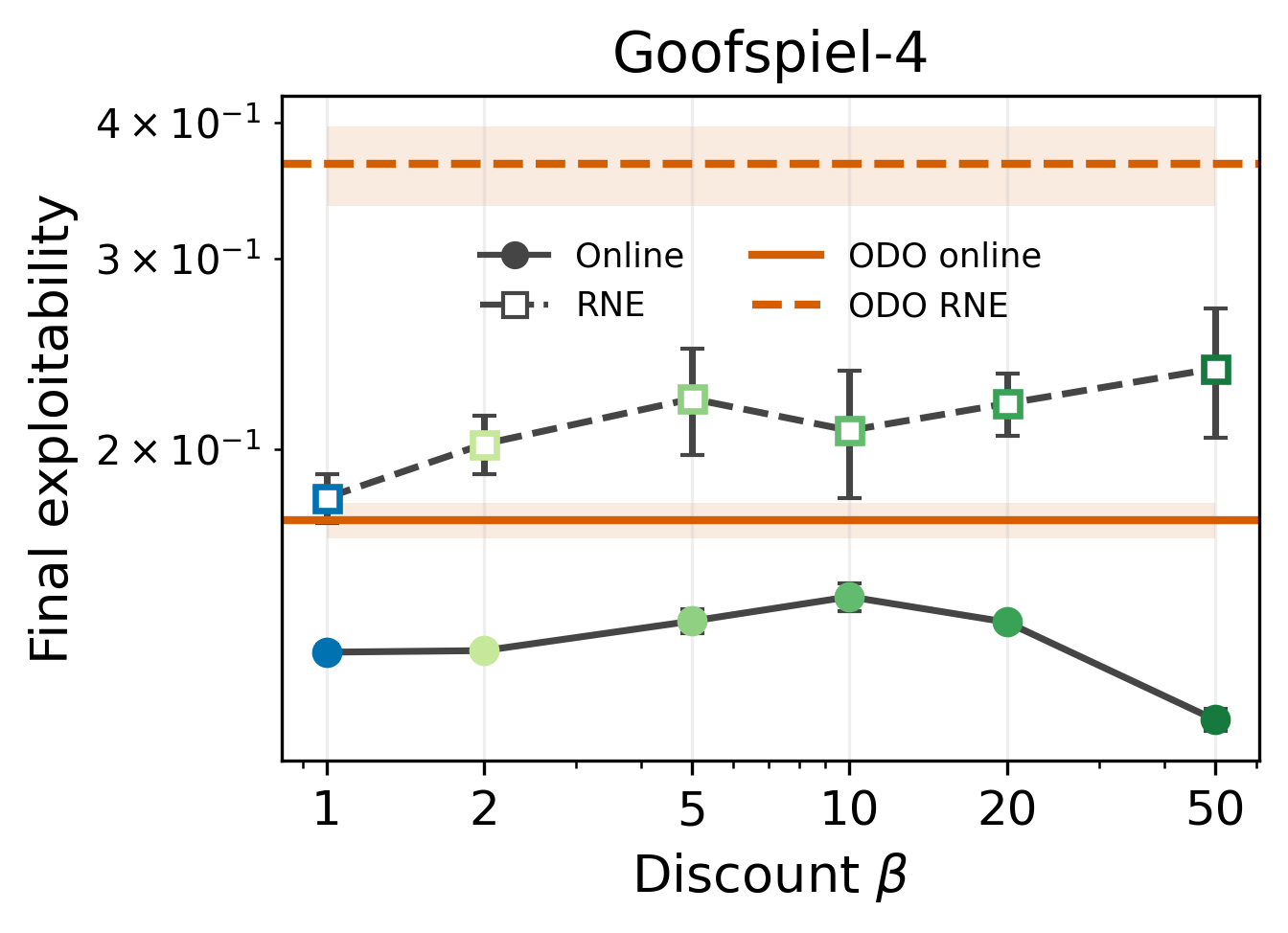}
\caption{Goofspiel-4.}
\end{subfigure}\hfill
\begin{subfigure}[b]{0.32\textwidth}
\includegraphics[width=\textwidth]{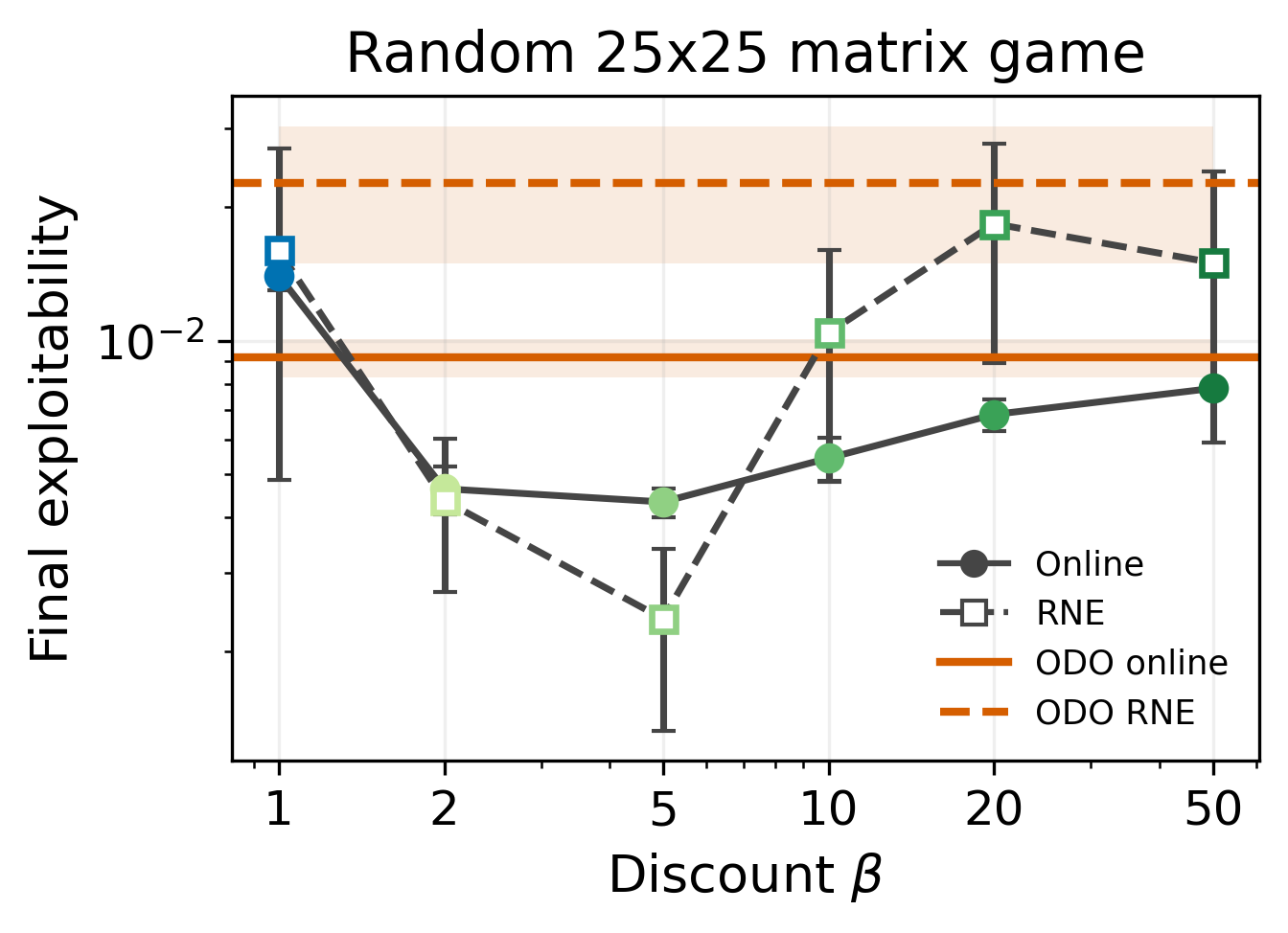}
\caption{Random $25\times 25$.}
\end{subfigure}
\\[4pt]
\begin{subfigure}[b]{0.32\textwidth}
\includegraphics[width=\textwidth]{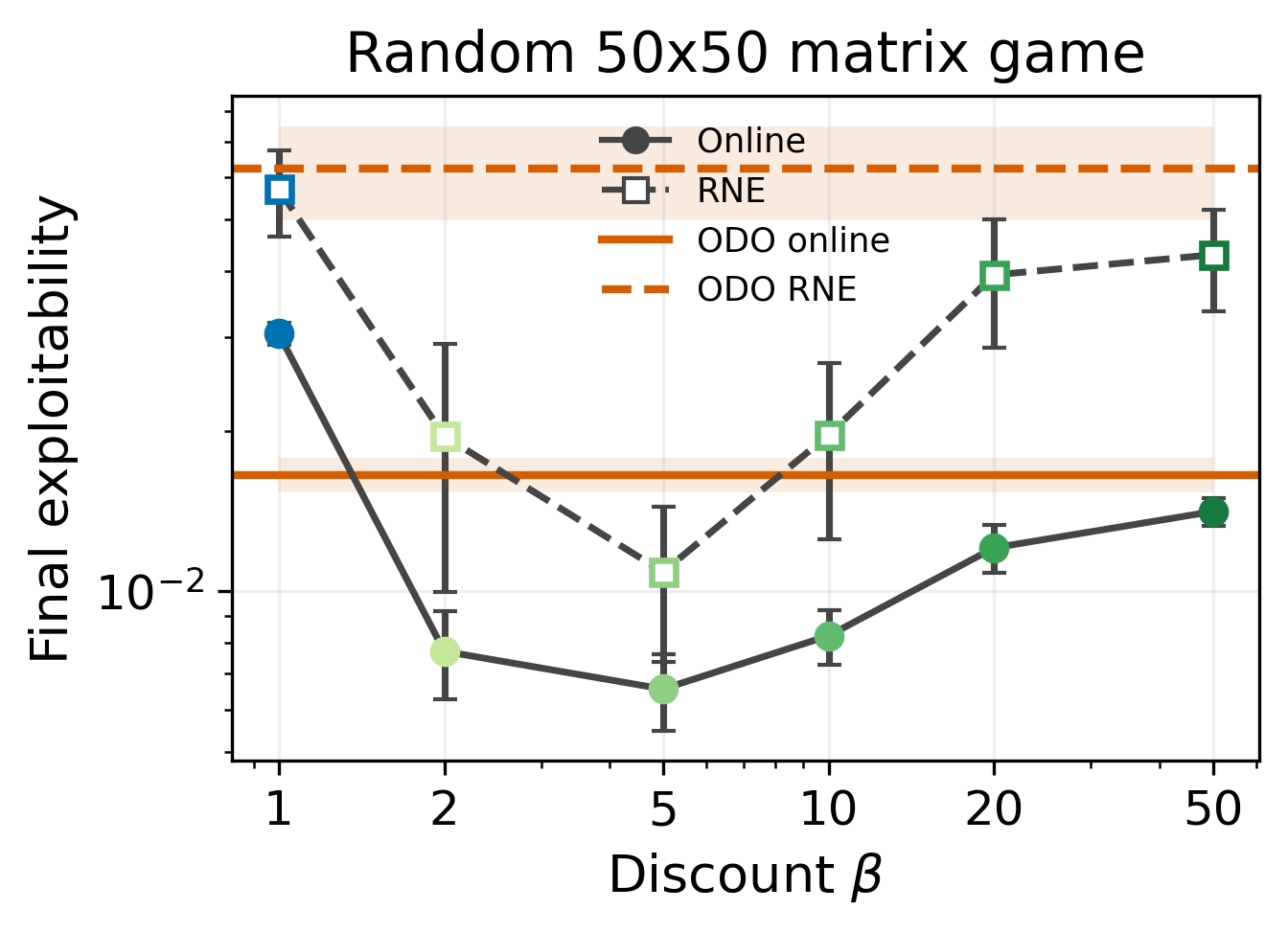}
\caption{Random $50\times 50$.}
\end{subfigure}\hfill
\begin{subfigure}[b]{0.32\textwidth}
\includegraphics[width=\textwidth]{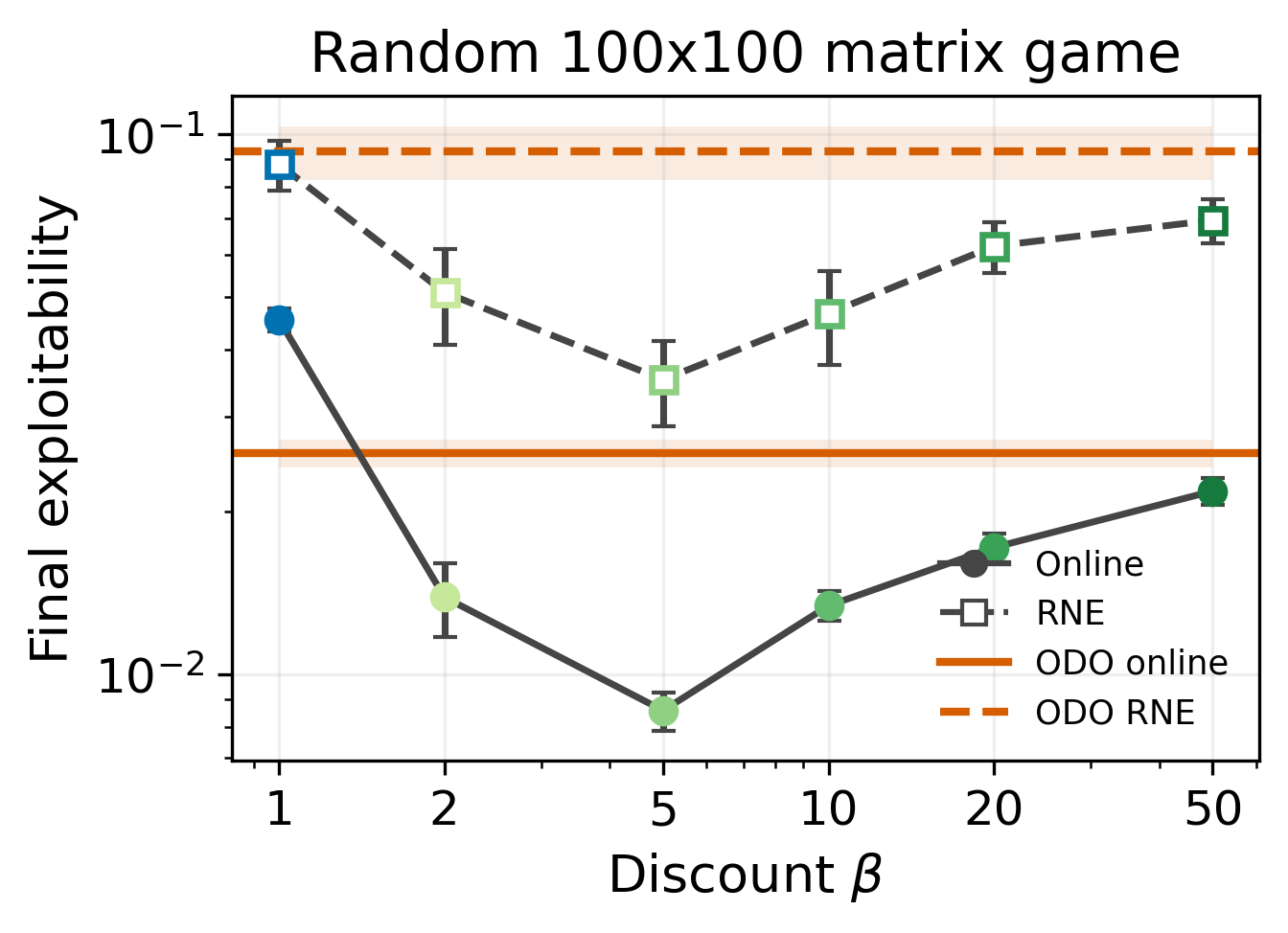}
\caption{Random $100\times 100$.}
\end{subfigure}\hfill
\begin{subfigure}[b]{0.32\textwidth}
\includegraphics[width=\textwidth]{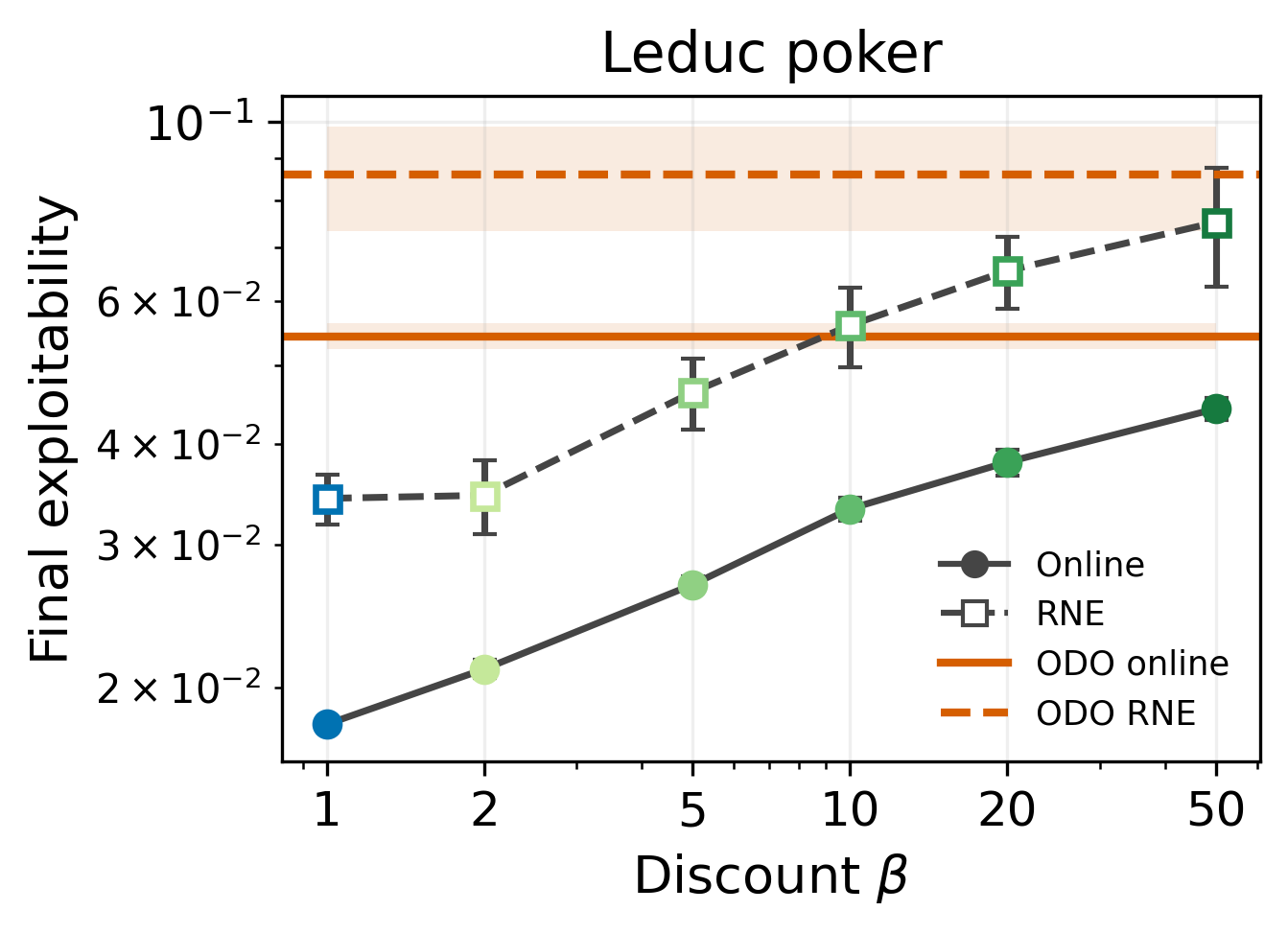}
\caption{Leduc poker.}
\end{subfigure}
\caption{Discount sensitivity with normalised training payoffs and per-player discovery. Points show the final event index (200 events per player), then average across seeds: ten for Kuhn, Leduc and each matrix size, and five for Goofspiel-4. All metrics are normalised half NashConv. Circles show online exploitability and squares show RNE. On Kuhn, all mean RNE values are below $10^{-15}$, so only online exploitability is shown. The Goofspiel-4 points are the endpoints of the exact-response trajectories in Figure~\ref{fig:traj_imperfect_info}(c).}
\label{fig:ushape_all}
\end{figure}

\begin{figure}[H]
\centering
\includegraphics[width=0.55\textwidth]{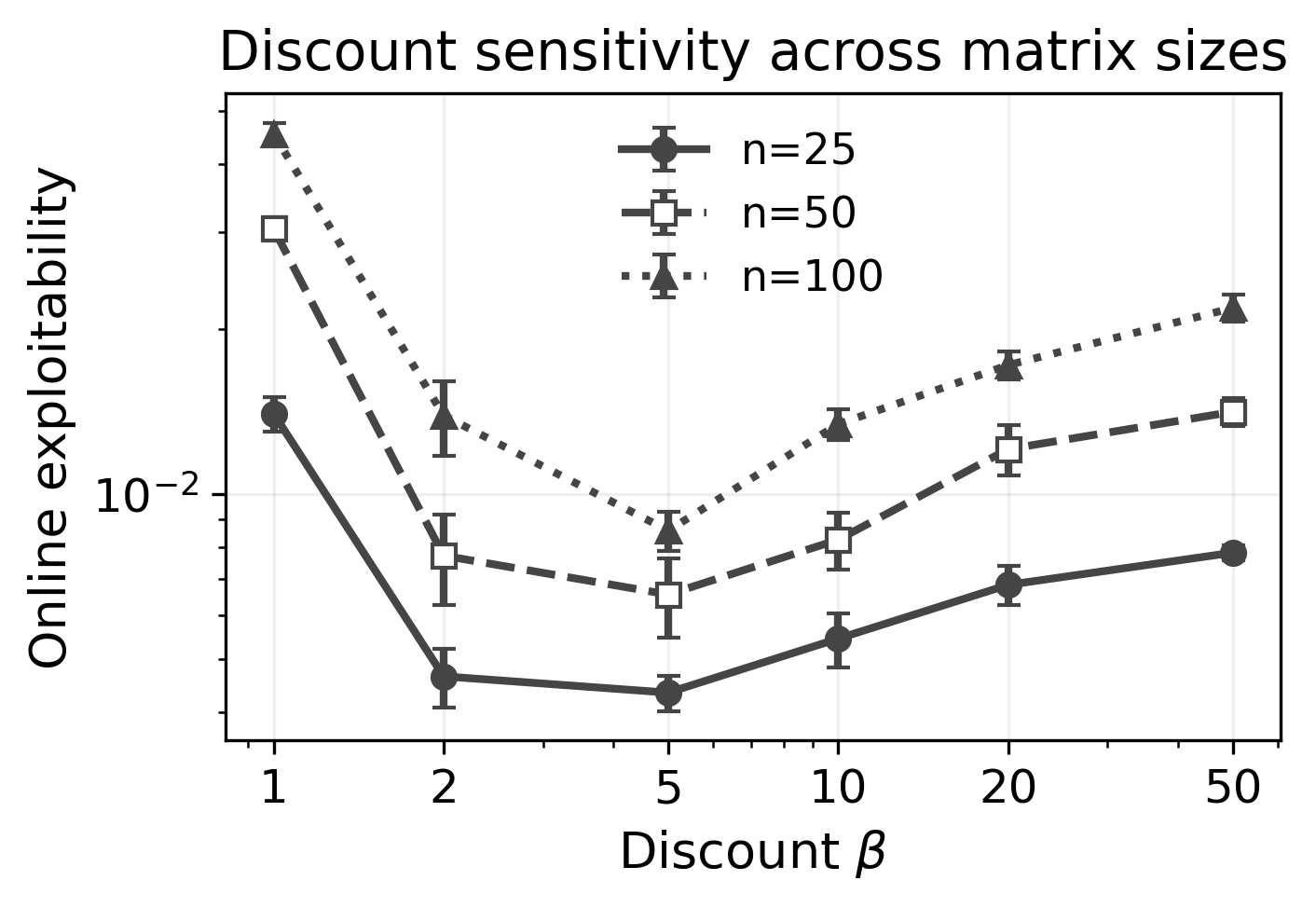}
\caption{Online half NashConv versus discount on three fixed normalised Gaussian matrices. Points use the final event index and ten seeds per size. The smallest mean is at $\disc=5$ for all three sizes.}
\label{fig:matrix_size}
\end{figure}

\paragraph{Discount sensitivity across matrix sizes.}
\label{app:matrix_size}
Figure~\ref{fig:matrix_size} compares online half NashConv over a common grid $\disc\in\{1,2,5,10,20,50\}$. The lowest sample mean at the final event index is at $\disc=5$ for each of the three fixed matrix instances; this is a descriptive comparison, not a scaling law.

\paragraph{Online and RNE exploitability on Leduc.} In Figure~\ref{fig:ushape_all}(f), VDO ($\disc=1$) has the lowest mean online and restricted-NE exploitability among the tested discounts, although the restricted-NE means at $\disc=1$ and $\disc=2$ are close. Both diagnostics use full-game best responses: the first measures time-averaged play, while the second measures a restricted equilibrium of the discovered population.

\subsection{Long-horizon illustration}
\label{app:long_horizon}

\subsection{Broader societal impact}
\label{app:broader_impact}

The contribution of this paper is
retrospective MWU-weight reconstruction and
convergence analysis for an existing game-solving
algorithmic family (Double Oracle / PSRO); we
introduce no new model, dataset, or deployable
system. Faster Nash solvers inherit the dual-use
considerations of the broader DO/PSRO
family---applications include adversarial planning
and security-game equilibria, both well-discussed
in that literature---and we see no foreseeable
direct path to harmful application that would not
equally apply to existing baselines such as ODO or
PSRO. The contribution is at the
meta-solver layer (axis~3 in our three-axis
framework), and any deployment downstream of a
faster meta-solver would still depend on the BR
oracle (axis~2) and the application context, which
our work does not modify. We do not release any
models, datasets, or pretrained artefacts that
would require additional safeguards.

\section{Results on Additional Games}
\label{app:results}

\begin{figure}[H]
\centering
\begin{subfigure}[b]{0.32\textwidth}
\includegraphics[width=\textwidth]{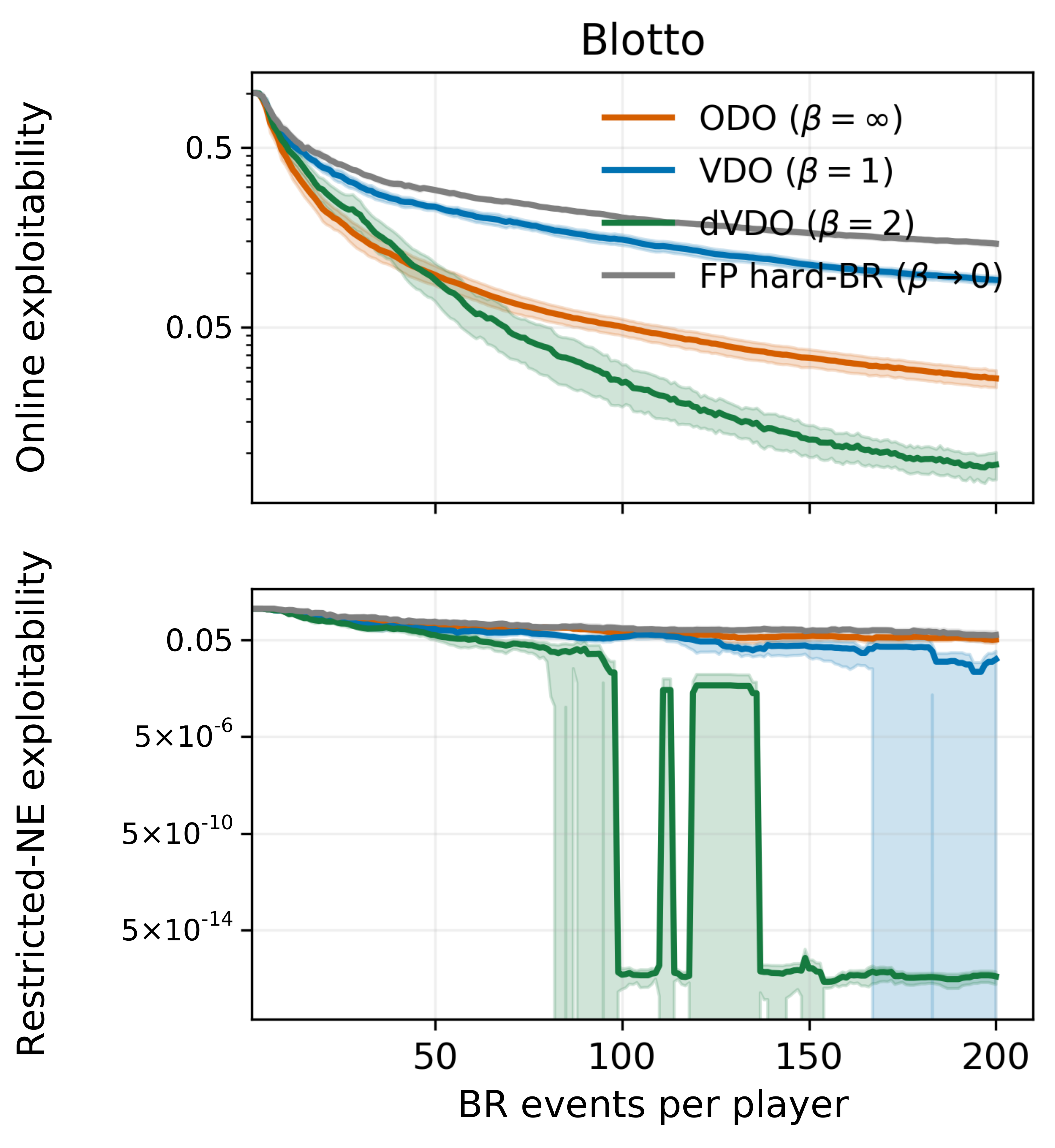}
\caption{Blotto.}
\end{subfigure}\hfill
\begin{subfigure}[b]{0.32\textwidth}
\includegraphics[width=\textwidth]{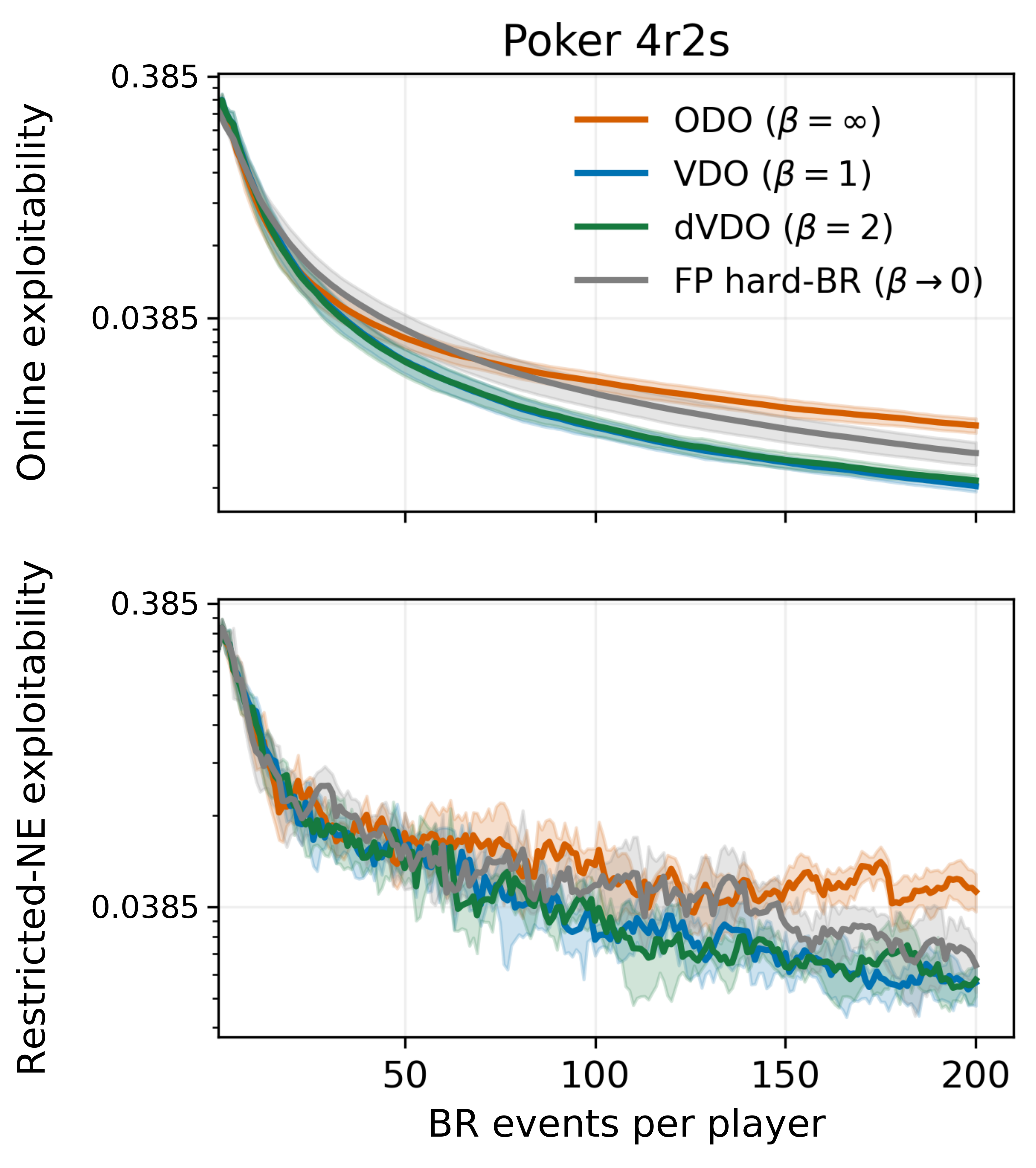}
\caption{Leduc-8.}
\end{subfigure}\hfill
\begin{subfigure}[b]{0.32\textwidth}
\includegraphics[width=\textwidth]{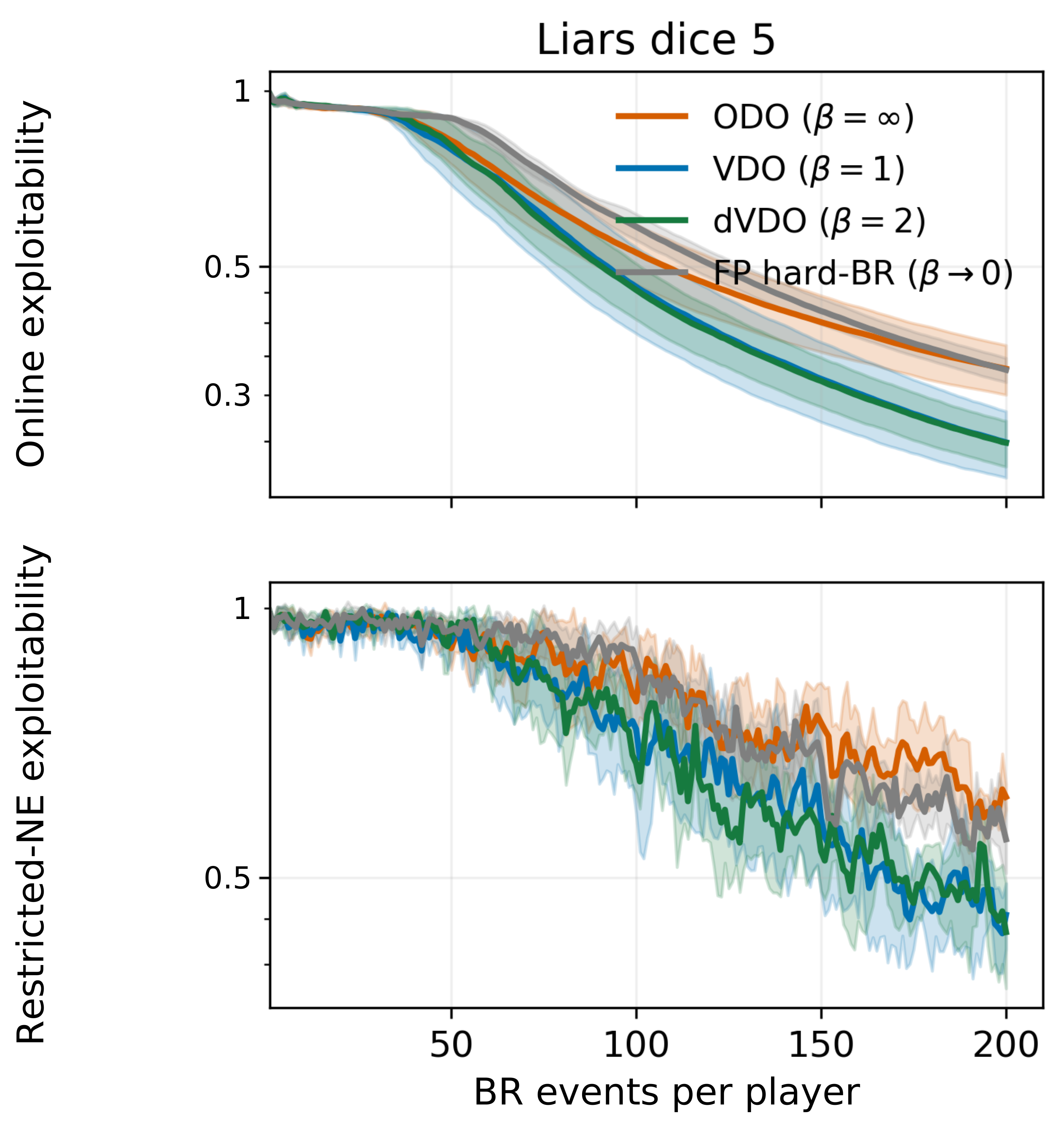}
\caption{Liar's Dice-5.}
\end{subfigure}
\caption{Exact best responses on the
additional games. Online (top) and restricted-NE
(bottom, RNE) exploitability versus BR queries,
with the same baselines as
Figure~\ref{fig:main_results}.}
\label{fig:exact-app}
\end{figure}

\begin{figure}[H]
\centering
\begin{subfigure}[b]{0.32\textwidth}
\includegraphics[width=\textwidth]{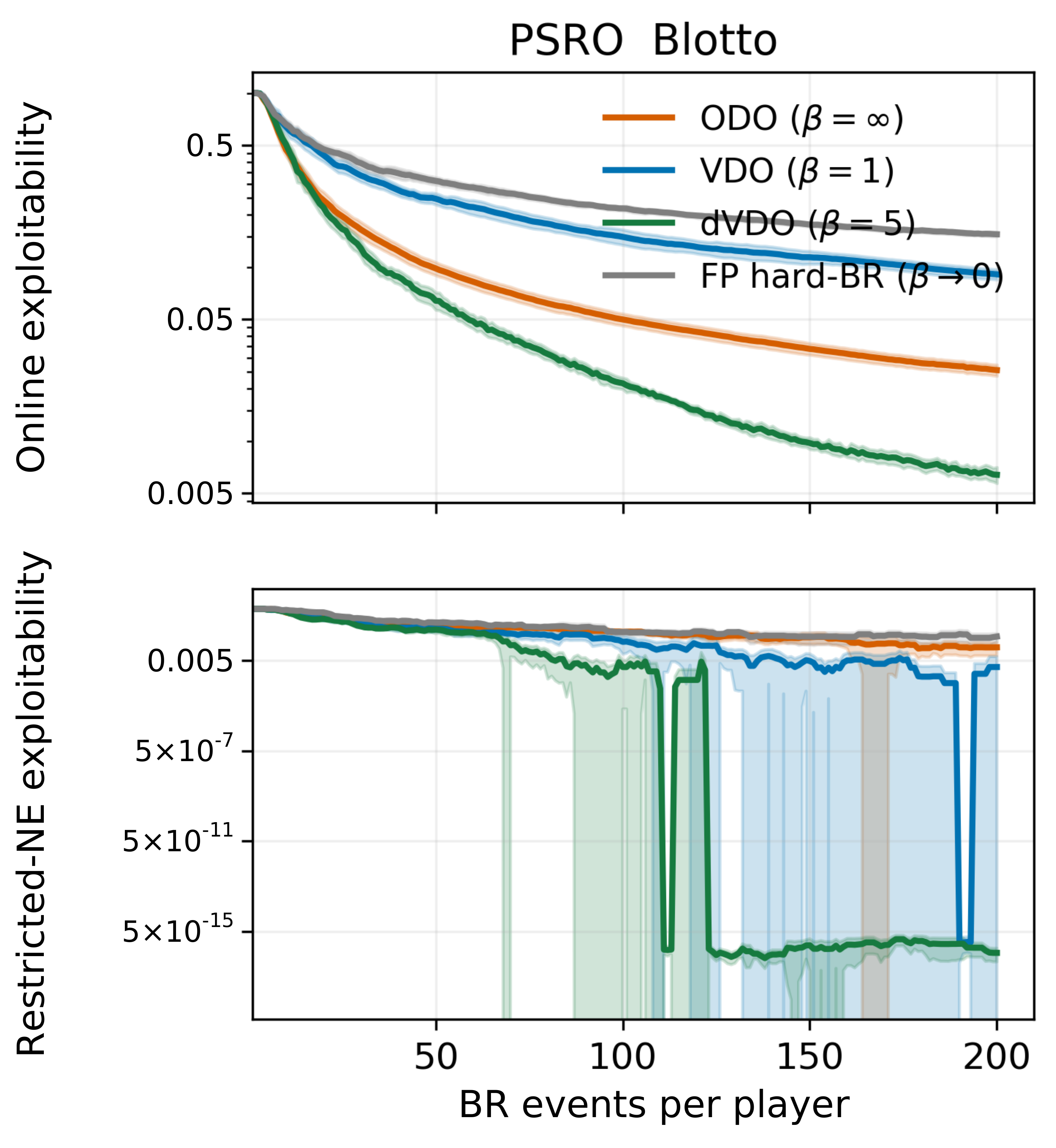}
\caption{Blotto.}
\end{subfigure}\hfill
\begin{subfigure}[b]{0.32\textwidth}
\includegraphics[width=\textwidth]{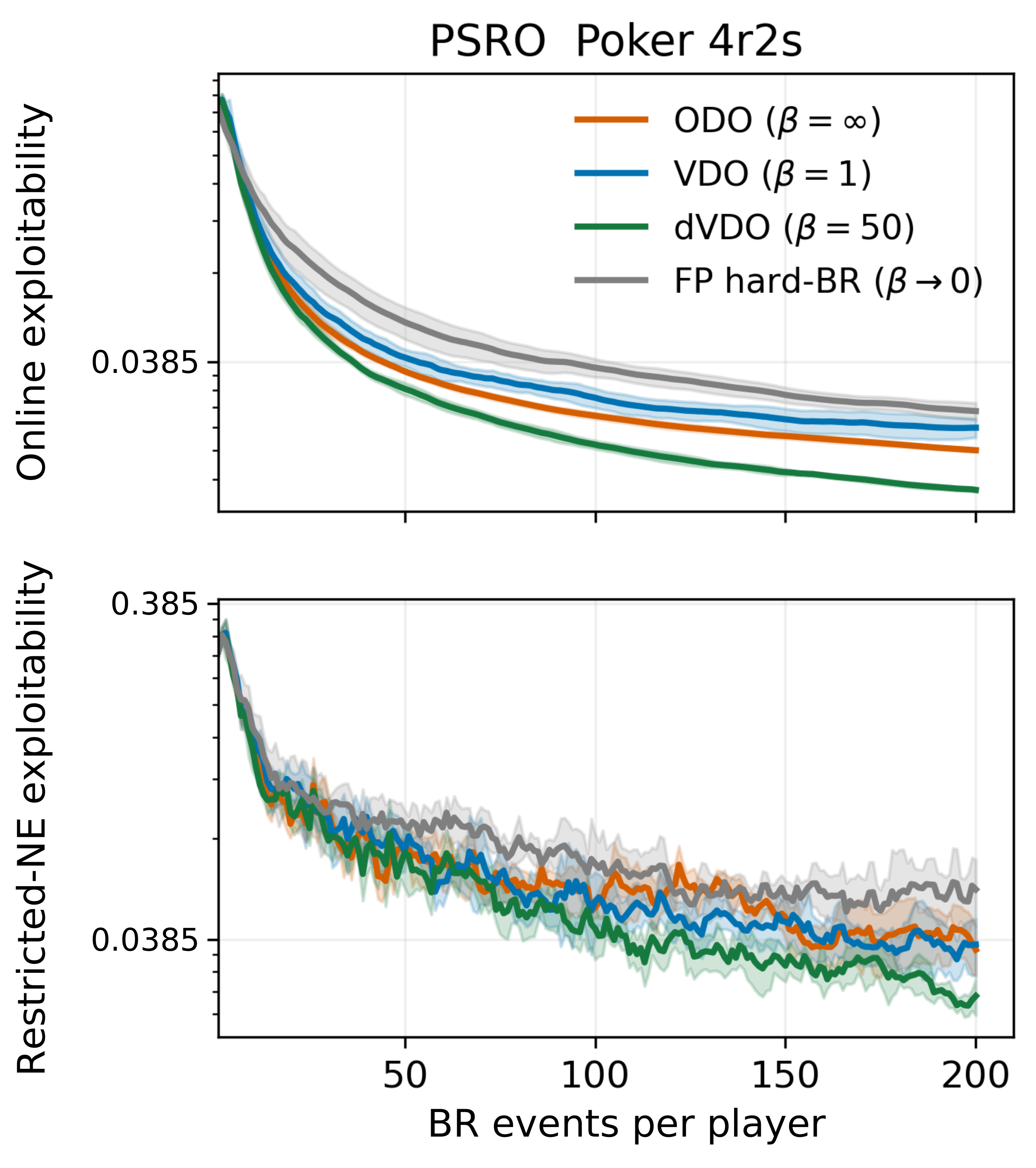}
\caption{Leduc-8.}
\end{subfigure}\hfill
\begin{subfigure}[b]{0.32\textwidth}
\includegraphics[width=\textwidth]{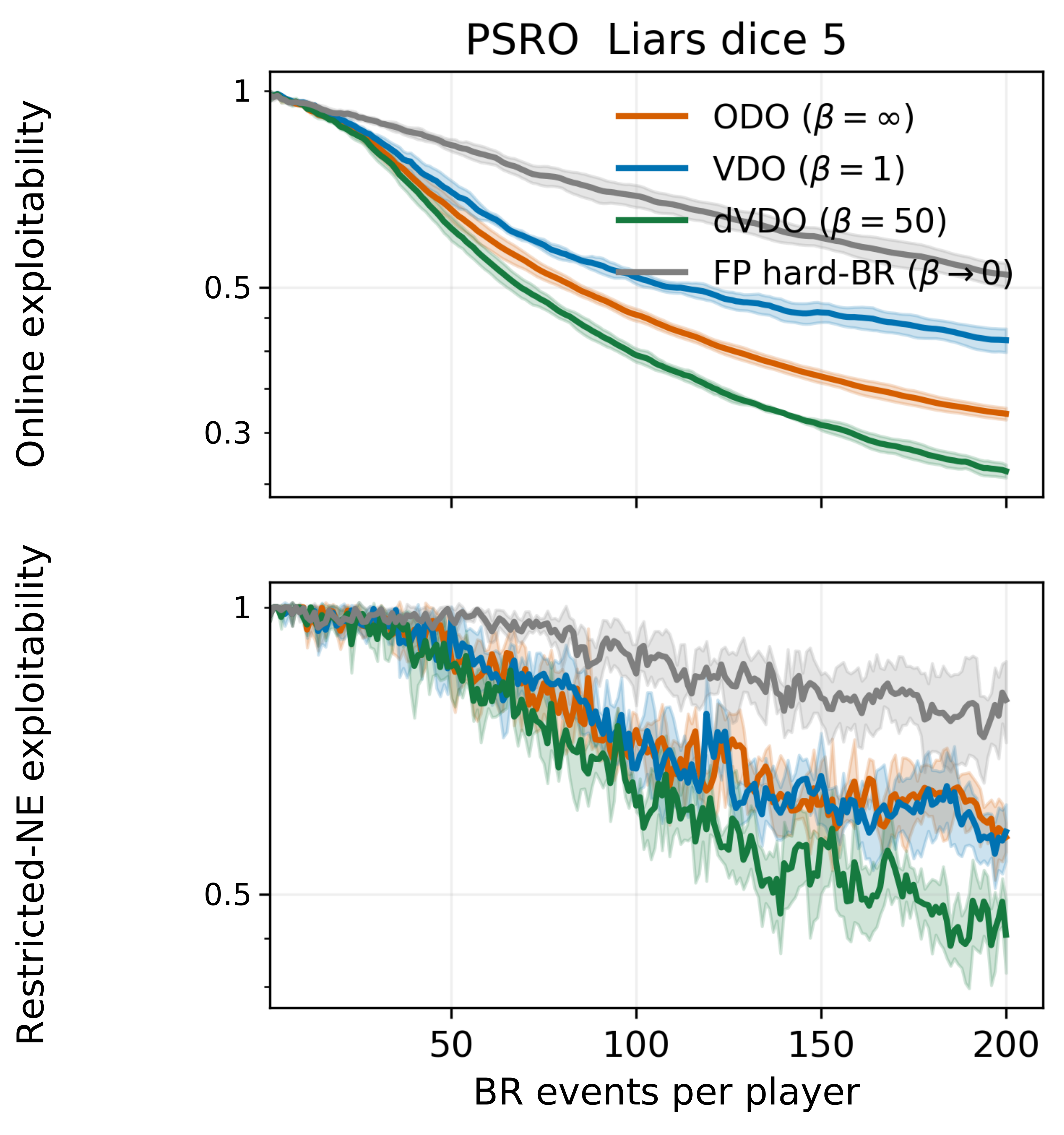}
\caption{Liar's Dice-5.}
\end{subfigure}
\caption{$\varepsilon$-perturbed best
responses ($\varepsilon=0.1$) on the additional
games. Online (top) and restricted-NE (bottom,
RNE) exploitability versus BR queries, with the
same baselines as Figure~\ref{fig:main_results}.
The preferred discount moves to $\disc^*=50$ on
Leduc-8 and Liar's Dice-5, while Blotto keeps a
small discount ($\disc^*=5$).}
\label{fig:eps-app}
\end{figure}

\begin{figure}[H]
\centering
\begin{subfigure}[b]{0.32\textwidth}
\includegraphics[width=\textwidth]{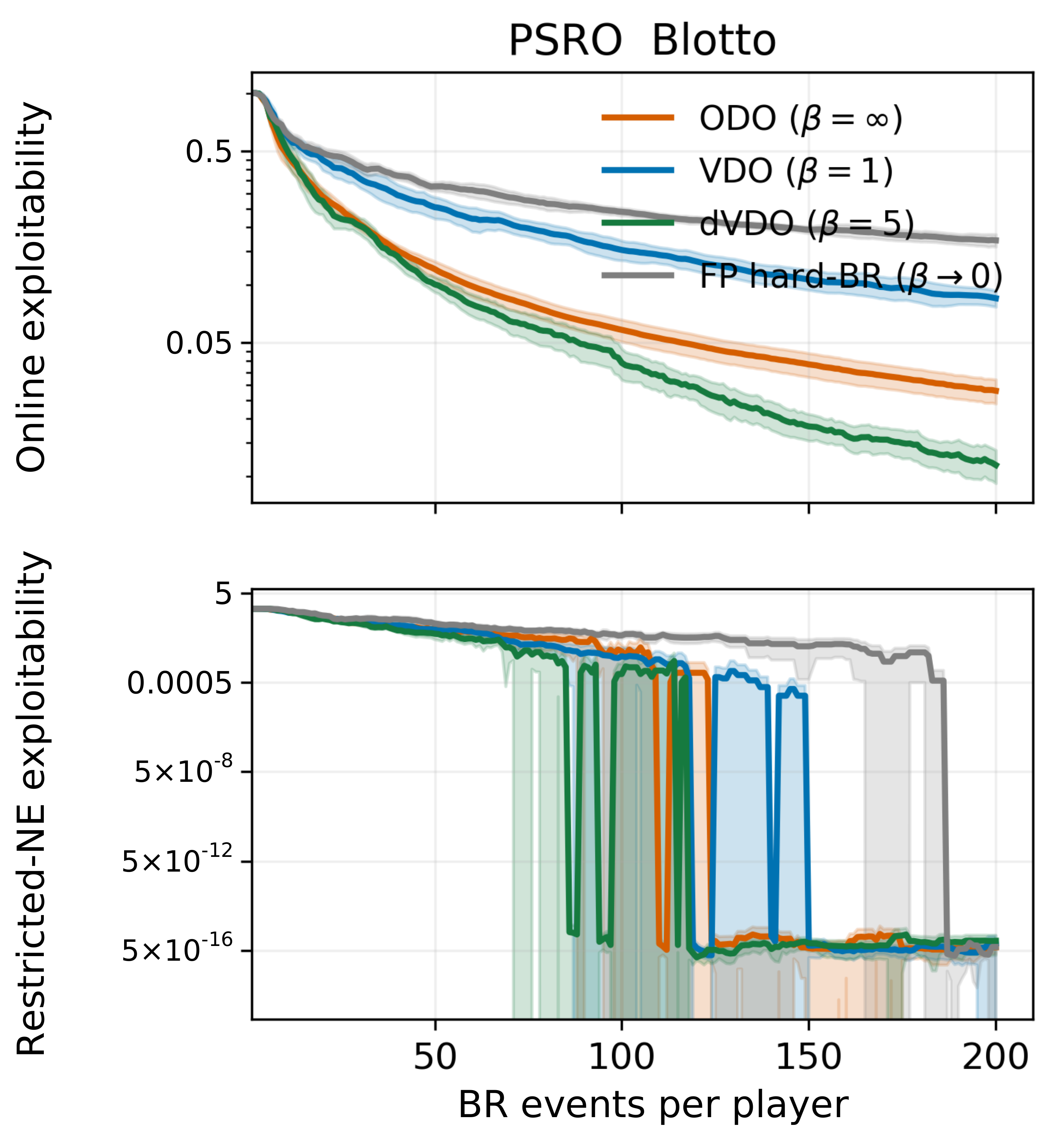}
\caption{Blotto.}
\end{subfigure}\hfill
\begin{subfigure}[b]{0.32\textwidth}
\includegraphics[width=\textwidth]{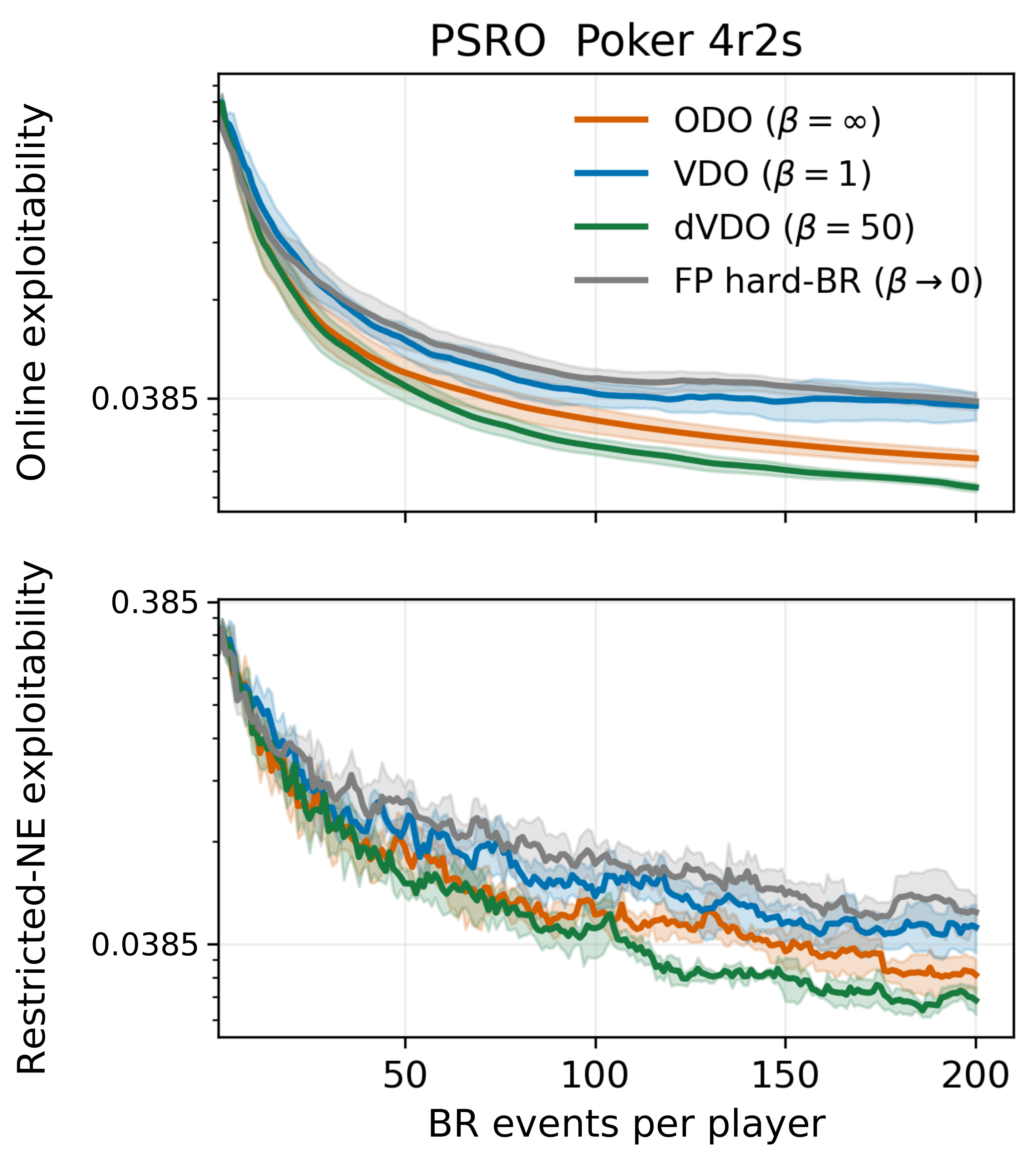}
\caption{Leduc-8.}
\end{subfigure}\hfill
\begin{subfigure}[b]{0.32\textwidth}
\includegraphics[width=\textwidth]{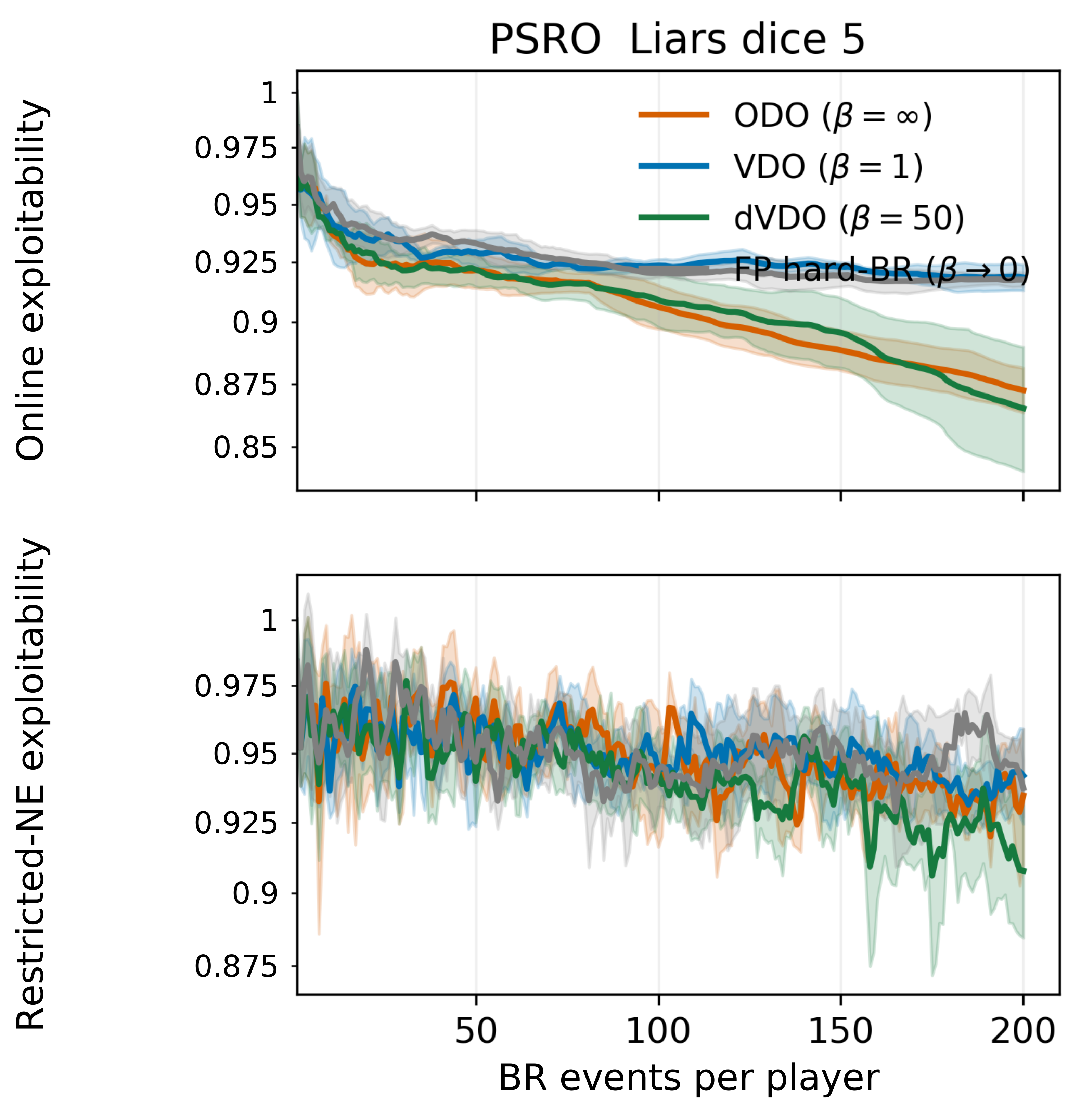}
\caption{Liar's Dice-5.}
\end{subfigure}
\caption{Sampled-target best responses
($N=10$ opponent samples) on the additional games.
Online (top) and restricted-NE (bottom, RNE)
exploitability versus BR queries, with the same
baselines as Figure~\ref{fig:main_results}. dVDO
($\disc^*=50$) is lowest online on Leduc-8 and
Liar's Dice-5, though on Liar's Dice-5 its margin
over ODO is within the seed spread. Blotto keeps a
small discount ($\disc^*=5$), and every method
reaches an RNE of $\approx 0$ on it.}
\label{fig:sampled-app}
\end{figure}

This appendix reports three games not
plotted in Section~\ref{sec:experiments}: Colonel
Blotto (10 coins over 3 fields; a single decision
per player, so effectively a $66\times 66$ matrix
game), Leduc-8 (Leduc's betting structure on an
eight-card deck of four ranks $\times$ two suits,
testing whether the Leduc results scale with deck
size) and Liar's Dice-5 (one five-sided die per
player). Figures~\ref{fig:exact-app},
\ref{fig:eps-app} and~\ref{fig:sampled-app} show
exact, $\varepsilon$-perturbed and sampled-target
best responses, respectively.
Table~\ref{tab:final} reports the final values on
these games and on Kuhn, Leduc and Goofspiel-4,
including DO. Baselines, styles and confidence
intervals are as in
Figure~\ref{fig:main_results}. Blotto uses 10
seeds; Leduc-8 and Liar's Dice-5 use 5 due to
higher per-seed cost. With exact BRs, dVDO-best is
the best $\disc$ from $\{2,5,10,20,50\}$ per game;
with approximate BRs the sweep is $\{2,5\}$ on Kuhn
and Blotto, $\{2,50\}$ on Leduc and $\{2,10,50\}$
on the remaining games. In every setting $\disc^*$
is the value with the lowest online exploitability
at response event 200 per player to the precision reported. On Kuhn with exact BRs, $\disc=2$ and $\disc=5$ both give $0.00120$ to three significant figures; the figure and table show $\disc=2$.

\begin{table}[H]
\centering
\caption{Mean normalised half NashConv at response event 200 per player.
Bold marks the lowest displayed value among ODO, VDO and dVDO.
The dVDO discount in parentheses is selected by mean online exploitability
and is used in both halves of the table. DO uses the exact oracle.
Values below $10^{-9}$ are shown as $\approx 0$.}
\label{tab:final}
\small
\setlength{\tabcolsep}{4pt}
\begin{tabular}{llccccc}
\toprule
Game & Oracle & ODO & VDO & dVDO-best ($\disc^*$) & FP hard-BR & DO \\
\midrule
\multicolumn{7}{l}{\emph{Online exploitability}} \\
\midrule
Kuhn poker & exact & 0.00186 & 0.00182 & \textbf{0.00120} (2) & 0.00821 & $\approx 0$ \\
 & $\varepsilon=0.1$ & 0.00391 & 0.00155 & \textbf{0.000496} (2) & 0.00843 & -- \\
 & sampled & 0.00243 & 0.00201 & \textbf{0.000888} (2) & 0.00866 & -- \\
\addlinespace
Leduc poker & exact & 0.0142 & \textbf{0.00969} & 0.00991 (2) & 0.0133 & 0.00481 \\
 & $\varepsilon=0.1$ & 0.0218 & 0.0307 & \textbf{0.0166} (50) & 0.0362 & -- \\
 & sampled & 0.0303 & 0.0443 & \textbf{0.0281} (50) & 0.0443 & -- \\
\addlinespace
Goofspiel-4 & exact & 0.172 & 0.130 & \textbf{0.113} (50) & 0.137 & 0.0120 \\
 & $\varepsilon=0.1$ & 0.208 & 0.300 & \textbf{0.155} (50) & 0.313 & -- \\
 & sampled & 0.227 & 0.307 & \textbf{0.193} (50) & 0.320 & -- \\
\addlinespace
Blotto & exact & 0.0260 & 0.0915 & \textbf{0.00861} (2) & 0.146 & $\approx 0$ \\
 & $\varepsilon=0.1$ & 0.0255 & 0.0907 & \textbf{0.00639} (5) & 0.154 & -- \\
 & sampled & 0.0280 & 0.0850 & \textbf{0.0115} (5) & 0.171 & -- \\
\addlinespace
Leduc-8 & exact & 0.0139 & \textbf{0.00780} & 0.00822 (2) & 0.0107 & 0.00504 \\
 & $\varepsilon=0.1$ & 0.0193 & 0.0230 & \textbf{0.0141} (50) & 0.0262 & -- \\
 & sampled & 0.0253 & 0.0366 & \textbf{0.0206} (50) & 0.0377 & -- \\
\addlinespace
Liar's Dice-5 & exact & 0.333 & 0.249 & \textbf{0.248} (2) & 0.331 & 0.104 \\
 & $\varepsilon=0.1$ & 0.320 & 0.415 & \textbf{0.262} (50) & 0.524 & -- \\
 & sampled & 0.872 & 0.919 & \textbf{0.865} (50) & 0.918 & -- \\
\midrule
\multicolumn{7}{l}{\emph{Restricted-NE (RNE) exploitability}} \\
\midrule
Kuhn poker & exact & $\mathbf{\approx 0}$ & $\mathbf{\approx 0}$ & $\mathbf{\approx 0}$ (2) & $\approx 0$ & $\approx 0$ \\
 & $\varepsilon=0.1$ & $\mathbf{\approx 0}$ & $\mathbf{\approx 0}$ & $\mathbf{\approx 0}$ (2) & $\approx 0$ & -- \\
 & sampled & $\mathbf{\approx 0}$ & $\mathbf{\approx 0}$ & $\mathbf{\approx 0}$ (2) & $\approx 0$ & -- \\
\addlinespace
Leduc poker & exact & 0.0513 & \textbf{0.0271} & 0.0273 (2) & 0.0329 & 0.00481 \\
 & $\varepsilon=0.1$ & 0.0462 & 0.0408 & \textbf{0.0311} (50) & 0.0490 & -- \\
 & sampled & 0.0341 & 0.0458 & \textbf{0.0305} (50) & 0.0515 & -- \\
\addlinespace
Goofspiel-4 & exact & 0.366 & \textbf{0.181} & 0.238 (50) & 0.236 & 0.0120 \\
 & $\varepsilon=0.1$ & 0.397 & 0.330 & \textbf{0.276} (50) & 0.365 & -- \\
 & sampled & 0.233 & 0.296 & \textbf{0.198} (50) & 0.321 & -- \\
\addlinespace
Blotto & exact & 0.0527 & 0.00800 & $\mathbf{\approx 0}$ (2) & 0.0816 & $\approx 0$ \\
 & $\varepsilon=0.1$ & 0.0197 & 0.00263 & $\mathbf{\approx 0}$ (5) & 0.0591 & -- \\
 & sampled & $\mathbf{\approx 0}$ & $\mathbf{\approx 0}$ & $\mathbf{\approx 0}$ (5) & $\approx 0$ & -- \\
\addlinespace
Leduc-8 & exact & 0.0433 & \textbf{0.0217} & 0.0221 (2) & 0.0247 & 0.00504 \\
 & $\varepsilon=0.1$ & 0.0358 & 0.0371 & \textbf{0.0261} (50) & 0.0542 & -- \\
 & sampled & 0.0313 & 0.0431 & \textbf{0.0263} (50) & 0.0477 & -- \\
\addlinespace
Liar's Dice-5 & exact & 0.616 & 0.455 & \textbf{0.435} (2) & 0.552 & 0.104 \\
 & $\varepsilon=0.1$ & 0.574 & 0.582 & \textbf{0.454} (50) & 0.800 & -- \\
 & sampled & 0.935 & 0.942 & \textbf{0.908} (50) & 0.937 & -- \\
\bottomrule
\end{tabular}
\end{table}

\clearpage
\begingroup
\section{Insertion-Cost Measurements}
\label{app:insertion_experiments}

We measure the insertion budgets of VDO and dVDO ($\disc=5$) in six
normalised settings: Gaussian matrix games of sizes $25$, $50$, $100$ and
$250$, Kuhn poker, and Leduc poker. We use seeds $101$--$103$ and horizons
$T\in\{1{,}024,4{,}096,16{,}384,65{,}536\}$. Each matrix size has three
independent Gaussian draws, generated with seed $1000+$ run seed and divided
by the largest absolute entry of that matrix. Kuhn and Leduc payoffs are
divided by $2$ and $13$, respectively. The poker seeds vary the initial
policies. The four horizons are prefixes of the same fixed-rate run.

Both methods start with one pure strategy per player and use $\eta=1/11$,
simultaneous MWU updates, and exact best responses to the full-history
opponent average. Each player uses the countdown
$\max\{1,\lfloor50\ln|\Pi_i^t|\rfloor\}$ and reinitialises only when its
own population expands. Historical weights are reconstructed at rate
$\gamma=\eta$ for VDO and $\gamma=\eta/5$ for dVDO. We use the released MWU
meta-solver with exact matrix utilities and an information-set
dynamic-programming oracle for poker. The poker best-response values were
checked against a sequence-form linear program on $40$ policy mixtures,
with maximum discrepancy $4.45\times10^{-16}$.

For each method, $C_{i,t}$ is the initial log-population term plus the
cumulative insertion costs evaluated at its reconstruction rate. We divide
this budget by the current number of distinct retained pure strategies,
$|\Pi_i^t|$, and record the ratio after every insertion. Costs are computed
from historical log-weight sums. Table~\ref{tab:insertion_cost} reports the
maximum over both players, all seeds, all horizons and every recorded
insertion prefix in each setting. These fixed-rate, finite-horizon
measurements support the plausibility of a linear budget in the tested
regime; the convergence guarantee remains conditional on the budget at
the chosen learning rate.

\begin{table}[H]
\centering
\caption{Maximum observed insertion-budget ratio $C_{i,t}/|\Pi_i^t|$
with normalised payoffs and global-history best responses. Costs use each
method's reconstruction rate.}
\label{tab:insertion_cost}
\begin{tabular}{lcc}
\toprule
Setting & VDO & dVDO ($\disc=5$) \\
\midrule
Gaussian $25\times25$ & 2.7707 & 0.8221 \\
Gaussian $50\times50$ & 3.1464 & 0.6582 \\
Gaussian $100\times100$ & 3.8767 & 0.7348 \\
Gaussian $250\times250$ & 3.9568 & 0.8537 \\
Kuhn poker & 0.7126 & 0.4514 \\
Leduc poker & 1.0270 & 0.3966 \\
\bottomrule
\end{tabular}
\end{table}
\endgroup

\end{document}